\documentclass[%
 reprint,
 amsmath,amssymb,
 aps,
nofootinbib,
superscriptaddress,
]{revtex4-2}
\usepackage{graphicx}
\usepackage{epstopdf}
\usepackage{color}
\usepackage{multirow}
\usepackage{subeqnarray}
\usepackage{setspace}
\usepackage{palatino}
\usepackage{dcolumn}
\usepackage{bm}
\usepackage[utf8]{inputenc}
\usepackage[T1]{fontenc}
\usepackage{mathptmx}
\usepackage{amsmath,amsfonts,amscd,amssymb,amsthm}
\usepackage{commath}

\usepackage{algorithm,algcompatible,algorithmicx}
\usepackage{algpseudocode}

\algnewcommand\algorithmicinput{\textbf{Input:}}
\algnewcommand\INPUT{\item[\algorithmicinput]}
\algnewcommand\algorithmicoutput{\textbf{Output:}}
\algnewcommand\OUTPUT{\item[\algorithmicoutput]}
\algnewcommand\algorithmicoptional{\textbf{Optional:}}
\algnewcommand\OPTIONAL{\item[\algorithmicoptional]}
\usepackage{overpic}
\usepackage{cancel}
\usepackage{rotating}
\usepackage{url}
\usepackage{caption}
\usepackage{subcaption}
\usepackage{hyperref}
\usepackage{mathtools}

\DeclareMathOperator*{\argmin}{arg\,min}
\usepackage{setspace}
\usepackage[bottom,flushmargin,hang,multiple,symbol]{footmisc}
\newcommand\blfootnote[1]{
  \begingroup
  \renewcommand\thefootnote{}\footnote{#1}
  \addtocounter{footnote}{-1}
  \endgroup
}

\begin{document} 

\title{Benchmarking sparse system identification with low-dimensional chaos}

\author{Alan A. Kaptanoglu}\thanks{Corresponding author (akaptano@uw.edu).}
\affiliation{IREAP, University of Maryland, College Park, MD, USA\looseness=-1} 
\affiliation{
 Department of Mechanical Engineering, University of Washington, Seattle, WA, USA
}
\author{Lanyue Zhang}
\affiliation{
 Department of Applied Mathematics, University of Washington, Seattle, WA, USA  
}
\author{Zachary G. Nicolaou} 
\affiliation{
 Department of Applied Mathematics, University of Washington, Seattle, WA, USA
}
\author{Urban Fasel}
\affiliation{
 Department of Aeronautics, Imperial College, London, UK
}
\author{Steven L. Brunton}
\affiliation{
 Department of Mechanical Engineering, University of Washington, Seattle, WA, USA
}

 \begin{abstract}
 Sparse system identification is the data-driven process of obtaining parsimonious differential equations that describe the evolution of a dynamical system, balancing model complexity and accuracy. There has been rapid innovation in system identification across scientific domains, but there remains a gap in the literature for large-scale methodological comparisons that are evaluated on a variety of dynamical systems. 
 In this work, we systematically benchmark sparse regression variants by utilizing the dysts standardized database of chaotic systems introduced by Gilpin~\cite{gilpin2021chaos}.  
 In particular, we demonstrate how this open-source tool can be used to quantitatively compare different methods of system identification. 
 To illustrate how this benchmark can be utilized, we perform a large comparison of four algorithms for solving the sparse identification of nonlinear dynamics (SINDy) optimization problem, finding strong performance of the original algorithm and a recent mixed-integer discrete algorithm. In all cases, we used ensembling to improve the noise robustness of SINDy and provide statistical comparisons.
 In addition, we show very compelling evidence that the weak SINDy formulation provides significant improvements over the traditional method, even on clean data.
Lastly, we investigate how Pareto-optimal models generated from SINDy algorithms depend on the properties of the equations, finding that the performance shows no significant dependence on a set of dynamical properties that quantify the amount of chaos, scale separation, degree of nonlinearity, and the syntactic complexity. 
 \end{abstract}

 \maketitle

\section{Introduction\label{sec:intro}}
The governing equations of a dynamical system have traditionally been derived from first principles or phenomenology and confirmed with experimentation. However, the availability of enormous volumes of data in the modern era is facilitating \textit{data-driven} model discovery, which is now widely applied to many areas such as mathematics, physics, chemistry, biology, engineering, and economics. 
A large number of such approaches have been developed in recent years, such as the dynamic mode decomposition~\cite{schmid2010dynamic,Rowley2009jfm,Kutz2016book}, Koopman theory~\cite{Mezic2005nd,Brunton2021koopman}, nonlinear autoregressive algorithms~\cite{Billings2013book}, neural networks~\cite{pathak2018model,vlachas2018data,Raissi2019jcp}, Gaussian process regression~\cite{raissi2017machine}, operator inference and reduced-order modeling~\cite{Benner2015siamreview,peherstorfer2016data,qian2020lift}, symbolic regression~\cite{Bongard2007pnas,schmidt_distilling_2009}, divide-and-conquer strategies~\cite{udrescu2020ai}, and sparse regression~\cite{Brunton2016pnas}. Even within each of these approaches, there has been prolific methodological innovation.

There are several factors that must be considered when choosing a data-driven modeling strategy, including the quantity and quality of the data and the ultimate use of the model. 
A key tradeoff for all of these methods is balancing accuracy with model complexity, since a naíve determination of many free, nonzero parameters will likely result in a model that is overfit to the training data, especially as real-world data will contain noise of various types. Despite the variety of new methods for data-driven model discovery, there is a relative lack of comparisons between such methods on a large set of nontrivial dynamical systems. It is this gap in the literature that we intend to address by providing a systematic comparison of sparse system identification variants on a recently developed large, curated database of known, chaotic, dynamical systems~\cite{gilpin2021chaos}. 

Sparse system identification is an umbrella term for system identification methods that address this key tradeoff between model complexity and accuracy by promoting sparsity in the parameterization of the model. This facilitates a-priori parsimonious descriptions that model the data with as few terms as necessary. The sparse identification of nonlinear dynamics (SINDy)~\cite{Brunton2016pnas} is a leading method for parsimonious modeling, and it and its variants will be investigated here. It is based on sparsity-regularized linear regression, so it can be computed quickly and robustly. Moreover, identified models and optimization results can be readily understood by researchers, as there is a wealth of literature on generalized linear regression and sparse optimization. 
Another compelling reason to investigate variations of SINDy is that it has seen rapid application and extension since its introduction in 2016. SINDy has been widely applied for model identification in applications such as chemical reaction dynamics~\cite{hoffmann2019reactive}, chemical processes~\cite{bhadriraju2020operable,scheffold2021gray}, air pollutant modeling~\cite{rubio2022modeling}, biological transport~\cite{lagergren2020learning}, stellar dynamics~\cite{pasquato2022sparse}, disease transmission~\cite{jiang2021modeling}, convective heat transfer~\cite{zucatti2020assessment}, nonlinear optics~\cite{Sorokina2016oe}, power systems~\cite{stankovic2020data,cai2022online}, hydraulics~\cite{narasingam2018data}, scientific computing~\cite{Thaler2019SparseIO}, human behavior models~\cite{dale2018equations}, fluid dynamics~\cite{loiseau2018constrained,Loiseau2018jfm,loiseau2020data,el2018sparse,chang2019machine,deng2020low,fukami2021sparse,callaham2022role,khoo2022sparse,deng2021galerkin,xiao2023construction,foster2022estimating}, turbulence modeling~\cite{schmelzer2020discovery,beetham2020formulating,beetham2021multiphase,beetham2021sparse,callaham2022empirical,sansica2022system}, plasma physics~\cite{Dam2017pf,kaptanoglu2021physics,alves2022data}, structural modeling~\cite{lai2019sparse}, among others~\cite{narasingam2018data,de2020discovery,pan2021sparsity,subramanian2021white,brenner2022tractable,zhang2022knowledge,golden2022physically,joshi2022data}.
It has also been extended to handle more complex modeling scenarios such as  PDEs~\cite{Schaeffer2017prsa,Rudy2017sciadv}, delay equations~\cite{sandoz2022sindy}, stochastic differential equations~\cite{klimovskaia2016sparse,bruckner2020inferring,dai2020detecting,callaham2021nonlinear,huang2022sparse}, Bayesian modelling~\cite{hirsh2022sparsifying}, dimensional analysis~\cite{bakarji2022dimensionally}, systems with inputs or control~\cite{brunton2016sparse,Kaiser2018prsa,kaiser2018discovering}, systems with implicit dynamics~\cite{Mangan2016ieee,kaheman2020sindy}, hybrid systems~\cite{mangan2019model,thiele2020system}, to enforce physical constraints~\cite{loiseau2018constrained, champion2020unified, kaptanoglu2021physics}, to incorporate information theory~\cite{mangan2017model} or group sparsity~\cite{dong2022improved} or global stability~\cite{kaptanoglu2021promoting}, to identify models from corrupt or limited data~\cite{tran2017exact,schaeffer2018extracting,delahunt2022toolkit,wentz2022derivative,kaheman2022automatic}, to identify models with partial measurements of the state space~\cite{somacal2022uncovering,bakarji2022discovering,conti2022reduced,gao2022bayesian}, to identify models with clever subsampling strategies~\cite{zhao2022adaptive},
and ensembles of initial conditions~\cite{wu2019numerical}, to perform cross-validation with ensemble methods~\cite{fasel2022ensemble,gao2023convergence}, and extending to related objective functions~\cite{jiang2023regularized}, a weak or integral formulation~\cite{Schaeffer2017pre,gurevich2019robust,reinbold2020using,messenger2021weakpde,kageorge2021data,reinbold2021robust,gurevich2021learning,russo2022convergence,messenger2022asymptotic,messenger2022learning,messenger2022online}, tensor representations~\cite{gelss2019multidimensional,goessmann2020tensor}, and stochastic forcing~\cite{boninsegna2018sparse}.

Most of the work in this field demonstrates performance on just a few canonical dynamical systems, such as the Lorenz63 system~\cite{lorenz1963deterministic}.
Moreover, there has been little systematic investigation of the comparative performance of different sparse system identification techniques.
We posit two plausible reasons for the relative lack of these investigations in the literature: (1) for more sophisticated variants of system identification, it can be computationally intensive to compute thousands or hundreds of thousands of dynamical models to draw statistical conclusions like those presented here, and (2) an analysis of the performance of system identification is more straightforward if the true governing equations are actually known. This is clearly not an option for real experimental data. 
With benchmarks, proper comparisons can be made between new and existing methods to understand the way in which a new algorithm produces novel benefits and functionality. Moreover, essentially all system identification methods require hyperparameter tuning and a proper comparison between methods necessitates that only the optimally tuned models from each method are compared on a large set of examples.

At first glance, there may seem to be too many confounding variables for a thorough comparison of methods.  
In principle, sparse system identification performance can depend on the equation complexity, the searchable space of functional forms, the size and diversity of the training data, the amount of computing power available for hyperparameter scans, the nonlinearity of the underlying dynamics, the amount and type of noise in the data, the quality of the algorithm used to solve the problem (and its associated hyperparameters), and potentially many more factors. 

\subsection{Contributions of this work}\label{sec:contributions}
To control for these parameters, we have carefully designed our experiments in a number of ways. We use a fixed functional space for fitting all systems, containing all possible terms necessary for successful model identification. Moreover, the dynamics and size of the training data are standardized by using all of the polynomial dynamical systems in the dysts database~\cite{gilpin2021chaos}, and the amount of computing power is negligible by using sparse regression algorithms with minimal hyperparameters. For generating robust conclusions, the following steps have been taken: $d\%$ noise is sampled from a zero-mean Gaussian $\mathcal{N}(0, \sigma=d \|\bm X_\text{train}\|_2 / 100)$ and added to every data point in the training data $\bm X_\text{train}$ (and the noise is subsequently amplified by computing $\dot{\bm X}_\text{train}$ from $\bm X_\text{train}$). Moreover, a large ensemble of SINDy models are generated by sub-sampling the same data~\cite{fasel2022ensemble}, hyperparameter scans are performed, and finally, only statistical model performance metrics are reported. This work begins by detailing this standardization and methodology.

The entirety of the methods and results in the present work can be reproduced with the \href{https://github.com/dynamicslab/pysindy}{PySINDy code}~\cite{silva2020pysindy,Kaptanoglu2022}. Moreover, we have included a discussion in Appendix~\ref{sec:appendix_benchmark_use} about how to effectively use this tool in future work, as researchers design new algorithms, libraries, databases, or other functionality. The present work illustrates the scope and flexibility of this tool by performing a few large-scale investigations regarding SINDy optimization and performance.

First, we generate a statistical comparison of different optimization algorithms for solving the SINDy model identification problem across a large database. The STLSQ~\cite{Brunton2016pnas}, SR3~\cite{zheng2019unified}, Lasso~\cite{tibshirani2015statistical}, MIOSR~\cite{bertsimas2022learning}, and weak form~\cite{schaeffer2017learning} algorithms are  compared. To our knowledge, this is the first large-scale and statistical comparison of these algorithms. We find that the original STLSQ algorithm~\cite{Brunton2016pnas} holds up well for performing SINDy optimization and the recent MIOSR algorithm is high performance particularly for low-dimensional or constrained problems. Moreover, we find compelling evidence that the weak formulation of the SINDy problem provides significant performance improvements, even in the clean data setting. 
Next, we classify the Pareto-optimal model performance against properties of the true underlying dynamics. We find that the SINDy method has no significant dependence on the amount of chaos, scale separation, degree of nonlinearity, and even the syntactical complexity of the equations, albeit with some interesting deviations that could merit additional future investigation.

\section{Methodology\label{sec:methods}}
Robust conclusions about the performance of system identification methods require that only standardized comparisons are made. 
We now proceed by explaining the dataset, metrics for measuring sparse system identification performance, how we use hyperparameter scanning to find Pareto optimal models for each system, and explicitly define the dynamical properties to be investigated. This section is intended to clarify how we have carefully managed confounding variables in the system identification in order to draw conclusions with respect to different optimization algorithms and with respect to the dynamical properties of the database.

\begin{figure}
    \centering
    \includegraphics[width=0.97\linewidth]{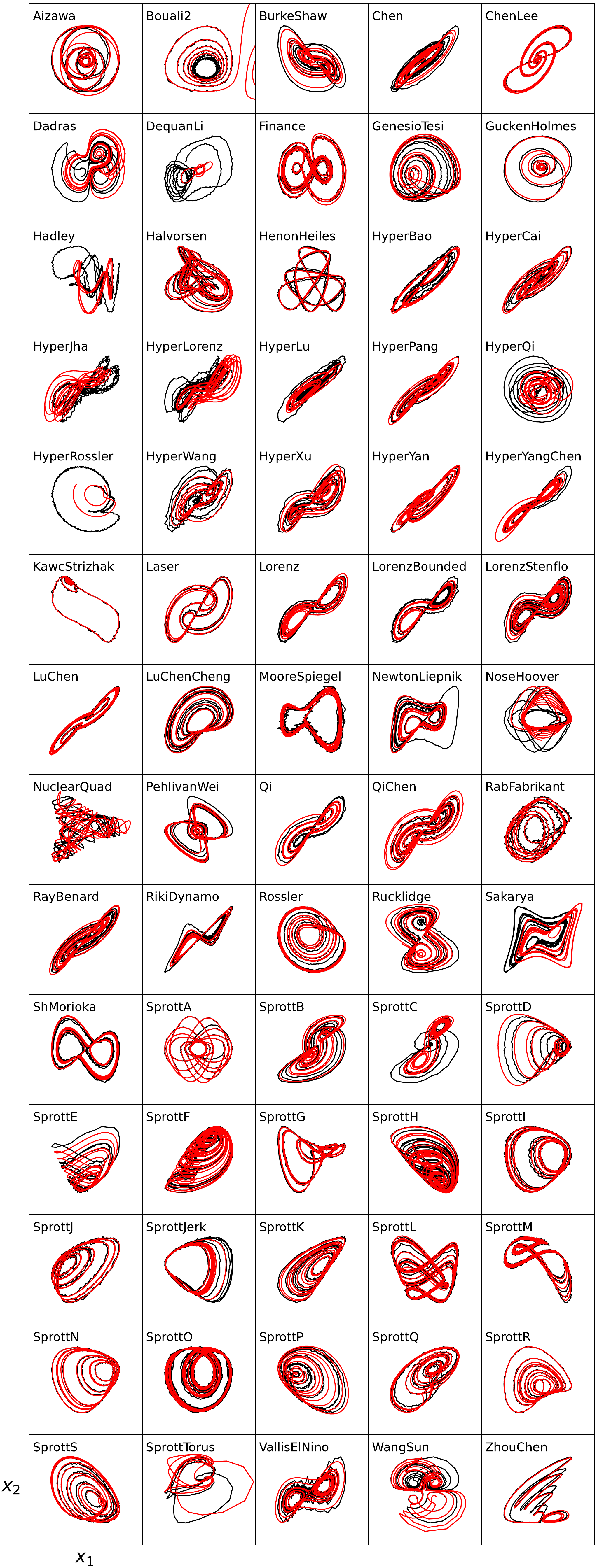}
    \caption{A training trajectory in the $(x_1, x_2)$ subspace with $1\%$ added noise (black), along with a clean testing trajectory (red) for the systems in the present work, taken from~\cite{gilpin2021chaos}. Some of the dynamical systems have multiple attractors, and the data may not be sufficient to fully ``cover'' the attractor. }
    \label{fig:dysts_visualization}
\end{figure}

\subsection{Dysts database}
The dysts database~\cite{gilpin2021chaos}, introduced by Gilpin in 2021, is ideal for evaluating system identification performance, because it is a large and highly standardized set of chaotic systems with known equations of motion.
Many of the chaotic systems exhibit syntactically simple equations, e.g. Lorenz63 is a three-state system with quadratic nonlinearity and only seven terms in the differential equation. However, this database generates far more complicated and dynamical data than that of many existing databases, such as the Nguyen symbolic regression database~\cite{uy2011semantically} used previously for benchmarking symbolic regression algorithms~\cite{petersen2019deep}. The database also provides data, equations, and dynamical properties for over 100 chaotic systems exhibiting strange attractors and coming from disparate scientific fields. It can be used for testing sparse system identification because the dynamical properties have been pre-computed and there is a high degree of standardization across the systems. For instance, phase surrogate significance testing is used to select optimal integration timesteps and sampling rates for all the systems, so that dynamics across systems are aligned with respect to their smallest and largest timescales~\cite{kantz2004nonlinear}. This is a critical step for generating similar time series for each model; the minimum significant time scale in this dataset varies by more than four orders of magnitude. Most importantly, the true governing equations are available for evaluating the model performance.

To narrow the scope of our comparison, we consider 70 systems in the database, representing systems of ordinary differential equations (ODEs) that have polynomial nonlinearities, with degree no more than four. We exclusively consider these systems and define a single ``feature library'' (space of functions that can be used to fit the data) for every model fit. Polynomial models are often the natural choice for capturing leading-order dynamics for general nonlinear systems near an attracting set or fixed point, where the dynamics might be reasonably expanded in a Taylor series. 

Unlike some other dynamical systems, chaotic systems exhibit a continuous spectrum of frequencies, making the dynamical data particularly challenging to systematically distinguish from noise, e.g. by using frequency filters. Our chosen systems exhibit a specific subclass of chaos, as they are all bounded and are characterized by strange attractor(s) with the default system parameters used in the database. Moreover, 54 of the systems have a state space dimension of three, the remaining 16 have a dimension of four, and there is a single Hamiltonian system in the data. 51 of the systems are ``hyper-chaotic'' systems that have been estimated to exhibit two or more positive Lyapunov exponents. A training and testing trajectory of 1000 data points are each visualized in Fig.~\ref{fig:dysts_visualization} so that the strange attractors are apparent.

To generate statistical conclusions, we generate five training trajectories and five testing trajectories representing ten periods of motion, all sampled at 100 points per period for a total of 1000 data points for each trajectory. Each trajectory is generated from a different initial condition on the attractor. 

To summarize, the underlying dynamics are chaotic, and we assume that the dynamical equations can be represented as systems of ODEs with polynomial nonlinearities up to fourth order. Additionally, full state measurements are available that, at various points in the coming analysis, contain added zero-mean Gaussian noise. 

\subsection{The SINDy method}
SINDy utilizes sparse regression to  discover governing equations from data~\cite{Brunton2016pnas}. The aim is to identify differential equation models of the form
\begin{align}
    \label{eq:dx_fx}
    \dot{\bm x} = \bm f(\bm x) \approx \Theta(\bm x)\bm \Xi,
\end{align}
where $\bm f(\bm x)$ is a nonlinear function of the state $\bm x$, $\bm \Theta$ is a library of nonlinear functions, and $\bm \Xi$ is a matrix of constant coefficients. Together $\bm \Theta\bm \Xi$ represents a linear combination of nonlinear functions that are chosen to best approximate $\bm f(\bm x)$. The choice of nonlinear functions in $\bm \Theta$ is clearly important for successful approximation. 
Before moving from continuous state space to discrete measurement data, recall that high order differential equations can always be reduced to first order coupled differential equations, so this form is fairly general. Extensions of Eq.~\eqref{eq:dx_fx} for control inputs~\cite{brunton2016sparse}, partial differential equations~\cite{Rudy2017sciadv}, implicit terms~\cite{kaheman2020sindy}, and other variations are available with PySINDy~\cite{kaptanoglu2021exploration}. 

In order to identify Eq.~\eqref{eq:dx_fx} from data, we need measurements of the state $\bm x$.
First, we uniformly sample the full-state data on a set of training trajectories and a data matrix $\bm{X} \in \mathbb{R}^{M\times d}$ is formed from the time-series data of the state, $\bm{x}(t_1),\bm{x}(t_2),...,\bm{x}(t_M)$:
\begin{eqnarray}
\bm{X} = \overset{\text{\normalsize state}}{\left.\overrightarrow{\overset{~~}{\begin{bmatrix}
x_1(t_1) & x_2(t_1) & \cdots & x_d(t_1)\\
x_1(t_2) & x_2(t_2) & \cdots & x_d(t_2)\\
\vdots & \vdots & \ddots & \vdots \\
x_1(t_M) & x_2(t_M) & \cdots & x_d(t_M)
\end{bmatrix}}}\right\downarrow}\begin{rotate}{270}\hspace{-.125in}time~~\end{rotate} \hspace{.125in}
\label{Eq:DataMatrix}.
\end{eqnarray}
$M$ is the number of time points and $d$ is the size of the state space, i.e. three or four for our dataset. 
A matrix of derivatives in time, $\dot{\bm{X}}$, is defined similarly and can be numerically computed from $\bm{X}$, for instance by finite differences or more sophisticated differentiation~\cite{van2020numerical,van2022pynumdiff,wentz2022derivative}. Then we form a feature library $\bm \Theta(\bm X)$ of possible terms that could describe the evolution of the system, and attempt to find a sparse vector of coefficients $\bm \xi$ representing the best linear combination of terms in the library. In general, the feature library $\bm \Theta$ can include a wide range of generalized functions, and designing this library often requires domain-knowledge; however, the systems of interest in the present work are all polynomial. Therefore, we fix $\bm \Theta$ to be a polynomial library in $\bm X$, up to fourth order. This is an important point; in general system identification methods rely on adequate functional forms in the feature library, and here we have completely bypassed this issue, isolating and disambiguating the optimization performance from the choice of library. We know definitively that the feature library contains the necessary terms in order to describe all 70 chaotic systems considered in the present work. 

The vector of coefficients $\bm{\xi}$ is determined via the following sparse optimization problem:
\begin{align}
\label{eq:optimization_vanilla}
\argmin_{\bm{\xi}}&\left[ \frac{1}{2}\|\bm{\Theta}\bm{\xi}-\dot{\bm{X}}\|^2 + \lambda \|\bm{\xi}\|_0\right].
\end{align}
The first term in the SINDy optimization problem in Eq.~\eqref{eq:optimization_vanilla} is a least-squares fit of a system of ODEs $\bm{\Theta}\bm{\xi}$ to the given data in $\dot{\bm{X}}$.
The $l_0$ loss, $\|\bm{\xi}\|_0$, is a function that counts the number of nonzero elements of $\bm{\xi}$. Using the $l_0$ loss or alternative regularization to promote sparsity in the coefficients of an identified model tends to improve robustness to numerical or experimental noise, addressing the tradeoff between accuracy and generalizability and avoiding overfitting to the training data.
The $l_0$ loss is nonconvex and nonsmooth, and therefore many convex relaxations of this problem have been introduced~\cite{Brunton2016pnas,Rudy2017sciadv,champion2020unified}. The first major result presented in this work compares a number of different algorithms for solving Eq.~\eqref{eq:optimization_vanilla}, including the sequentially-thresholded least-squares (STLSQ) algorithm from the original SINDy paper~\cite{Brunton2016pnas}, the traditional convex relaxation of Eq.~\eqref{eq:optimization_vanilla} using the Lasso~\cite{tibshirani2015statistical}, the SR3 algorithm~\cite{zheng2019unified,champion2020unified}, and a recent mixed-integer discrete algorithm called MIOSR~\cite{bertsimas2022learning}.

As much as is feasible, we use default PySINDy parameters to reduce the number of hyperparameters that must be scanned. For instance, in the present work $\dot{\bm X}$ is calculated by simple finite differences. Improved performance is immediately available for future work using more sophisticated differentation methods~\cite{van2020numerical,van2022pynumdiff,wentz2022derivative}, many of which are already available in the PySINDy code. Weak or integral formulations of SINDy~\cite{Schaeffer2017pre} are also known to mitigate the noise amplification inherent to finite differencing, and we illustrate in Sec.~\ref{sec:weak_form} that the weak formulation implemented in PySINDy produces substantial performance benefits.

\subsection{Metrics for performance}\label{sec:metrics}
Methodological comparisons require appropriate metrics for determining the relative performance quality of different methods.
Unlike real-world data, we have access to the true equations of motion of all 70 systems to be tested. Therefore we can compute a number of useful metrics to characterize the performance of the resulting sparse symbolic models on these examples. We define the normalized coefficient and root-mean-square errors
\begin{align}
\label{eq:model_error}
    \text{E}_\text{coef} &= \frac{\|\bm{\xi}_\text{True}- \bm{\xi}_\text{SINDy}\|_2}{\|\bm{\xi}_\text{True}\|_2}, \\ \label{eq:prediction_error}
    \text{E}_\text{RMSE} &= \frac{\|\dot{\bm{X}}_\text{True}- \dot{\bm{X}}_\text{SINDy}\|_\text{F}}{\|\dot{\bm{X}}_\text{True}\|_\text{F}}.
\end{align}
Other works have also explored the $\|\cdot\|_1$ and $\|\cdot\|_0$ norm versions of Eq.~\eqref{eq:model_error} to quantify model mismatch~\cite{kaheman2020sindy,petersen2019deep}. 
Obtaining very small values for $\text{E}_\text{RMSE}$ (or variants such as the mean absolute error, regular mean absolute percent error, etc.) is simple with a basic system identification algorithm applied to clean data from our chosen database. This is because models with many parameters can overfit using many small but nonzero terms and potentially produce very accurate testing trajectories if we only investigate regions that are nearby to the strange attractor(s). Despite this caveat, for forecasting, prediction, reconstruction of the strange attractor, calculation of the Lyapunov spectrum, and other important metrics for capturing the behavior of a dynamical system, models with low $\text{E}_\text{RMSE}$ are often more than adequate. Moreover, the usual situation of interest is one in which the dynamical equations and therefore the coefficient errors are unknown. We comment further on errors and stability in Appendix~\ref{sec:appendix_stability}.

Given that we know the true coefficients, we can check for overfitting by considering the coefficient errors, which can be incorrect even when $\text{E}_\text{RMSE}$ is extremely small$^*$. Indeed, on the strange attractor, the dynamics may be significantly reduced compared to the dynamics on the full state space. A connection to invariant manifolds and the reduced effective dimension and entropy of the dynamical system are discussed in Appendix~\ref{sec:appendix_dynamical_properties}.
\blfootnote{$^*$Another common metric is the recovery rate~\cite{petersen2019deep}, which returns a discrete 0 or 1 value for each correctly identified term in the equations. However, small errors may result from insufficient hyperparameter tuning, noise in the data, insufficient optimization convergence, and many other sources. Moreover, it is often useful to penalize large coefficient errors more than small ones. Thus, we omit the recovery rate metric in this work but mention it for future investigation.
}

\begin{figure*}
    \centering
    \includegraphics[width=0.95\linewidth]{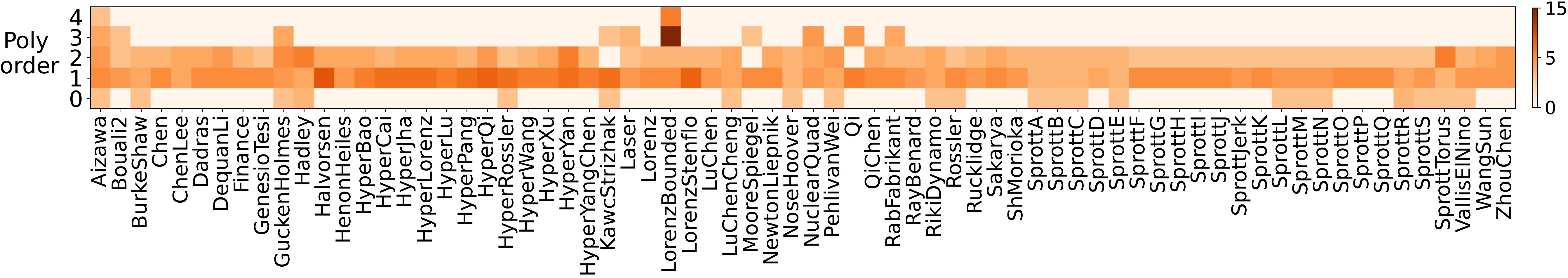}
    \caption{Total number of polynomial terms in each equation, for each polynomial degree. It can be seen that quadratic nonlinearities are dominantly represented.}
    \label{fig:nonlinearity_summary}
\end{figure*}

\subsection{Automated hyperparameter scanning with ensembles}\label{sec:Pareto_scan}
Before any conclusions can be drawn about system identification performance, each algorithm's hyperparameters must be tuned. Sweeping the sparsity-related hyperparameters generates a spectrum ranging from very sparse models with limited accuracy to dense models with high prediction accuracy, which is typically referred to as a Pareto-front~\cite{blasco2008new}. 
The dataset is too large to manually tune hyperparameters for optimal system identification performance on each chaotic trajectory, 
so for hyperparameter scanning, we use the following procedure. For all the algorithms, we use ensembling~\cite{fasel2022ensemble}. We subsample the five training trajectories by choosing 50$\%$ of the total trajectory data without replacement in order to generate $10$ models for each value of the threshold hyperparameter. For STLSQ, we then sweep over 300 values of increasingly large hard thresholds, and determine the best threshold by the minimal, finite-sample-size-corrected, Akaike information criterion (AIC)~\cite{mangan2017model} averaged over the 10 models generated by subsampling,
\begin{align}
    \text{AIC} = M \log(\|\dot{\bm X}_\text{True} - \dot{\bm X}_\text{SINDy}\|_2^2) + 2k + \frac{2k(k + 1)}{(M - k - 1)},
\end{align}
where $k = \|\bm \xi_{SINDy}\|_0$ and $\dot{\bm X}_\text{True}$ and $\dot{\bm X}_\text{SINDy}$ are both computed from the test trajectories.
The AIC is a well-known statistical metric for comparing models while balancing sparsity and accuracy~\cite{mangan2017model}. This systematic process requires the identification of $300 \times 10 \times 70 = 210,000$ models. Despite the large number of models to generate, the entire procedure for STLSQ takes about 10 minutes to run on single Intel Core i9-7920X 2.90GHz CPU. 
After the best threshold is determined via this Pareto sweep, the corresponding ten models allow us to characterize a distribution of identified coefficients and errors. The other SINDy algorithms are tuned similarly for their respective hyperparameters. The SR3 algorithm has an extra parameter $\nu$ that we set to either 1 or $0.1$ since these values tend to work well on clean data and it avoids additional hyperparameter scanning. 

The use of SINDy, rather than more sophisticated system identification methods, facilitates very fast calculation of many models.
For instance, previous work~\cite{la2021contemporary} investigating mostly genetic programming algorithms for symbolic regression required roughly 200 core-hours per algorithm per dynamical system in order to generate a best model from the Pareto front. In contrast, using STLSQ we require roughly $\sim 10^{-3}$ core-hours per dynamical system to generate our Pareto-optimal model, consistent with findings elsewhere~\cite{gilpin2021chaos, orzechowski2018we}. This speed is required because the systematic process we have outlined requires generating $\mathcal{O}(10^5)$ SINDy models per algorithm.

\subsection{Dynamical properties}\label{sec:dynamical_properties}
\begin{figure}
    \centering
    \includegraphics[width=0.95\linewidth]{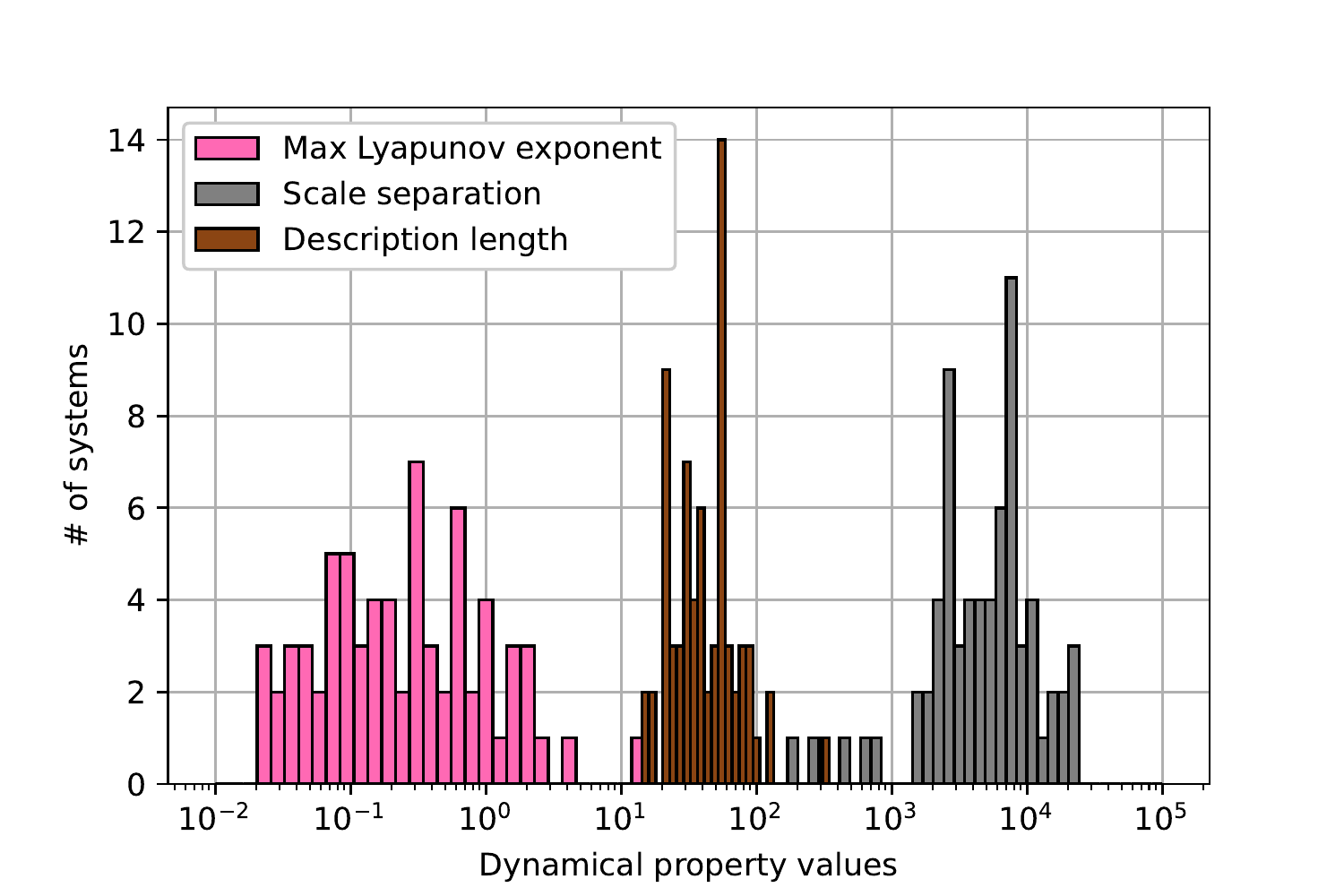}
    \caption{Distributions of the different properties of the dynamical system used in this work.}
    \label{fig:metrics_histograms}
\end{figure}

Systems of differential equations can often be effectively classified by various dynamical properties. 
Now that the dynamical systems used in the present work have been introduced, we define a set of dynamical properties for testing sparse system identification performance: the degree of chaos, the degree of scale separation, the syntactic complexity of the underlying equations, and the amount of nonlinearity. In the present work, the degree of chaos is measured by the largest Lyapunov exponent. The degree of scale separation is taken as the system's dominant timescale divided by the system's smallest significant timescale as determined by the method in Gilpin~\cite{gilpin2021chaos}. The syntactic complexity is quantified as the description length, representing the number of bits to describe every object in the equations. Lastly, the amount of nonlinearity is taken as sum of the total number of terms in the equations, weighted by each term's polynomial degree, so it is a mix between the degree of nonlinearity and the number of equation terms. For more precise definitions and discussion of the metric choices, see Appendix~\ref{sec:appendix_dynamical_properties}. 
A summary of the nonlinearities appearing in the dynamical equations is shown in Fig.~\ref{fig:nonlinearity_summary}, indicating that most of the systems exhibit strong quadratic nonlinearity. Quadratic nonlinearity is common in reduced-order models of fluids and many other dynamical systems, e.g. the Lorenz or Rossler systems, because of the quadratic nonlinearity coming from the convective term of the material derivative. Ten systems exhibit up to cubic nonlinearity and only two systems have quartic nonlinearity.
Fig.~\ref{fig:metrics_histograms} summarizes the distribution of the remaining dynamical properties, indicating a wide range of values taken by the dataset. 

\begin{figure*}
    \centering
    \begin{subfigure}[b]{0.64\textwidth}
         \centering
        \includegraphics[width=\linewidth]{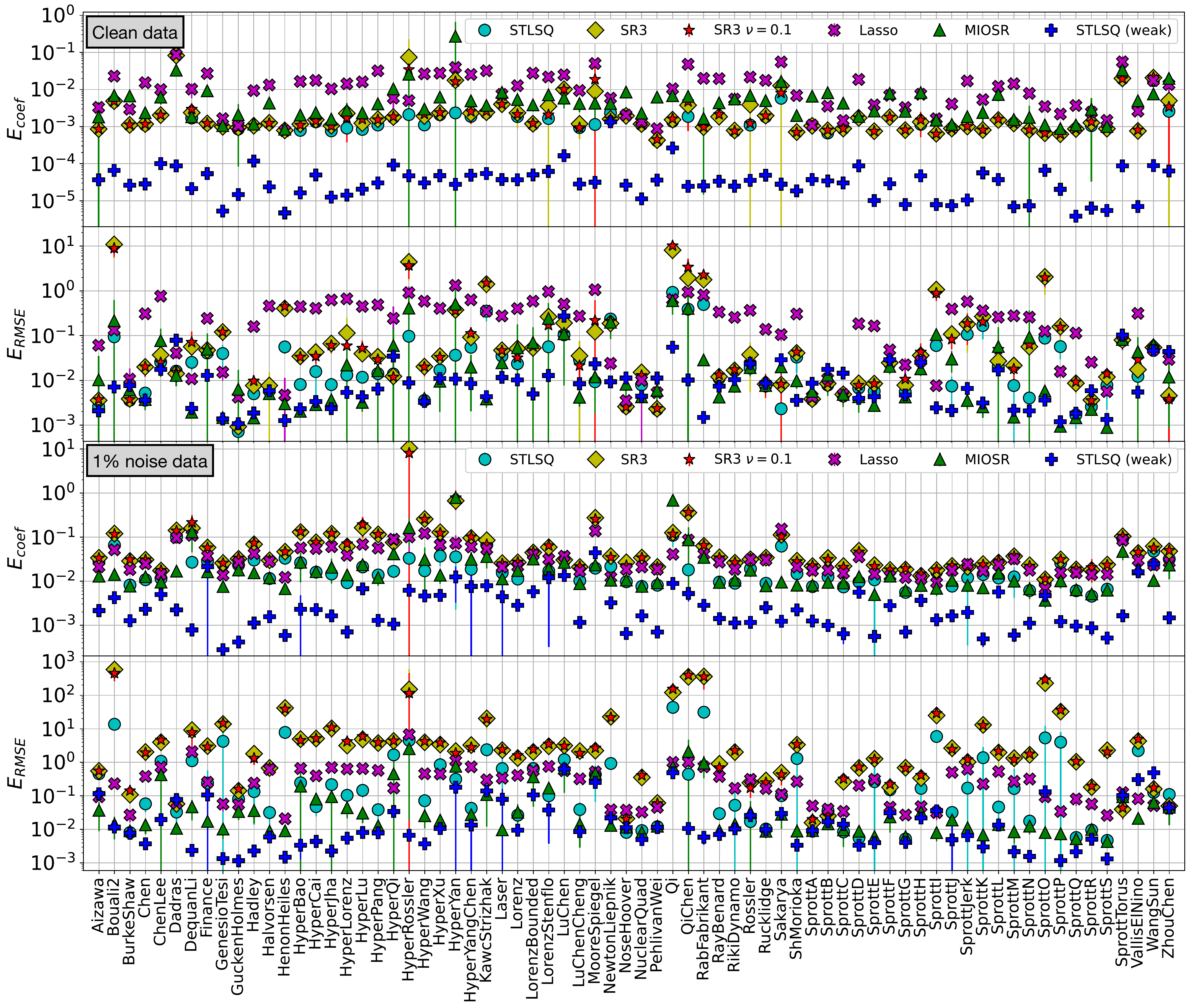}
    \caption{}
         \label{fig:optimizer_comparison}
     \end{subfigure}
    \begin{subfigure}[b]{0.35\textwidth}
    \includegraphics[width=\linewidth]{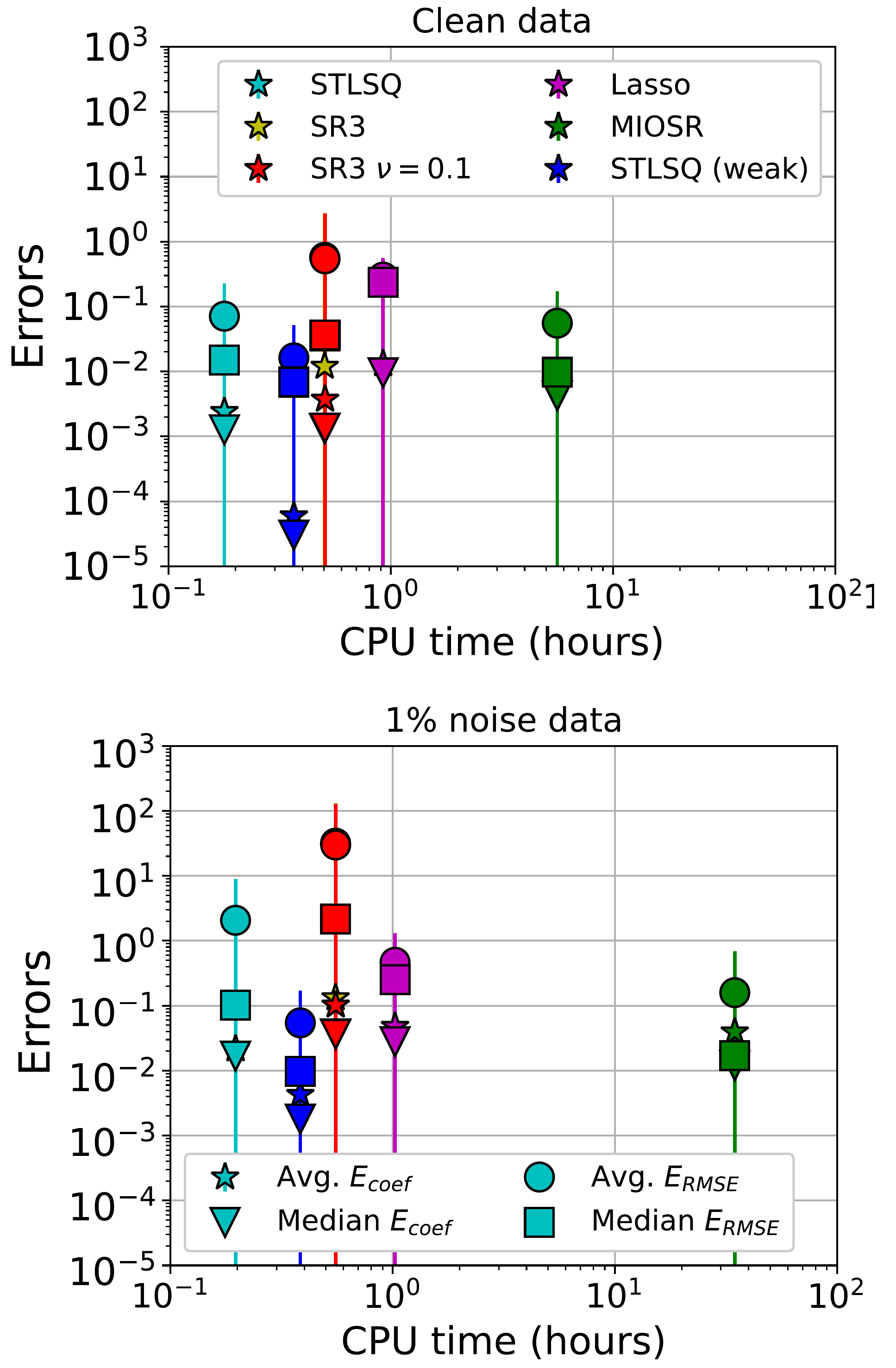}
    \caption{}
    \label{fig:error_summary}
    \end{subfigure}
    \caption{(a) Summary of the average errors for each optimization algorithm, without noise (top two rows) and with $1\%$ noise (bottom two rows), on all 70 chaotic systems. STLSQ and MIOSR generally produce the best models of the traditional optimizers. Lasso has weak performance, and SR3 seems to require additional hyperparameter scanning to avoid substantial errors when noise is present. The weak formulation with STLSQ identifies the coefficients extremely well and demonstrates the substantial advantages of using the weak form. (b) Algorithm average and median performance aggregated over all the dynamical systems. Marker colors indicate different optimizers, and the marker shapes indicate the type of error. All runs were performed serially with an Intel Core i9-7920X 2.90GHz CPU.}
\end{figure*}

\section{Results\label{sec:results}}
We now demonstrate the utility of combining dysts with PySINDy by illustrating a series of benchmark experiments to better understand performance tradeoffs of sparse system identification variants.
Now that we have explained the characteristics of the database, the sparse system identification algorithm, the error metrics, and the dynamical properties of interest, we illustrate the results of our investigations with the standardized, Pareto-optimal SINDy models. We emphasize that the \href{https://github.com/dynamicslab/pysindy/tree/master/examples/16_noise_robustness}{entirety of these results} are reproduced as examples in the PySINDy code and the following investigations use Pareto-optimal identified models that have been chosen by producing the minimal AIC in a hyperparameter scan. 

\subsection{Comparison of SINDy algorithms}
As an illustration of this methodology and database, we compare Pareto-optimal models for a class of SINDy algorithms: STLSQ, Lasso, SR3 (fixing the $\nu$ hyperparameter to $1$ or $0.1$), and MIOSR. Additionally, we use the STLSQ optimizer with the weak form implementation in the PySINDy code in order to compare with the traditional method. Although not shown in this work, the weak formulation can also be used with any of the other PySINDy optimizers with minimal modification. Weak form runs were performed with the other algorithms and similar trends were seen between algorithms, so we have omitted these results. The weak form uses $200$ subdomains, which becomes 200 points to use in the optimization, rather than the 1000 trajectory points used in the traditional optimization. This many subdomains is probably excessive; it was found empirically that, after 100 subdomains, there is essentially no further improvement (in fact just 10 subdomains is much faster to compute and still performs quite well). The MIOSR optimizer is set to use no more than five seconds to solve and prove optimality per state variable. The limit of five seconds is rarely invoked with clean data but more often invoked once the data has $1\%$ added noise. 

\begin{figure}
    \centering
    \includegraphics[width=0.97\linewidth]{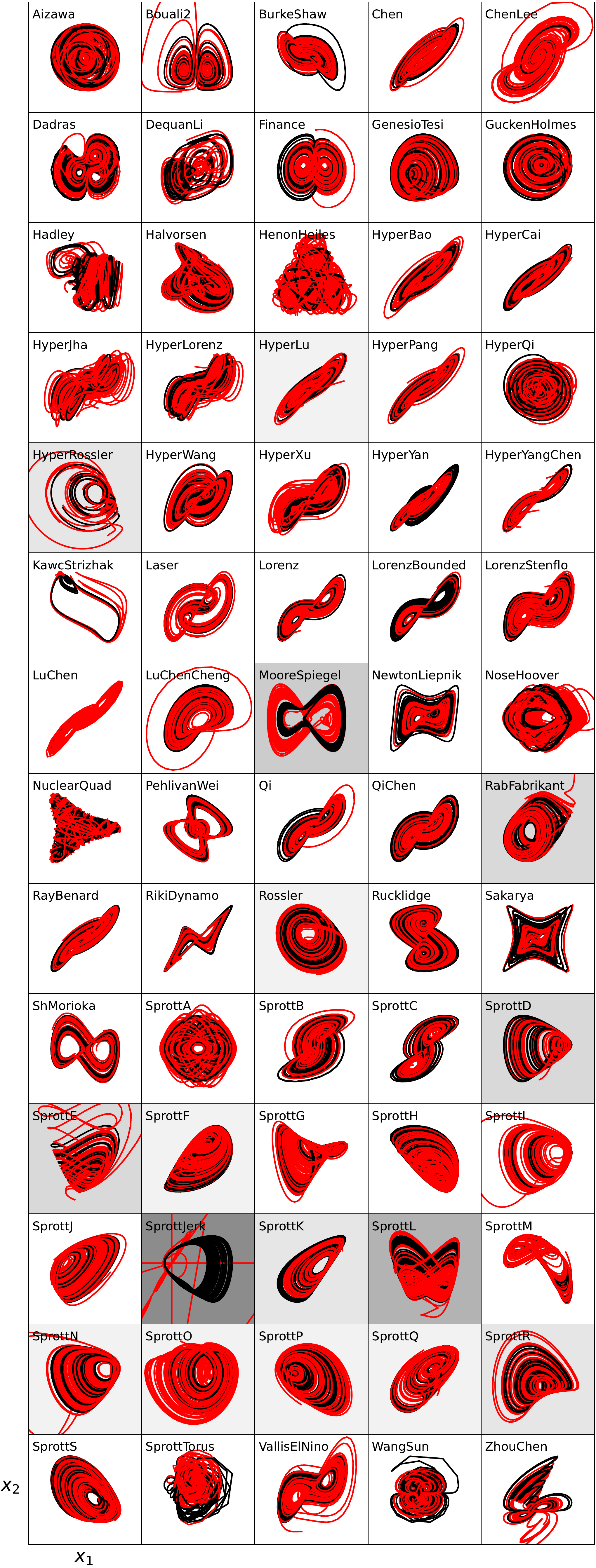}
    \caption{Pareto-optimal SINDy STLSQ model simulations (red) vs true trajectories (black) in the $(x_1, x_2)$ subspace using ten new initial conditions. Greyer backgrounds indicate more unstable trajectories, which are otherwise not pictured.}
    \label{fig:trajectories_zero_noise}
\end{figure}

The results were generated by the process outlined earlier in Section~\ref{sec:Pareto_scan}. 
Figure~\ref{fig:optimizer_comparison} illustrates the Pareto-optimal model performances for each of the SINDy algorithms. Results are shown for both clean data and data with $1\%$ added noise. These results illustrate the known insufficiency of the Lasso for generating high-performance system identification results, although the Lasso degrades less than some of the other optimizers as noise is added. On the clean data, the remainder of the optimizers regularly produce models with $E_\text{coef} < 1\%$ and $E_\text{RMSE} < 10\%$. 

For additional evidence that these are accurate dynamical models, Fig.~\ref{fig:trajectories_zero_noise} illustrates the Pareto-optimal STLSQ model simulations of ten trajectories from new initial conditions starting slightly off each of the strange attractors. An increasingly grey-tinted background indicates an increasing number of simulated trajectories that went unstable in the ODE solver. Overall, the major conclusion is that, without noise, Pareto-optimal SINDy models regularly produce new trajectories that recreate the strange attractor of each chaotic system, meaning that the dynamical features such as the Lyapunov spectrum can be accurately computed. 

\subsection{Weak formulation results}\label{sec:weak_form}
The weak or integral form of SINDy~\cite{Schaeffer2017pre,gurevich2019robust,messenger2021weakpde} subdivides the time span of the trajectories and considers integrals of the ODEs over these subdomains. This method then avoids the calculation of derivatives on noisy data and subsequent noise amplification by integrating by parts when appropriate. 

The results in Figure~\ref{fig:optimizer_comparison} indicate that the weak formulation performs extraordinarily well in producing the correct equation coefficients. The weak form $E_\text{RMSE}$ results appear slightly less impressive for a rather technical reason. We are computing the pointwise $E_\text{RMSE}$ defined in Eq.~\eqref{eq:prediction_error} in order to compare against the other optimizers. Unlike the other optimizers, $E_\text{RMSE}$ (of the training data) is not directly minimized in the weak formulation of the problem; the weak form solves its optimization problem by minimizing an error metric that is integrated over the subdomains. Despite not directly minimizing these quantities, the weak form still generally obtains the smallest values of $E_\text{RMSE}$. 

These results provide compelling evidence for the considerable strength of the weak formulation. Interestingly, the typical justification for the high performance of weak formulations of sparse regression is that it allows the user to avoid amplifying noise when computing derivatives of the data. However, we find significant improvements with the weak formulation even when there is no added noise -- only the intrinsic and unavoidable ``noise'' of finite sampling rates. Even further, the weak formulation results are not truly Pareto-optimal; the threshold value for STLSQ is still scanned but the additional hyperparameters available in the weak form are just fixed to a set of plausible values. Combined, we consider these results persuasive for using the weak formulation whenever possible. For flexible usage, the weak formulation is entirely integrated into the feature library functionality of the PySINDy code, and therefore can be used with any of the current or future optimizers, or with any generic library terms of interest. 

\subsection{Greedy vs exact algorithms}\label{sec:greedy}
MIOSR is an exact mixed-integer algorithm that can often solve the $l_0$-regularized (i.e. nonconvex) SINDy problem to optimality. In contrast, the STLSQ algorithm solves this problem greedily, meaning it chooses the best model on a subset of the coefficients at each iteration, and it cannot recover from thresholding mistakes that occur in earlier iterations.

On the noisy data, the MIOSR algorithm seems to outperform the others, while with clean data all of the non-weak algorithms perform similarly well except for Lasso. With clean data, the SR3 algorithms very well. Some variation is seen with SR3 between the $\nu = 1$ and $\nu = 0.1$ cases but a clear trend is not distinguishable. The SR3 errors become significantly larger as noise is added and these results seem to indicate that hyperparameter scans over $\nu$ are required for high performance when noise is present. All of the optimizer performances degrade somewhat when noise is added. These overall takeaways are summarized further in Figure~\ref{fig:error_summary}.

At $1\%$ noise, STLSQ and MIOSR show performance advantages over the other optimizers. As measured by the average coefficient errors, MIOSR leads the group. STLSQ at $1\%$ noise exhibits a small number of very large RMSE errors and this tends to reduce the average performance. This is consistent with the explanation that greedy methods can increasingly make catastrophic mistakes as the noise levels increase. This trend can also been seen to an extent in the weak formulation results since the optimization is still being performed greedily with STLSQ.

One takeaway is that the MIOSR algorithm performs well, as we might expect for an exact mixed-integer optimization. However, Fig.~\ref{fig:error_summary} shows that, at larger levels of noise, the MIOSR algorithm requires much more computing time than the other algorithms. This is true despite putting a time limit on the computation, but it is reassuring that strong performance is seen even if the optimizer does not finish finding a solution to optimality. Indeed, as is pointed out in detail in Bertsimas et al.~\cite{bertsimas2022learning},  MIOSR runs slower when the regression is harder, e.g. in the noisy setting, but the computational efficiency scales very well with the amount of data and additionally sees substantial speedups with parallelization or better CPUs than used in this work. However, our results appear to show MIOSR does not significantly dominate the other algorithms in our chosen performance metrics. This difference with the results in Bertsimas et al.~\cite{bertsimas2022learning} may be because their performance metric focus is primarily on the true positive rate of the model coefficients, and their MIOSR results are most impressive in the low-data limit. 

In fact, these results also highlight the strong performance of the STLSQ algorithm; STLSQ appears relatively robust to noise, very fast to compute (regardless of the noise level), and, for the most part, generates models of comparable performance to MIOSR.
Even at $1\%$ noise, the errors that accumulate from the greedy nature of STLSQ appear tolerable. These optimizers were not tested at larger values of noise. It is well known that weak or integral formulations of the optimization problem are required for retaining high performance in the high-noise setting~\cite{Schaeffer2017pre,reinbold2020using,fasel2022ensemble}. This fact is clearly reinforced by the impressive weak form performance in Fig.~\ref{fig:optimizer_comparison}.

\subsection{Dynamical properties}\label{sec:results_zero_noise}
A pressing question for practitioners of system identification is whether certain types of dynamical systems are more difficult to identify than others. For instance, some methods are specifically designed to search for Hamiltonian systems~\cite{bhat2019learning,chu2020discovering,bertalan2019learning}. Another example is that traditional RNNs should be modified for modeling chaotic systems in order to avoid unbounded gradients~\cite{mikhaeil2021difficulty}. 

\begin{figure*}
    \centering
    \includegraphics[width=0.9\linewidth]{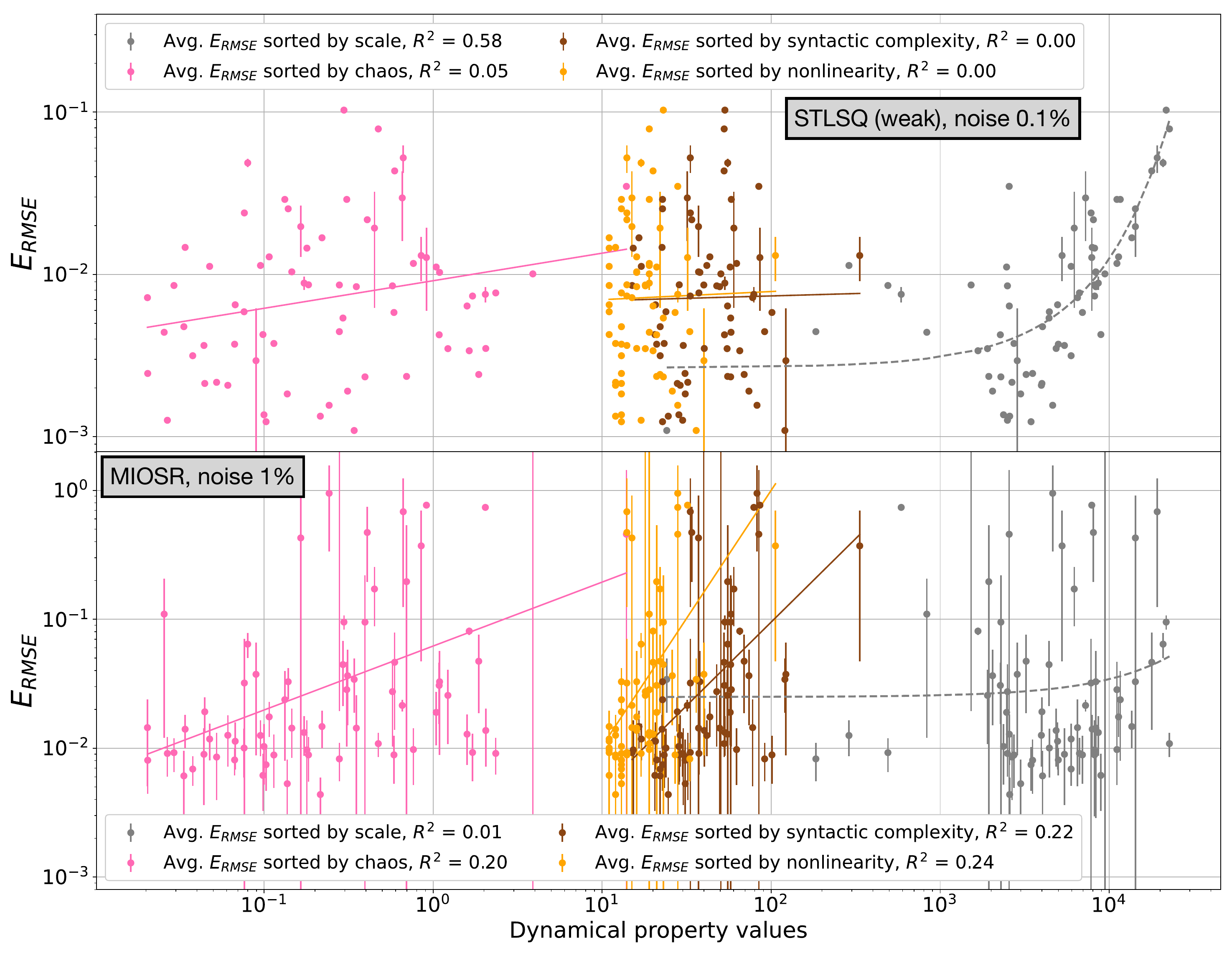}
    \caption{Two illustrations of the $E_\text{RMSE}$ results for the Pareto-optimal STLSQ optimizer (weak form) on data with $0.1\%$ noise and for the MIOSR optimizer at $1\%$ noise. The STLSQ result suggests that $E_\text{RMSE}$ grows quickly with the scale separation. The MIOSR algorithm produces weak log-log trends with the other three dynamical properties tested in this work, although these trends strengthen as noise increases.}
    \label{fig:dynamical_properties_summary}
\end{figure*}

\begin{figure*}
    \centering
    \includegraphics[width=0.7\linewidth]{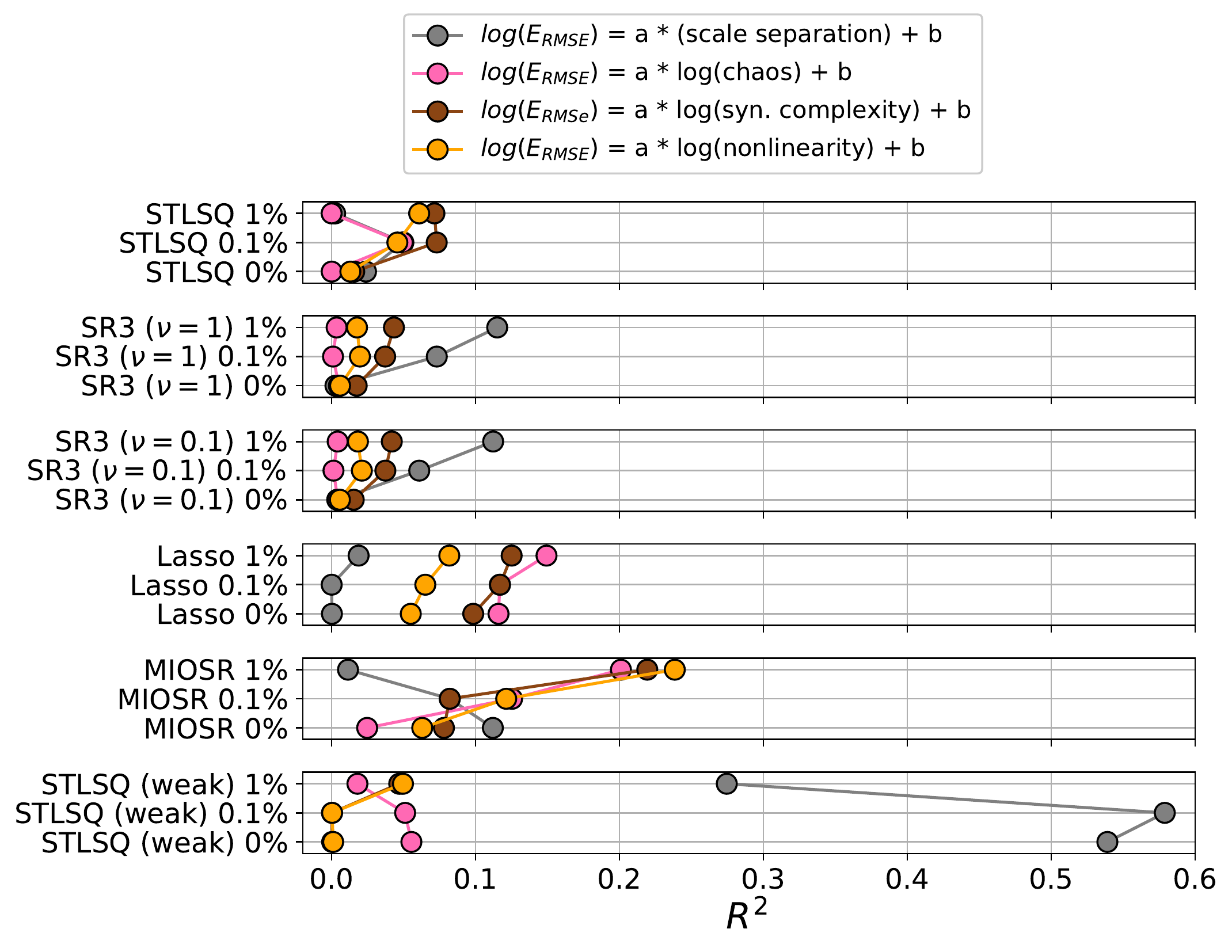}
    \caption{Summary of the  weak $R^2$ coefficients of determination found between each optimizer's Pareto-optimal $E_\text{RMSE}$ and the dynamical properties, choosing the best fit from linear, log-linear, and log-log regressions. Some differing trends can be observed for each optimizer. Even weaker correlations are found between $E_\text{coef}$ and the dynamical properties (not pictured).}
    \label{fig:dynamical_properties_total}
\end{figure*}

We now present an investigation into the importance of the dynamical properties with respect to the performance of the various SINDy optimizers. The required data and computations for this analysis have already been generated in the previous section.
For each chaotic system in our dataset, the errors of the Pareto-optimal models are sorted by the dynamical property in question, and some simple data fits are attempted: linear, log-linear, and log-log. 
Linear fits between $\log(E_{RMSE})$ and the scale separation seemed to produce the largest $R^2$ coefficient of determination for all the optimizers, and similarly for log-log fits between $\log(E_{RMSE})$ and the logarithm of the remaining dynamical properties. 
We make no claims that these simple fits to the data best capture the correlations, but rather they are used to roughly quantify if there are correlations between increasing dynamical property values and worsening system identification performance. 

Visual inspection of the different dynamical property fits for the weak form STLSQ optimizer at $0.1\%$ noise in Fig.~\ref{fig:dynamical_properties_summary} does seem to indicate that, on average, the optimizer performance rapidly drops as the scale separation increases. Similarly, visual inspection of the different dynamical property fits for the MIOSR optimizer at $1\%$ noise in Fig.~\ref{fig:dynamical_properties_summary} indicates that log-log fits to the data weakly capture the optimizer performance for all of the dynamical properties except the scale separation, which shows no clear trend. 

All of the best $R^2$ coefficients of determination are illustrated in Fig.~\ref{fig:dynamical_properties_total} for each optimizer, each dynamical property, and for noise levels of 0, 0.1, and $1\%$. Most of the $R^2$ values are below 0.25, indicating that the dominant behavior is that the optimizers produce results that are approximately independent of the dynamical properties. Nonetheless, there appear to be some intriguing trends; for instance, the weak formulation $E_\text{RMSE}$ shows substantial correlation with the scale separation. MIOSR seems to increasingly correlate with the dynamical properties (except that of scale separation) at larger values of noise.

These results suggest that there are some relatively persistent but quite weak correlations with the underlying dynamical properties of the equations. Moreover, which correlations dominate appears to depend on the level of noise in the data and the choice of optimization algorithm.

Our large experiments are a strong affirmation of the ``Task 4'' result in Gilpin indicating that SINDy model results were independent of the level of chaos~\cite{gilpin2021chaos}. Our metric for the degree of chaos in the system is a rather coarse measure, but similar results were found in both Gilpin and the present work when alternative metrics for chaos were used. The lack of correlation with the description length is an interesting and potentially counter-intuitive discovery. To be more specific, if the dynamical terms live in the subspace of the feature library, the quality of Pareto-optimal models generated by sparse regression onto data seems to be approximately uncorrelated with the length or complexity of the underlying dynamical equations. This could be seen as a positive result. If adequate functional terms are available to describe the underlying dynamics, and the data is high-quality, the complexity of the equation factors out. This motivates large and expansive libraries, although this will generally increase the condition number of the feature library and therefore require additional regularization and shift the Pareto-front.

At $1\%$ noise levels, the quality of the Pareto-optimal models generally drops; for instance, a typical result with STLSQ produces only a few models with less than $10\%$ $E_\text{RMSE}$ and eight of the models are not able to produce any new trajectories (starting from slightly off the attractor) that either stay bounded or remotely match the qualitative behavior of the strange attractor.

\section{Conclusions\label{sec:conclusion}}
The methodology and analysis in the present work has shown that sparse system identification can be used to identify the dynamics of 70 polynomial chaotic systems, reproducing the strange attractors and therefore reproducing the fundamental dynamics. We extensively used sub-sampling to generate large ensembles of models, utilized the AIC to identify ensembles at the optimal hyperparameter values, and relied on the computational speed of sparse linear regression to compute over one million dynamical models in just two days worth of CPU time (mostly expended during the MIOSR algorithm).

This methodology was used to benchmark a variety of different SINDy optimizers, finding that: STLSQ $\&$ MIOSR produce the best performance, Lasso is relatively robust to noise but otherwise lower performance, and large-scale fitting with SR3 requires additional hyperparameter scanning for high performance. As significant noise is added to the data, STLSQ retains its fast computational speed but starts making ``greedy'' mistakes that cannot be recovered from, while conversely, MIOSR slows down considerably but can retain its strong performance for most systems. 
Lastly, we find very persuasive evidence that the weak formulation provides significant performance improvements across the database, even in the zero-added-noise setting. The weak form can be used with any of the PySINDy optimizers and we recommend using the weak formulation for almost all cases. Similarly, sub-sampling the data to create an ensemble of SINDy models~\cite{fasel2022ensemble} is highly recommended, especially in the presence of noise.

We have also presented one of the first large-scale and quantitative analyses into how sparse regression depends on the dynamical properties of the underlying governing equations. Overall, we found very weak correlations between the dynamical properties and the performance of the Pareto-optimal SINDy models, with some noteworthy deviations, providing a foundation for more sophisticated investigations in future work.

There are many opportunities for future work that builds on the database and analysis presented here. Examples include further improvements to recent and important work addressing partial measurements~\cite{bakarji2022discovering,ouala2023bounded}, the low-data high-noise regime~\cite{fasel2022ensemble,bertsimas2022learning}, extended feature libraries (which can already be built in PySINDy), and algorithm comparisons with convex constraints. For instance, most of the optimizers, STLSQ, SR3, Lasso, and MIOSR, can in principle accommodate general convex constraints. However, future PySINDy work is required to implement these options. Currently, constraints with STLSQ~\cite{loiseau2018constrained} and Lasso are not supported, while SR3 constraints in PySINDy (except linear equalities) rely on a CVXPY~\cite{diamond2016cvxpy} backend which significantly slows down the algorithm. See e.g. Kaptanoglu et al.~\cite{kaptanoglu2022permanent} for a much faster SR3 algorithm with quadratic inequality constraints. 

There are also numerous other dynamical metrics, sparse system identification algorithms, subsampling methods, types of dynamical systems (i.e. not just those defined by polynomial dynamics), and other parameters to systematically test. Although it is beyond the scope of this work, the dysts database facilitates another very interesting line of inquiry -- building data-driven models for predicting chaotic bifurcations in global dynamical behavior, e.g. between exhibiting a strange attractor and a stable periodic solution. Additional future work could perform a quantitative analysis of how invariant manifolds affect the performance and usefulness of system identification; this topic is discussed in Appendix~\ref{sec:appendix_dynamical_properties}.
Lastly, many dynamical or engineering-focused metrics, such as the description length of the equations, can be directly minimized during the optimization and this may change the qualitative conclusions found in this work with the baseline SINDy optimization problem, especially since the present work found that the correlations are weak between model performance and the metrics tested. 

\section{Acknowledgements}
AAK would like to acknowledge useful conversations with William Gilpin, Chris Hansen, and Nathan Kutz. ZGN is a Washington Research Foundation Postdoctoral Fellow. The authors acknowledge support from the National Science Foundation AI Institute in Dynamic Systems under grant number 2112085.

\appendix 

\section{How to use this benchmark}\label{sec:appendix_benchmark_use}
The goal of the present work is to provide a systematic approach to benchmark new and existing innovations in the field of sparse system identification. In order to do so, we have endeavored to provide simple, intuitive scripts, built on the open-source PySINDy code. 
A script for performing the type of Pareto-optimal scans in this work can be found in one of the PySINDy code \href{https://github.com/dynamicslab/pysindy/tree/master/examples/16_noise_robustness}{examples}. It contains options for the user to specify (1) the data (including the number of points, sampling rates, etc. of the training and testing trajectories), (2) the optimization algorithm, and (3) the feature library (including whether to normalize the feature library, how to subsample the data for making model ensembles, and the number of models to generate for making statistical conclusions). For the dysts database, it shows additionally how to aggregate all the equations and their associated dynamical properties. The entirety of the results in this work can be reproduced by running the run$\_$all.py script in the same directory.

The PySINDy code is regularly updated with new methods from the literature and now contains many SINDy variations and advanced functionality. 
New algorithms, feature libraries, subsampling strategies, and more can be readily added into the PySINDy code and then immediately used with the scripts mentioned above. Much of the backend for forming general SINDy libraries and new optimizers is already pre-made in the code and should significantly reduce the overhead designing and testing new algorithms.

\section{Small model errors and associated instability}\label{sec:appendix_stability}
True sparse regression is a nonconvex problem because of the $l_0$ loss term in the objective function. The $l_1$ loss term is often used in place of the $l_0$ since, under certain conditions, it will produce the same solution with high probability~\cite{wainwright2009sharp}. When these conditions are not met, the Lasso is known to weakly choose irrelevant features~\cite{bertsimas_review}. Furthermore, even the optimally-solved $l_0$-regularized problem can fail to predict exactly the correct features when the data is sufficiently corrupted by noise. In the context of system identification, small model errors can generate significant issues.

For instance, very small errors in the model identification can cause finite-time and unphysical instability when generating new trajectories from the identified models (although this is not an issue if the user merely seeks to find an approximation of the equations rather than to \textit{use} them for generating new trajectories). In the case of fourth order polynomials, we can use the following heuristic. We would expect that small coefficient errors on the third and fourth order terms will almost inevitably produce instability on \textit{some} trajectory since these terms will be negligible near the origin (or near the attractor) but an initial condition can always be chosen far enough from the origin in order to make high order terms dominant. Only in the constrained~\cite{loiseau2018constrained} and stability-promoted~\cite{kaptanoglu2021promoting} settings should we expect that most (or all) trajectories generated from the identified models are stable. Finite-time blow up of the solution obviously prevents a reasonable comparison of the RMSE error of $\bm X$ (rather than $\dot{\bm X}$) and additionally prevents calculation of the Lyapunov spectrum and other dynamical properties of interest.

There are two half-remedies: only test the model with trajectories with initial conditions reasonably close to the strange attractor(s) and avoid rescaling the SINDy matrix. The first remedy comes from the intuition that small coefficient errors on high-order terms will only become dominant in regions of the state space sufficiently far from the origin. The latter remedy is also intuitive if we consider the STLSQ algorithm. It has been previously noted, e.g. in Delahunt et al.~\cite{delahunt2022toolkit}, that if $x(t)\sim 10$ on the strange attractor, the two terms $x(t)$ and $0.1x^2(t)$ will contribute similarly to the time evolution, yet the latter term will be the first term to be truncated by an algorithm that relies on some version of hard-thresholding the smallest coefficients in the equations. Rather than seeing this as a problem of the algorithm, we can view this instead as an algorithmic bias towards describing each system with low order polynomial terms, which we expect should contribute indirectly to boundedness. 
Unfortunately, the opposite is true if $x(t) \sim 0.1$, so it can be useful to rescale  the different dynamical systems so that this thresholding mismatch is all in the same direction for the entire database. Although not explored in this work, minimizing an objective function related to the $N_t$-step SINDy model prediction as in e.g. Kaheman et al.~\cite{kaheman2022automatic} or Bakarji et al.~\cite{bakarji2022discovering}, rather than the traditional single time-step prediction, should generally improve the model stability but at the expense of nonconvexity (it is no longer sparse \textit{linear} regression).

Lastly, even when the identified models are essentially perfect and reproduce the strange attractor(s), reporting the $\bm X$ RMSE error for chaotic systems can often be profoundly misleading, since inevitably there will be (initially exponentially large and then bounded) errors in the trajectory RMSEs from tiny errors in the integration, identified coefficients, and so on. We found it much more convincing to simply show visually that, with new initial conditions, the identified models can approximately reproduce the correct attractors, and therefore accurate calculations of the Lyapunov spectrum and other dynamical properties are possible. 

\section{Dynamical property definitions}\label{sec:appendix_dynamical_properties}
In this section we provide precise definitions for the dynamical properties computed and analyzed in the present work. 
\begin{itemize}
\item[1.] \textit{The amount of chaos} can be measured in numerous ways~\cite{gilpin2021chaos}, and a common choice is the largest Lyapunov exponent. The largest Lyapunov exponent measures the rate at which trajectories from two initially infinitesimally-close points exponentially separate as time evolves~\cite{sommerer1993particles}. This exponent can have very large variation over a trajectory, so it is usually considered as a global average. There is some evidence~\cite{gilpin2021chaos} that deep symbolic regression~\cite{petersen2019deep} and recurrent neural networks see degraded performance at higher levels of chaos~\cite{mikhaeil2021difficulty}.
\item [2.] \textit{The degree of scale separation} could be defined through a ratio of the largest timescale to smallest timescale in the underlying equations. However, these timescales are not typically well-defined even for polynomially nonlinear systems, so we could instead use the ratio of the largest to smallest Lyapunov exponents. As before, the Lyapunov spectrum can often vary tremendously over a trajectory and therefore this is quite a coarse measure of scale separation. For a much improved metric to measure scale separation, we use the dominant timescale divided by the smallest significant timescale as defined in Gilpin~\cite{gilpin2021chaos}. 
To see why this measure of scale separation might be dynamically relevant to symbolic regression, consider the following simplified model
\begin{align}\label{eq:timescale_eqns}
    \dot{x}_1 &= f(x_1, x_2), \\ \notag 
    \dot{x}_2 &= \frac{1}{\epsilon}g(x_1, x_2),
\end{align}
for $\epsilon \ll 1$ and assuming the existence of a strange attractor, as in the case for our dataset. Following the standard derivation for invariant manifolds~\cite{pavliotis2008multiscale}, on long time intervals (compared to the $\epsilon$ timescale) $x_2(t)$ quickly finds an equilibrium  parametrized as $x_2(t) \sim \eta(x_1(t))$, and is functionally controlled by the $x_1(t)$ evolution. The conclusion is the existence of an invariant manifold defined by 
\begin{align}\label{eq:invariant_manifold}
    \dot{x}_1 &= f(x_1, \eta(x_1)), \\ \notag
    \dot{x}_2 &= \partial_{x_1}\eta(x_1)\dot{x}_1 = \partial_{x_1}\eta(x_1)f(x_1, \eta(x_1)),
\end{align}
which is a manifold in the $(x_1, x_2)$ state space. If few data points are available for the period when $t \sim \epsilon$ (or, equivalently, if the initial value $x_2(0)$ starts close to the equilibrium or strange attractor), we cannot reasonably expect that Eqs.~\eqref{eq:timescale_eqns} and~\eqref{eq:invariant_manifold} can be distinguished. In other words, we might expect that for the best symbolic regression results (i.e. results that correctly capture the dynamical terms necessary to converge to the strange attractor from initial conditions starting substantially off the attractor) we need to use initial conditions far from the equilibrium, and resolved at the smallest time scale. This intuition is given some additional practical and theoretical justification in Bucci et al.~\cite{bucci2022}. It is shown explicitly that trajectories from linearly unstable fixed points contain less entropy, i.e. less information, than trajectories starting on the attractors, since near the fixed points only the linear terms are active and information about the nonlinearities cannot be gleaned. Similarly, on the attractor there is a balance of terms that reduces the information available about the equation terms, relative to some generic trajectory in the phase space. Some chaotic systems can have effective dimensions on the attractor that are significantly smaller than the state space dimension. 

There is another important note to make here. We know that Eqs.~\eqref{eq:timescale_eqns} are discoverable with the pre-defined, fourth-order polynomial feature library used for symbolic regression. This is not guaranteed for the equations defining the invariant manifold, and thus we may still find the correct equations simply because the more specific invariant manifold equations are not accessible in the regression. 
Proceeding down this line of analysis is complex and further complicated by the fact that often  polynomial terms are very close to linearly dependent and therefore multiple sparse models may fit the data well~\cite{delahunt2022toolkit}. 

In order to sample trajectories while they are off the attractor, i.e. when $t \sim \epsilon$, we need to sample each period such that $\sim \epsilon^{-1}$ points are generated, greatly increasing the required sampling resolution.  Alternatively, a more sophisticated weighting strategy may be employed. There has been some recent symbolic regression work~\cite{bramburger2020sparse,cenedese2022data,szalai2022data,axaas2022fast} connecting system identification and invariant manifolds. An interesting future investigation could attempt to identify the map $\eta(x_1(t))$ by searching for Eq.~\eqref{eq:invariant_manifold} with the constraint that library terms for $\dot{x}_2$ must be proportional to $\dot{x}_1$. These types of constraints are already implemented in PySINDy.

\item[3.] \textit{The syntactic complexity} of the model has been considered in earlier work as useful for both direct optimization and as a post-fit metric~\cite{udrescu2020ai2}. There are many possible definitions of syntactic complexity, but we adopt the approximate description-length metric~\cite{grunwald2005advances} since the description length has been used in a number of contexts for direct optimization and has a natural interpretation as approximating the number of bits to describe each object in the equations.  

\item[4.] \textit{The amount of nonlinearity} is also not typically well-defined. The highest degree of nonlinearity can be defined as the largest polynomial appearing in the equations governing each system, and previous work has used this definition for a complexity measure in sparse regression~\cite{vladislavleva2008order}. However, some systems have many quadratic nonlinear terms, and we might consider such systems very nonlinear despite the lack of higher-order polynomials. As a basic metric to trade-off between these considerations, we consider a weight vector $w_i = [1, 2, 3, 4, 5]$ and for each system simply report
\begin{align}
    N_\text{nonlinearity} = \sum_{j=1}^d\sum_{i=1}^5 w_i \|x^{i-1} \text{ terms in equation j}\|_0.
\end{align}
In other words, we sum all of the terms present in the coupled set of ODEs, weighted by the polynomial degree of each term. This includes the linear and constant terms, so it is a mix between the degree of nonlinearity and the number of equation terms. 
Because this definition takes into account the number of terms, there is presumably some overlap with the information captured by the description length metric. 
\end{itemize}

We make no claims that any of these metrics are the most compelling choice for capturing the particular dynamical information, but each metric seems to capture an important aspect. A substantial discussion on some of these points can be found in Murdoch et al.~\cite{murdoch2019definitions}. Future work could investigate more sophisticated metrics such as the stiffness or integrability of the identified ODE systems, or investigating the corresponding stability features of the models. For generic nonlinear ODE systems, investigating stability seems intractable, but there is a significant volume of literature for finding Lyapunov functions for polynomially nonlinear systems~\cite{tesi1996stability,ahmadi2013complexity}. Optimizing stiffness~\cite{dikeman2022stiffness} and stability would be additionally useful for finding models that can be used for applying chaos control~\cite{ott1990controlling}.

 \bibliography{attractor.bib}

\providecommand{\noopsort}[1]{}\providecommand{\singleletter}[1]{#1}%
\begin{thebibliography}{150}%
\makeatletter
\providecommand \@ifxundefined [1]{%
 \@ifx{#1\undefined}
}%
\providecommand \@ifnum [1]{%
 \ifnum #1\expandafter \@firstoftwo
 \else \expandafter \@secondoftwo
 \fi
}%
\providecommand \@ifx [1]{%
 \ifx #1\expandafter \@firstoftwo
 \else \expandafter \@secondoftwo
 \fi
}%
\providecommand \natexlab [1]{#1}%
\providecommand \enquote  [1]{``#1''}%
\providecommand \bibnamefont  [1]{#1}%
\providecommand \bibfnamefont [1]{#1}%
\providecommand \citenamefont [1]{#1}%
\providecommand \href@noop [0]{\@secondoftwo}%
\providecommand \href [0]{\begingroup \@sanitize@url \@href}%
\providecommand \@href[1]{\@@startlink{#1}\@@href}%
\providecommand \@@href[1]{\endgroup#1\@@endlink}%
\providecommand \@sanitize@url [0]{\catcode `\\12\catcode `\$12\catcode
  `\&12\catcode `\#12\catcode `\^12\catcode `\_12\catcode `\%12\relax}%
\providecommand \@@startlink[1]{}%
\providecommand \@@endlink[0]{}%
\providecommand \url  [0]{\begingroup\@sanitize@url \@url }%
\providecommand \@url [1]{\endgroup\@href {#1}{\urlprefix }}%
\providecommand \urlprefix  [0]{URL }%
\providecommand \Eprint [0]{\href }%
\providecommand \doibase [0]{https://doi.org/}%
\providecommand \selectlanguage [0]{\@gobble}%
\providecommand \bibinfo  [0]{\@secondoftwo}%
\providecommand \bibfield  [0]{\@secondoftwo}%
\providecommand \translation [1]{[#1]}%
\providecommand \BibitemOpen [0]{}%
\providecommand \bibitemStop [0]{}%
\providecommand \bibitemNoStop [0]{.\EOS\space}%
\providecommand \EOS [0]{\spacefactor3000\relax}%
\providecommand \BibitemShut  [1]{\csname bibitem#1\endcsname}%
\let\auto@bib@innerbib\@empty
\bibitem [{\citenamefont {Gilpin}(2021)}]{gilpin2021chaos}%
  \BibitemOpen
  \bibfield  {author} {\bibinfo {author} {\bibfnamefont {W.}~\bibnamefont
  {Gilpin}},\ }\bibfield  {title} {\bibinfo {title} {Chaos as an interpretable
  benchmark for forecasting and data-driven modelling},\ }\href@noop {}
  {\bibfield  {journal} {\bibinfo  {journal} {Advances in Neural Information
  Processing Systems (NeurIPS), arXiv:2110.05266}\ } (\bibinfo {year}
  {2021})}\BibitemShut {NoStop}%
\bibitem [{\citenamefont {Schmid}(2010)}]{schmid2010dynamic}%
  \BibitemOpen
  \bibfield  {author} {\bibinfo {author} {\bibfnamefont {P.~J.}\ \bibnamefont
  {Schmid}},\ }\bibfield  {title} {\bibinfo {title} {Dynamic mode decomposition
  of numerical and experimental data},\ }\href@noop {} {\bibfield  {journal}
  {\bibinfo  {journal} {Journal of Fluid Mechanics}\ }\textbf {\bibinfo
  {volume} {656}},\ \bibinfo {pages} {5} (\bibinfo {year} {2010})}\BibitemShut
  {NoStop}%
\bibitem [{\citenamefont {Rowley}\ \emph {et~al.}(2009)\citenamefont {Rowley},
  \citenamefont {Mezi\'c}, \citenamefont {Bagheri}, \citenamefont {Schlatter},\
  and\ \citenamefont {Henningson}}]{Rowley2009jfm}%
  \BibitemOpen
  \bibfield  {author} {\bibinfo {author} {\bibfnamefont {C.~W.}\ \bibnamefont
  {Rowley}}, \bibinfo {author} {\bibfnamefont {I.}~\bibnamefont {Mezi\'c}},
  \bibinfo {author} {\bibfnamefont {S.}~\bibnamefont {Bagheri}}, \bibinfo
  {author} {\bibfnamefont {P.}~\bibnamefont {Schlatter}},\ and\ \bibinfo
  {author} {\bibfnamefont {D.}~\bibnamefont {Henningson}},\ }\bibfield  {title}
  {\bibinfo {title} {Spectral analysis of nonlinear flows},\ }\href@noop {}
  {\bibfield  {journal} {\bibinfo  {journal} {Journal of fluid mechanics}\
  }\textbf {\bibinfo {volume} {641}},\ \bibinfo {pages} {115} (\bibinfo {year}
  {2009})}\BibitemShut {NoStop}%
\bibitem [{\citenamefont {Kutz}\ \emph {et~al.}(2016)\citenamefont {Kutz},
  \citenamefont {Brunton}, \citenamefont {Brunton},\ and\ \citenamefont
  {Proctor}}]{Kutz2016book}%
  \BibitemOpen
  \bibfield  {author} {\bibinfo {author} {\bibfnamefont {J.~N.}\ \bibnamefont
  {Kutz}}, \bibinfo {author} {\bibfnamefont {S.~L.}\ \bibnamefont {Brunton}},
  \bibinfo {author} {\bibfnamefont {B.~W.}\ \bibnamefont {Brunton}},\ and\
  \bibinfo {author} {\bibfnamefont {J.~L.}\ \bibnamefont {Proctor}},\
  }\href@noop {} {\emph {\bibinfo {title} {Dynamic Mode Decomposition:
  Data-Driven Modeling of Complex Systems}}}\ (\bibinfo  {publisher} {SIAM},\
  \bibinfo {year} {2016})\BibitemShut {NoStop}%
\bibitem [{\citenamefont {Mezi{\'c}}(2005)}]{Mezic2005nd}%
  \BibitemOpen
  \bibfield  {author} {\bibinfo {author} {\bibfnamefont {I.}~\bibnamefont
  {Mezi{\'c}}},\ }\bibfield  {title} {\bibinfo {title} {Spectral properties of
  dynamical systems, model reduction and decompositions},\ }\href@noop {}
  {\bibfield  {journal} {\bibinfo  {journal} {Nonlinear Dynamics}\ }\textbf
  {\bibinfo {volume} {41}},\ \bibinfo {pages} {309} (\bibinfo {year}
  {2005})}\BibitemShut {NoStop}%
\bibitem [{\citenamefont {Brunton}\ \emph {et~al.}(2022)\citenamefont
  {Brunton}, \citenamefont {Budišić}, \citenamefont {Kaiser},\ and\
  \citenamefont {Kutz}}]{Brunton2021koopman}%
  \BibitemOpen
  \bibfield  {author} {\bibinfo {author} {\bibfnamefont {S.~L.}\ \bibnamefont
  {Brunton}}, \bibinfo {author} {\bibfnamefont {M.}~\bibnamefont {Budišić}},
  \bibinfo {author} {\bibfnamefont {E.}~\bibnamefont {Kaiser}},\ and\ \bibinfo
  {author} {\bibfnamefont {J.~N.}\ \bibnamefont {Kutz}},\ }\bibfield  {title}
  {\bibinfo {title} {Modern {K}oopman theory for dynamical systems},\
  }\href@noop {} {\bibfield  {journal} {\bibinfo  {journal} {SIAM Review}\
  }\textbf {\bibinfo {volume} {64}},\ \bibinfo {pages} {229} (\bibinfo {year}
  {2022})}\BibitemShut {NoStop}%
\bibitem [{\citenamefont {Billings}(2013)}]{Billings2013book}%
  \BibitemOpen
  \bibfield  {author} {\bibinfo {author} {\bibfnamefont {S.~A.}\ \bibnamefont
  {Billings}},\ }\href@noop {} {\emph {\bibinfo {title} {Nonlinear system
  identification: NARMAX methods in the time, frequency, and spatio-temporal
  domains}}}\ (\bibinfo  {publisher} {John Wiley \& Sons},\ \bibinfo {year}
  {2013})\BibitemShut {NoStop}%
\bibitem [{\citenamefont {Pathak}\ \emph {et~al.}(2018)\citenamefont {Pathak},
  \citenamefont {Hunt}, \citenamefont {Girvan}, \citenamefont {Lu},\ and\
  \citenamefont {Ott}}]{pathak2018model}%
  \BibitemOpen
  \bibfield  {author} {\bibinfo {author} {\bibfnamefont {J.}~\bibnamefont
  {Pathak}}, \bibinfo {author} {\bibfnamefont {B.}~\bibnamefont {Hunt}},
  \bibinfo {author} {\bibfnamefont {M.}~\bibnamefont {Girvan}}, \bibinfo
  {author} {\bibfnamefont {Z.}~\bibnamefont {Lu}},\ and\ \bibinfo {author}
  {\bibfnamefont {E.}~\bibnamefont {Ott}},\ }\bibfield  {title} {\bibinfo
  {title} {Model-free prediction of large spatiotemporally chaotic systems from
  data: a reservoir computing approach},\ }\href@noop {} {\bibfield  {journal}
  {\bibinfo  {journal} {Physical review letters}\ }\textbf {\bibinfo {volume}
  {120}},\ \bibinfo {pages} {024102} (\bibinfo {year} {2018})}\BibitemShut
  {NoStop}%
\bibitem [{\citenamefont {Vlachas}\ \emph {et~al.}(2018)\citenamefont
  {Vlachas}, \citenamefont {Byeon}, \citenamefont {Wan}, \citenamefont
  {Sapsis},\ and\ \citenamefont {Koumoutsakos}}]{vlachas2018data}%
  \BibitemOpen
  \bibfield  {author} {\bibinfo {author} {\bibfnamefont {P.~R.}\ \bibnamefont
  {Vlachas}}, \bibinfo {author} {\bibfnamefont {W.}~\bibnamefont {Byeon}},
  \bibinfo {author} {\bibfnamefont {Z.~Y.}\ \bibnamefont {Wan}}, \bibinfo
  {author} {\bibfnamefont {T.~P.}\ \bibnamefont {Sapsis}},\ and\ \bibinfo
  {author} {\bibfnamefont {P.}~\bibnamefont {Koumoutsakos}},\ }\bibfield
  {title} {\bibinfo {title} {Data-driven forecasting of high-dimensional
  chaotic systems with long short-term memory networks},\ }\href@noop {}
  {\bibfield  {journal} {\bibinfo  {journal} {Proc. R. Soc. A}\ }\textbf
  {\bibinfo {volume} {474}},\ \bibinfo {pages} {20170844} (\bibinfo {year}
  {2018})}\BibitemShut {NoStop}%
\bibitem [{\citenamefont {Raissi}\ \emph {et~al.}(2019)\citenamefont {Raissi},
  \citenamefont {Perdikaris},\ and\ \citenamefont
  {Karniadakis}}]{Raissi2019jcp}%
  \BibitemOpen
  \bibfield  {author} {\bibinfo {author} {\bibfnamefont {M.}~\bibnamefont
  {Raissi}}, \bibinfo {author} {\bibfnamefont {P.}~\bibnamefont {Perdikaris}},\
  and\ \bibinfo {author} {\bibfnamefont {G.}~\bibnamefont {Karniadakis}},\
  }\bibfield  {title} {\bibinfo {title} {Physics-informed neural networks: A
  deep learning framework for solving forward and inverse problems involving
  nonlinear partial differential equations},\ }\href@noop {} {\bibfield
  {journal} {\bibinfo  {journal} {Journal of Computational Physics}\ }\textbf
  {\bibinfo {volume} {378}},\ \bibinfo {pages} {686} (\bibinfo {year}
  {2019})}\BibitemShut {NoStop}%
\bibitem [{\citenamefont {Raissi}\ \emph {et~al.}(2017)\citenamefont {Raissi},
  \citenamefont {Perdikaris},\ and\ \citenamefont
  {Karniadakis}}]{raissi2017machine}%
  \BibitemOpen
  \bibfield  {author} {\bibinfo {author} {\bibfnamefont {M.}~\bibnamefont
  {Raissi}}, \bibinfo {author} {\bibfnamefont {P.}~\bibnamefont {Perdikaris}},\
  and\ \bibinfo {author} {\bibfnamefont {G.~E.}\ \bibnamefont {Karniadakis}},\
  }\bibfield  {title} {\bibinfo {title} {Machine learning of linear
  differential equations using {G}aussian processes},\ }\href@noop {}
  {\bibfield  {journal} {\bibinfo  {journal} {Journal of Computational
  Physics}\ }\textbf {\bibinfo {volume} {348}},\ \bibinfo {pages} {683}
  (\bibinfo {year} {2017})}\BibitemShut {NoStop}%
\bibitem [{\citenamefont {Benner}\ \emph {et~al.}(2015)\citenamefont {Benner},
  \citenamefont {Gugercin},\ and\ \citenamefont
  {Willcox}}]{Benner2015siamreview}%
  \BibitemOpen
  \bibfield  {author} {\bibinfo {author} {\bibfnamefont {P.}~\bibnamefont
  {Benner}}, \bibinfo {author} {\bibfnamefont {S.}~\bibnamefont {Gugercin}},\
  and\ \bibinfo {author} {\bibfnamefont {K.}~\bibnamefont {Willcox}},\
  }\bibfield  {title} {\bibinfo {title} {A survey of projection-based model
  reduction methods for parametric dynamical systems},\ }\href@noop {}
  {\bibfield  {journal} {\bibinfo  {journal} {SIAM review}\ }\textbf {\bibinfo
  {volume} {57}},\ \bibinfo {pages} {483} (\bibinfo {year} {2015})}\BibitemShut
  {NoStop}%
\bibitem [{\citenamefont {Peherstorfer}\ and\ \citenamefont
  {Willcox}(2016)}]{peherstorfer2016data}%
  \BibitemOpen
  \bibfield  {author} {\bibinfo {author} {\bibfnamefont {B.}~\bibnamefont
  {Peherstorfer}}\ and\ \bibinfo {author} {\bibfnamefont {K.}~\bibnamefont
  {Willcox}},\ }\bibfield  {title} {\bibinfo {title} {Data-driven operator
  inference for nonintrusive projection-based model reduction},\ }\href@noop {}
  {\bibfield  {journal} {\bibinfo  {journal} {Computer Methods in Applied
  Mechanics and Engineering}\ }\textbf {\bibinfo {volume} {306}},\ \bibinfo
  {pages} {196} (\bibinfo {year} {2016})}\BibitemShut {NoStop}%
\bibitem [{\citenamefont {Qian}\ \emph {et~al.}(2020)\citenamefont {Qian},
  \citenamefont {Kramer}, \citenamefont {Peherstorfer},\ and\ \citenamefont
  {Willcox}}]{qian2020lift}%
  \BibitemOpen
  \bibfield  {author} {\bibinfo {author} {\bibfnamefont {E.}~\bibnamefont
  {Qian}}, \bibinfo {author} {\bibfnamefont {B.}~\bibnamefont {Kramer}},
  \bibinfo {author} {\bibfnamefont {B.}~\bibnamefont {Peherstorfer}},\ and\
  \bibinfo {author} {\bibfnamefont {K.}~\bibnamefont {Willcox}},\ }\bibfield
  {title} {\bibinfo {title} {Lift \& {L}earn: {P}hysics-informed machine
  learning for large-scale nonlinear dynamical systems},\ }\href@noop {}
  {\bibfield  {journal} {\bibinfo  {journal} {Physica D: Nonlinear Phenomena}\
  }\textbf {\bibinfo {volume} {406}},\ \bibinfo {pages} {132401} (\bibinfo
  {year} {2020})}\BibitemShut {NoStop}%
\bibitem [{\citenamefont {Bongard}\ and\ \citenamefont
  {Lipson}(2007)}]{Bongard2007pnas}%
  \BibitemOpen
  \bibfield  {author} {\bibinfo {author} {\bibfnamefont {J.}~\bibnamefont
  {Bongard}}\ and\ \bibinfo {author} {\bibfnamefont {H.}~\bibnamefont
  {Lipson}},\ }\bibfield  {title} {\bibinfo {title} {Automated reverse
  engineering of nonlinear dynamical systems},\ }\href
  {https://doi.org/10.1073/pnas.0609476104} {\bibfield  {journal} {\bibinfo
  {journal} {Proc. Natl. Acad. Sciences}\ }\textbf {\bibinfo {volume} {104}},\
  \bibinfo {pages} {9943} (\bibinfo {year} {2007})}\BibitemShut {NoStop}%
\bibitem [{\citenamefont {Schmidt}\ and\ \citenamefont
  {Lipson}(2009)}]{schmidt_distilling_2009}%
  \BibitemOpen
  \bibfield  {author} {\bibinfo {author} {\bibfnamefont {M.}~\bibnamefont
  {Schmidt}}\ and\ \bibinfo {author} {\bibfnamefont {H.}~\bibnamefont
  {Lipson}},\ }\bibfield  {title} {\bibinfo {title} {Distilling free-form
  natural laws from experimental data},\ }\href@noop {} {\bibfield  {journal}
  {\bibinfo  {journal} {Science}\ }\textbf {\bibinfo {volume} {324}},\ \bibinfo
  {pages} {81} (\bibinfo {year} {2009})}\BibitemShut {NoStop}%
\bibitem [{\citenamefont {Udrescu}\ and\ \citenamefont
  {Tegmark}(2020)}]{udrescu2020ai}%
  \BibitemOpen
  \bibfield  {author} {\bibinfo {author} {\bibfnamefont {S.-M.}\ \bibnamefont
  {Udrescu}}\ and\ \bibinfo {author} {\bibfnamefont {M.}~\bibnamefont
  {Tegmark}},\ }\bibfield  {title} {\bibinfo {title} {{AI F}eynman: {A}
  physics-inspired method for symbolic regression},\ }\href@noop {} {\bibfield
  {journal} {\bibinfo  {journal} {Science Advances}\ }\textbf {\bibinfo
  {volume} {6}},\ \bibinfo {pages} {eaay2631} (\bibinfo {year}
  {2020})}\BibitemShut {NoStop}%
\bibitem [{\citenamefont {Brunton}\ \emph
  {et~al.}(2016{\natexlab{a}})\citenamefont {Brunton}, \citenamefont
  {Proctor},\ and\ \citenamefont {Kutz}}]{Brunton2016pnas}%
  \BibitemOpen
  \bibfield  {author} {\bibinfo {author} {\bibfnamefont {S.~L.}\ \bibnamefont
  {Brunton}}, \bibinfo {author} {\bibfnamefont {J.~L.}\ \bibnamefont
  {Proctor}},\ and\ \bibinfo {author} {\bibfnamefont {J.~N.}\ \bibnamefont
  {Kutz}},\ }\bibfield  {title} {\bibinfo {title} {Discovering governing
  equations from data by sparse identification of nonlinear dynamical
  systems},\ }\href@noop {} {\bibfield  {journal} {\bibinfo  {journal}
  {Proceedings of the National Academy of Sciences}\ }\textbf {\bibinfo
  {volume} {113}},\ \bibinfo {pages} {3932} (\bibinfo {year}
  {2016}{\natexlab{a}})}\BibitemShut {NoStop}%
\bibitem [{\citenamefont {Hoffmann}\ \emph {et~al.}(2019)\citenamefont
  {Hoffmann}, \citenamefont {Fr{\"o}hner},\ and\ \citenamefont
  {No{\'e}}}]{hoffmann2019reactive}%
  \BibitemOpen
  \bibfield  {author} {\bibinfo {author} {\bibfnamefont {M.}~\bibnamefont
  {Hoffmann}}, \bibinfo {author} {\bibfnamefont {C.}~\bibnamefont
  {Fr{\"o}hner}},\ and\ \bibinfo {author} {\bibfnamefont {F.}~\bibnamefont
  {No{\'e}}},\ }\bibfield  {title} {\bibinfo {title} {Reactive {SIND}y:
  Discovering governing reactions from concentration data},\ }\href@noop {}
  {\bibfield  {journal} {\bibinfo  {journal} {The Journal of chemical physics}\
  }\textbf {\bibinfo {volume} {150}},\ \bibinfo {pages} {025101} (\bibinfo
  {year} {2019})}\BibitemShut {NoStop}%
\bibitem [{\citenamefont {Bhadriraju}\ \emph {et~al.}(2020)\citenamefont
  {Bhadriraju}, \citenamefont {Bangi}, \citenamefont {Narasingam},\ and\
  \citenamefont {Kwon}}]{bhadriraju2020operable}%
  \BibitemOpen
  \bibfield  {author} {\bibinfo {author} {\bibfnamefont {B.}~\bibnamefont
  {Bhadriraju}}, \bibinfo {author} {\bibfnamefont {M.~S.~F.}\ \bibnamefont
  {Bangi}}, \bibinfo {author} {\bibfnamefont {A.}~\bibnamefont {Narasingam}},\
  and\ \bibinfo {author} {\bibfnamefont {J.~S.-I.}\ \bibnamefont {Kwon}},\
  }\bibfield  {title} {\bibinfo {title} {Operable adaptive sparse
  identification of systems: {A}pplication to chemical processes},\ }\href@noop
  {} {\bibfield  {journal} {\bibinfo  {journal} {AIChE Journal}\ }\textbf
  {\bibinfo {volume} {66}},\ \bibinfo {pages} {e16980} (\bibinfo {year}
  {2020})}\BibitemShut {NoStop}%
\bibitem [{\citenamefont {Scheffold}\ \emph {et~al.}(2021)\citenamefont
  {Scheffold}, \citenamefont {Finkler},\ and\ \citenamefont
  {Piechottka}}]{scheffold2021gray}%
  \BibitemOpen
  \bibfield  {author} {\bibinfo {author} {\bibfnamefont {L.}~\bibnamefont
  {Scheffold}}, \bibinfo {author} {\bibfnamefont {T.}~\bibnamefont {Finkler}},\
  and\ \bibinfo {author} {\bibfnamefont {U.}~\bibnamefont {Piechottka}},\
  }\bibfield  {title} {\bibinfo {title} {Gray-box system modeling using
  symbolic regression and nonlinear model predictive control of a semibatch
  polymerization},\ }\href@noop {} {\bibfield  {journal} {\bibinfo  {journal}
  {Computers \& Chemical Engineering}\ }\textbf {\bibinfo {volume} {146}},\
  \bibinfo {pages} {107204} (\bibinfo {year} {2021})}\BibitemShut {NoStop}%
\bibitem [{\citenamefont {Rubio-Herrero}\ \emph {et~al.}(2022)\citenamefont
  {Rubio-Herrero}, \citenamefont {Marrero},\ and\ \citenamefont
  {Fan}}]{rubio2022modeling}%
  \BibitemOpen
  \bibfield  {author} {\bibinfo {author} {\bibfnamefont {J.}~\bibnamefont
  {Rubio-Herrero}}, \bibinfo {author} {\bibfnamefont {C.~O.}\ \bibnamefont
  {Marrero}},\ and\ \bibinfo {author} {\bibfnamefont {W.-T.~L.}\ \bibnamefont
  {Fan}},\ }\bibfield  {title} {\bibinfo {title} {Modeling atmospheric data and
  identifying dynamics temporal data-driven modeling of air pollutants},\
  }\href@noop {} {\bibfield  {journal} {\bibinfo  {journal} {Journal of Cleaner
  Production}\ }\textbf {\bibinfo {volume} {333}},\ \bibinfo {pages} {129863}
  (\bibinfo {year} {2022})}\BibitemShut {NoStop}%
\bibitem [{\citenamefont {Lagergren}\ \emph {et~al.}(2020)\citenamefont
  {Lagergren}, \citenamefont {Nardini}, \citenamefont {Michael~Lavigne},
  \citenamefont {Rutter},\ and\ \citenamefont
  {Flores}}]{lagergren2020learning}%
  \BibitemOpen
  \bibfield  {author} {\bibinfo {author} {\bibfnamefont {J.~H.}\ \bibnamefont
  {Lagergren}}, \bibinfo {author} {\bibfnamefont {J.~T.}\ \bibnamefont
  {Nardini}}, \bibinfo {author} {\bibfnamefont {G.}~\bibnamefont
  {Michael~Lavigne}}, \bibinfo {author} {\bibfnamefont {E.~M.}\ \bibnamefont
  {Rutter}},\ and\ \bibinfo {author} {\bibfnamefont {K.~B.}\ \bibnamefont
  {Flores}},\ }\bibfield  {title} {\bibinfo {title} {Learning partial
  differential equations for biological transport models from noisy
  spatio-temporal data},\ }\href@noop {} {\bibfield  {journal} {\bibinfo
  {journal} {Proceedings of the Royal Society A}\ }\textbf {\bibinfo {volume}
  {476}},\ \bibinfo {pages} {20190800} (\bibinfo {year} {2020})}\BibitemShut
  {NoStop}%
\bibitem [{\citenamefont {Pasquato}\ \emph {et~al.}(2022)\citenamefont
  {Pasquato}, \citenamefont {Abbas}, \citenamefont {Trani}, \citenamefont
  {Nori}, \citenamefont {Kwiecinski}, \citenamefont {Trevisan}, \citenamefont
  {Braga}, \citenamefont {Bono},\ and\ \citenamefont
  {Macci{\`o}}}]{pasquato2022sparse}%
  \BibitemOpen
  \bibfield  {author} {\bibinfo {author} {\bibfnamefont {M.}~\bibnamefont
  {Pasquato}}, \bibinfo {author} {\bibfnamefont {M.}~\bibnamefont {Abbas}},
  \bibinfo {author} {\bibfnamefont {A.~A.}\ \bibnamefont {Trani}}, \bibinfo
  {author} {\bibfnamefont {M.}~\bibnamefont {Nori}}, \bibinfo {author}
  {\bibfnamefont {J.~A.}\ \bibnamefont {Kwiecinski}}, \bibinfo {author}
  {\bibfnamefont {P.}~\bibnamefont {Trevisan}}, \bibinfo {author}
  {\bibfnamefont {V.~F.}\ \bibnamefont {Braga}}, \bibinfo {author}
  {\bibfnamefont {G.}~\bibnamefont {Bono}},\ and\ \bibinfo {author}
  {\bibfnamefont {A.~V.}\ \bibnamefont {Macci{\`o}}},\ }\bibfield  {title}
  {\bibinfo {title} {Sparse identification of variable star dynamics},\
  }\href@noop {} {\bibfield  {journal} {\bibinfo  {journal} {The Astrophysical
  Journal}\ }\textbf {\bibinfo {volume} {930}},\ \bibinfo {pages} {161}
  (\bibinfo {year} {2022})}\BibitemShut {NoStop}%
\bibitem [{\citenamefont {Jiang}\ \emph {et~al.}(2021)\citenamefont {Jiang},
  \citenamefont {Xiong}, \citenamefont {Zhang}, \citenamefont {Wang},
  \citenamefont {Li},\ and\ \citenamefont {Du}}]{jiang2021modeling}%
  \BibitemOpen
  \bibfield  {author} {\bibinfo {author} {\bibfnamefont {Y.-X.}\ \bibnamefont
  {Jiang}}, \bibinfo {author} {\bibfnamefont {X.}~\bibnamefont {Xiong}},
  \bibinfo {author} {\bibfnamefont {S.}~\bibnamefont {Zhang}}, \bibinfo
  {author} {\bibfnamefont {J.-X.}\ \bibnamefont {Wang}}, \bibinfo {author}
  {\bibfnamefont {J.-C.}\ \bibnamefont {Li}},\ and\ \bibinfo {author}
  {\bibfnamefont {L.}~\bibnamefont {Du}},\ }\bibfield  {title} {\bibinfo
  {title} {Modeling and prediction of the transmission dynamics of {COVID}-19
  based on the {SIND}y-{LM} method},\ }\href@noop {} {\bibfield  {journal}
  {\bibinfo  {journal} {Nonlinear Dynamics}\ }\textbf {\bibinfo {volume}
  {105}},\ \bibinfo {pages} {2775} (\bibinfo {year} {2021})}\BibitemShut
  {NoStop}%
\bibitem [{\citenamefont {Zucatti}\ \emph {et~al.}(2020)\citenamefont
  {Zucatti}, \citenamefont {Lui}, \citenamefont {Pitz},\ and\ \citenamefont
  {Wolf}}]{zucatti2020assessment}%
  \BibitemOpen
  \bibfield  {author} {\bibinfo {author} {\bibfnamefont {V.}~\bibnamefont
  {Zucatti}}, \bibinfo {author} {\bibfnamefont {H.~F.}\ \bibnamefont {Lui}},
  \bibinfo {author} {\bibfnamefont {D.~B.}\ \bibnamefont {Pitz}},\ and\
  \bibinfo {author} {\bibfnamefont {W.~R.}\ \bibnamefont {Wolf}},\ }\bibfield
  {title} {\bibinfo {title} {Assessment of reduced-order modeling strategies
  for convective heat transfer},\ }\href@noop {} {\bibfield  {journal}
  {\bibinfo  {journal} {Numerical Heat Transfer, Part A: Applications}\
  }\textbf {\bibinfo {volume} {77}},\ \bibinfo {pages} {702} (\bibinfo {year}
  {2020})}\BibitemShut {NoStop}%
\bibitem [{\citenamefont {Sorokina}\ \emph {et~al.}(2016)\citenamefont
  {Sorokina}, \citenamefont {Sygletos},\ and\ \citenamefont
  {Turitsyn}}]{Sorokina2016oe}%
  \BibitemOpen
  \bibfield  {author} {\bibinfo {author} {\bibfnamefont {M.}~\bibnamefont
  {Sorokina}}, \bibinfo {author} {\bibfnamefont {S.}~\bibnamefont {Sygletos}},\
  and\ \bibinfo {author} {\bibfnamefont {S.}~\bibnamefont {Turitsyn}},\
  }\bibfield  {title} {\bibinfo {title} {Sparse identification for nonlinear
  optical communication systems: {SINO} method},\ }\href@noop {} {\bibfield
  {journal} {\bibinfo  {journal} {Optics express}\ }\textbf {\bibinfo {volume}
  {24}},\ \bibinfo {pages} {30433} (\bibinfo {year} {2016})}\BibitemShut
  {NoStop}%
\bibitem [{\citenamefont {Stankovi{\'c}}\ \emph {et~al.}(2020)\citenamefont
  {Stankovi{\'c}}, \citenamefont {Sari{\'c}}, \citenamefont {Sari{\'c}},\ and\
  \citenamefont {Transtrum}}]{stankovic2020data}%
  \BibitemOpen
  \bibfield  {author} {\bibinfo {author} {\bibfnamefont {A.~M.}\ \bibnamefont
  {Stankovi{\'c}}}, \bibinfo {author} {\bibfnamefont {A.~A.}\ \bibnamefont
  {Sari{\'c}}}, \bibinfo {author} {\bibfnamefont {A.~T.}\ \bibnamefont
  {Sari{\'c}}},\ and\ \bibinfo {author} {\bibfnamefont {M.~K.}\ \bibnamefont
  {Transtrum}},\ }\bibfield  {title} {\bibinfo {title} {Data-driven symbolic
  regression for identification of nonlinear dynamics in power systems},\ }in\
  \href@noop {} {\emph {\bibinfo {booktitle} {2020 IEEE Power \& Energy Society
  General Meeting (PESGM)}}}\ (\bibinfo {organization} {IEEE},\ \bibinfo {year}
  {2020})\ pp.\ \bibinfo {pages} {1--5}\BibitemShut {NoStop}%
\bibitem [{\citenamefont {Cai}\ \emph {et~al.}(2022)\citenamefont {Cai},
  \citenamefont {Wang}, \citenamefont {Joos},\ and\ \citenamefont
  {Kamwa}}]{cai2022online}%
  \BibitemOpen
  \bibfield  {author} {\bibinfo {author} {\bibfnamefont {Y.}~\bibnamefont
  {Cai}}, \bibinfo {author} {\bibfnamefont {X.}~\bibnamefont {Wang}}, \bibinfo
  {author} {\bibfnamefont {G.}~\bibnamefont {Joos}},\ and\ \bibinfo {author}
  {\bibfnamefont {I.}~\bibnamefont {Kamwa}},\ }\bibfield  {title} {\bibinfo
  {title} {An online data-driven method to locate forced oscillation sources
  from power plants based on sparse identification of nonlinear dynamics
  {(SINDy)}},\ }\href@noop {} {\bibfield  {journal} {\bibinfo  {journal} {IEEE
  Transactions on Power Systems}\ } (\bibinfo {year} {2022})}\BibitemShut
  {NoStop}%
\bibitem [{\citenamefont {Narasingam}\ and\ \citenamefont
  {Kwon}(2018)}]{narasingam2018data}%
  \BibitemOpen
  \bibfield  {author} {\bibinfo {author} {\bibfnamefont {A.}~\bibnamefont
  {Narasingam}}\ and\ \bibinfo {author} {\bibfnamefont {J.~S.-I.}\ \bibnamefont
  {Kwon}},\ }\bibfield  {title} {\bibinfo {title} {Data-driven identification
  of interpretable reduced-order models using sparse regression},\ }\href@noop
  {} {\bibfield  {journal} {\bibinfo  {journal} {Computers \& Chemical
  Engineering}\ }\textbf {\bibinfo {volume} {119}},\ \bibinfo {pages} {101}
  (\bibinfo {year} {2018})}\BibitemShut {NoStop}%
\bibitem [{\citenamefont {Thaler}\ \emph {et~al.}(2019)\citenamefont {Thaler},
  \citenamefont {Paehler},\ and\ \citenamefont {Adams}}]{Thaler2019SparseIO}%
  \BibitemOpen
  \bibfield  {author} {\bibinfo {author} {\bibfnamefont {S.}~\bibnamefont
  {Thaler}}, \bibinfo {author} {\bibfnamefont {L.}~\bibnamefont {Paehler}},\
  and\ \bibinfo {author} {\bibfnamefont {N.~A.}\ \bibnamefont {Adams}},\
  }\bibfield  {title} {\bibinfo {title} {Sparse identification of truncation
  errors},\ }\href@noop {} {\bibfield  {journal} {\bibinfo  {journal} {J.
  Comput. Phys.}\ }\textbf {\bibinfo {volume} {397}} (\bibinfo {year}
  {2019})}\BibitemShut {NoStop}%
\bibitem [{\citenamefont {Dale}\ and\ \citenamefont
  {Bhat}(2018)}]{dale2018equations}%
  \BibitemOpen
  \bibfield  {author} {\bibinfo {author} {\bibfnamefont {R.}~\bibnamefont
  {Dale}}\ and\ \bibinfo {author} {\bibfnamefont {H.~S.}\ \bibnamefont
  {Bhat}},\ }\bibfield  {title} {\bibinfo {title} {Equations of mind: {D}ata
  science for inferring nonlinear dynamics of socio-cognitive systems},\
  }\href@noop {} {\bibfield  {journal} {\bibinfo  {journal} {Cognitive Systems
  Research}\ }\textbf {\bibinfo {volume} {52}},\ \bibinfo {pages} {275}
  (\bibinfo {year} {2018})}\BibitemShut {NoStop}%
\bibitem [{\citenamefont {Loiseau}\ and\ \citenamefont
  {Brunton}(2018)}]{loiseau2018constrained}%
  \BibitemOpen
  \bibfield  {author} {\bibinfo {author} {\bibfnamefont {J.-C.}\ \bibnamefont
  {Loiseau}}\ and\ \bibinfo {author} {\bibfnamefont {S.~L.}\ \bibnamefont
  {Brunton}},\ }\bibfield  {title} {\bibinfo {title} {Constrained sparse
  {G}alerkin regression},\ }\href@noop {} {\bibfield  {journal} {\bibinfo
  {journal} {Journal of Fluid Mechanics}\ }\textbf {\bibinfo {volume} {838}},\
  \bibinfo {pages} {42} (\bibinfo {year} {2018})}\BibitemShut {NoStop}%
\bibitem [{\citenamefont {Loiseau}\ \emph {et~al.}(2018)\citenamefont
  {Loiseau}, \citenamefont {Noack},\ and\ \citenamefont
  {Brunton}}]{Loiseau2018jfm}%
  \BibitemOpen
  \bibfield  {author} {\bibinfo {author} {\bibfnamefont {J.-C.}\ \bibnamefont
  {Loiseau}}, \bibinfo {author} {\bibfnamefont {B.~R.}\ \bibnamefont {Noack}},\
  and\ \bibinfo {author} {\bibfnamefont {S.~L.}\ \bibnamefont {Brunton}},\
  }\bibfield  {title} {\bibinfo {title} {Sparse reduced-order modeling:
  sensor-based dynamics to full-state estimation},\ }\href@noop {} {\bibfield
  {journal} {\bibinfo  {journal} {Journal of Fluid Mechanics}\ }\textbf
  {\bibinfo {volume} {844}},\ \bibinfo {pages} {459} (\bibinfo {year}
  {2018})}\BibitemShut {NoStop}%
\bibitem [{\citenamefont {Loiseau}(2020)}]{loiseau2020data}%
  \BibitemOpen
  \bibfield  {author} {\bibinfo {author} {\bibfnamefont {J.-C.}\ \bibnamefont
  {Loiseau}},\ }\bibfield  {title} {\bibinfo {title} {Data-driven modeling of
  the chaotic thermal convection in an annular thermosyphon},\ }\href@noop {}
  {\bibfield  {journal} {\bibinfo  {journal} {Theoretical and Computational
  Fluid Dynamics}\ }\textbf {\bibinfo {volume} {34}},\ \bibinfo {pages} {339}
  (\bibinfo {year} {2020})}\BibitemShut {NoStop}%
\bibitem [{\citenamefont {El~Sayed~M}\ \emph {et~al.}(2018)\citenamefont
  {El~Sayed~M}, \citenamefont {Semaan},\ and\ \citenamefont
  {Radespiel}}]{el2018sparse}%
  \BibitemOpen
  \bibfield  {author} {\bibinfo {author} {\bibfnamefont {Y.}~\bibnamefont
  {El~Sayed~M}}, \bibinfo {author} {\bibfnamefont {R.}~\bibnamefont {Semaan}},\
  and\ \bibinfo {author} {\bibfnamefont {R.}~\bibnamefont {Radespiel}},\
  }\bibfield  {title} {\bibinfo {title} {Sparse modeling of the lift gains of a
  high-lift configuration with periodic coanda blowing},\ }in\ \href@noop {}
  {\emph {\bibinfo {booktitle} {2018 AIAA Aerospace Sciences Meeting}}}\
  (\bibinfo {year} {2018})\ p.\ \bibinfo {pages} {1054}\BibitemShut {NoStop}%
\bibitem [{\citenamefont {Chang}\ and\ \citenamefont
  {Zhang}(2019)}]{chang2019machine}%
  \BibitemOpen
  \bibfield  {author} {\bibinfo {author} {\bibfnamefont {H.}~\bibnamefont
  {Chang}}\ and\ \bibinfo {author} {\bibfnamefont {D.}~\bibnamefont {Zhang}},\
  }\bibfield  {title} {\bibinfo {title} {Machine learning subsurface flow
  equations from data},\ }\href@noop {} {\bibfield  {journal} {\bibinfo
  {journal} {Computational Geosciences}\ }\textbf {\bibinfo {volume} {23}},\
  \bibinfo {pages} {895} (\bibinfo {year} {2019})}\BibitemShut {NoStop}%
\bibitem [{\citenamefont {Deng}\ \emph {et~al.}(2020)\citenamefont {Deng},
  \citenamefont {Noack}, \citenamefont {Morzy{\'n}ski},\ and\ \citenamefont
  {Pastur}}]{deng2020low}%
  \BibitemOpen
  \bibfield  {author} {\bibinfo {author} {\bibfnamefont {N.}~\bibnamefont
  {Deng}}, \bibinfo {author} {\bibfnamefont {B.~R.}\ \bibnamefont {Noack}},
  \bibinfo {author} {\bibfnamefont {M.}~\bibnamefont {Morzy{\'n}ski}},\ and\
  \bibinfo {author} {\bibfnamefont {L.~R.}\ \bibnamefont {Pastur}},\ }\bibfield
   {title} {\bibinfo {title} {Low-order model for successive bifurcations of
  the fluidic pinball},\ }\href@noop {} {\bibfield  {journal} {\bibinfo
  {journal} {Journal of Fluid Mechanics}\ }\textbf {\bibinfo {volume} {884}},\
  \bibinfo {pages} {A37} (\bibinfo {year} {2020})}\BibitemShut {NoStop}%
\bibitem [{\citenamefont {Fukami}\ \emph {et~al.}(2021)\citenamefont {Fukami},
  \citenamefont {Murata}, \citenamefont {Zhang},\ and\ \citenamefont
  {Fukagata}}]{fukami2021sparse}%
  \BibitemOpen
  \bibfield  {author} {\bibinfo {author} {\bibfnamefont {K.}~\bibnamefont
  {Fukami}}, \bibinfo {author} {\bibfnamefont {T.}~\bibnamefont {Murata}},
  \bibinfo {author} {\bibfnamefont {K.}~\bibnamefont {Zhang}},\ and\ \bibinfo
  {author} {\bibfnamefont {K.}~\bibnamefont {Fukagata}},\ }\bibfield  {title}
  {\bibinfo {title} {Sparse identification of nonlinear dynamics with
  low-dimensionalized flow representations},\ }\href@noop {} {\bibfield
  {journal} {\bibinfo  {journal} {Journal of Fluid Mechanics}\ }\textbf
  {\bibinfo {volume} {926}} (\bibinfo {year} {2021})}\BibitemShut {NoStop}%
\bibitem [{\citenamefont {Callaham}\ \emph
  {et~al.}(2022{\natexlab{a}})\citenamefont {Callaham}, \citenamefont
  {Brunton},\ and\ \citenamefont {Loiseau}}]{callaham2022role}%
  \BibitemOpen
  \bibfield  {author} {\bibinfo {author} {\bibfnamefont {J.~L.}\ \bibnamefont
  {Callaham}}, \bibinfo {author} {\bibfnamefont {S.~L.}\ \bibnamefont
  {Brunton}},\ and\ \bibinfo {author} {\bibfnamefont {J.-C.}\ \bibnamefont
  {Loiseau}},\ }\bibfield  {title} {\bibinfo {title} {On the role of nonlinear
  correlations in reduced-order modelling},\ }\href@noop {} {\bibfield
  {journal} {\bibinfo  {journal} {Journal of Fluid Mechanics}\ }\textbf
  {\bibinfo {volume} {938}} (\bibinfo {year} {2022}{\natexlab{a}})}\BibitemShut
  {NoStop}%
\bibitem [{\citenamefont {Khoo}\ \emph {et~al.}(2022)\citenamefont {Khoo},
  \citenamefont {Chan},\ and\ \citenamefont {Hwang}}]{khoo2022sparse}%
  \BibitemOpen
  \bibfield  {author} {\bibinfo {author} {\bibfnamefont {Z.~C.}\ \bibnamefont
  {Khoo}}, \bibinfo {author} {\bibfnamefont {C.~H.}\ \bibnamefont {Chan}},\
  and\ \bibinfo {author} {\bibfnamefont {Y.}~\bibnamefont {Hwang}},\ }\bibfield
   {title} {\bibinfo {title} {A sparse optimal closure for a reduced-order
  model of wall-bounded turbulence},\ }\href@noop {} {\bibfield  {journal}
  {\bibinfo  {journal} {Journal of Fluid Mechanics}\ }\textbf {\bibinfo
  {volume} {939}} (\bibinfo {year} {2022})}\BibitemShut {NoStop}%
\bibitem [{\citenamefont {Deng}\ \emph {et~al.}(2021)\citenamefont {Deng},
  \citenamefont {Noack}, \citenamefont {Morzy{\'n}ski},\ and\ \citenamefont
  {Pastur}}]{deng2021galerkin}%
  \BibitemOpen
  \bibfield  {author} {\bibinfo {author} {\bibfnamefont {N.}~\bibnamefont
  {Deng}}, \bibinfo {author} {\bibfnamefont {B.~R.}\ \bibnamefont {Noack}},
  \bibinfo {author} {\bibfnamefont {M.}~\bibnamefont {Morzy{\'n}ski}},\ and\
  \bibinfo {author} {\bibfnamefont {L.~R.}\ \bibnamefont {Pastur}},\ }\bibfield
   {title} {\bibinfo {title} {Galerkin force model for transient and
  post-transient dynamics of the fluidic pinball},\ }\href@noop {} {\bibfield
  {journal} {\bibinfo  {journal} {Journal of Fluid Mechanics}\ }\textbf
  {\bibinfo {volume} {918}} (\bibinfo {year} {2021})}\BibitemShut {NoStop}%
\bibitem [{\citenamefont {Xiao}\ \emph {et~al.}(2023)\citenamefont {Xiao},
  \citenamefont {Wang}, \citenamefont {Yang},\ and\ \citenamefont
  {Jiang}}]{xiao2023construction}%
  \BibitemOpen
  \bibfield  {author} {\bibinfo {author} {\bibfnamefont {Q.}~\bibnamefont
  {Xiao}}, \bibinfo {author} {\bibfnamefont {J.}~\bibnamefont {Wang}}, \bibinfo
  {author} {\bibfnamefont {X.}~\bibnamefont {Yang}},\ and\ \bibinfo {author}
  {\bibfnamefont {B.}~\bibnamefont {Jiang}},\ }\bibfield  {title} {\bibinfo
  {title} {Construction of a reduced-order model of an electroosmotic
  micromixer and discovery of attractors for petal structure},\ }\href@noop {}
  {\bibfield  {journal} {\bibinfo  {journal} {Physics of Fluids}\ } (\bibinfo
  {year} {2023})}\BibitemShut {NoStop}%
\bibitem [{\citenamefont {Foster}\ \emph {et~al.}(2022)\citenamefont {Foster},
  \citenamefont {Decuyper}, \citenamefont {De~Troyer},\ and\ \citenamefont
  {Runacres}}]{foster2022estimating}%
  \BibitemOpen
  \bibfield  {author} {\bibinfo {author} {\bibfnamefont {J.~A.}\ \bibnamefont
  {Foster}}, \bibinfo {author} {\bibfnamefont {J.}~\bibnamefont {Decuyper}},
  \bibinfo {author} {\bibfnamefont {T.}~\bibnamefont {De~Troyer}},\ and\
  \bibinfo {author} {\bibfnamefont {M.}~\bibnamefont {Runacres}},\ }\bibfield
  {title} {\bibinfo {title} {Estimating a sparse nonlinear dynamical model of
  the flow around an oscillating cylinder in a fluid flow using {SINDy}},\ }in\
  \href@noop {} {\emph {\bibinfo {booktitle} {Conference on Noise and Vibration
  Engineering ISMA 2022}}}\ (\bibinfo {organization} {ISMA 2022},\ \bibinfo
  {year} {2022})\ p.\ \bibinfo {pages} {tba}\BibitemShut {NoStop}%
\bibitem [{\citenamefont {Schmelzer}\ \emph {et~al.}(2020)\citenamefont
  {Schmelzer}, \citenamefont {Dwight},\ and\ \citenamefont
  {Cinnella}}]{schmelzer2020discovery}%
  \BibitemOpen
  \bibfield  {author} {\bibinfo {author} {\bibfnamefont {M.}~\bibnamefont
  {Schmelzer}}, \bibinfo {author} {\bibfnamefont {R.~P.}\ \bibnamefont
  {Dwight}},\ and\ \bibinfo {author} {\bibfnamefont {P.}~\bibnamefont
  {Cinnella}},\ }\bibfield  {title} {\bibinfo {title} {Discovery of algebraic
  {R}eynolds-stress models using sparse symbolic regression},\ }\href@noop {}
  {\bibfield  {journal} {\bibinfo  {journal} {Flow, Turbulence and Combustion}\
  }\textbf {\bibinfo {volume} {104}},\ \bibinfo {pages} {579} (\bibinfo {year}
  {2020})}\BibitemShut {NoStop}%
\bibitem [{\citenamefont {Beetham}\ and\ \citenamefont
  {Capecelatro}(2020)}]{beetham2020formulating}%
  \BibitemOpen
  \bibfield  {author} {\bibinfo {author} {\bibfnamefont {S.}~\bibnamefont
  {Beetham}}\ and\ \bibinfo {author} {\bibfnamefont {J.}~\bibnamefont
  {Capecelatro}},\ }\bibfield  {title} {\bibinfo {title} {Formulating
  turbulence closures using sparse regression with embedded form invariance},\
  }\href@noop {} {\bibfield  {journal} {\bibinfo  {journal} {Physical Review
  Fluids}\ }\textbf {\bibinfo {volume} {5}},\ \bibinfo {pages} {084611}
  (\bibinfo {year} {2020})}\BibitemShut {NoStop}%
\bibitem [{\citenamefont {Beetham}\ and\ \citenamefont
  {Capecelatro}(2021)}]{beetham2021multiphase}%
  \BibitemOpen
  \bibfield  {author} {\bibinfo {author} {\bibfnamefont {S.}~\bibnamefont
  {Beetham}}\ and\ \bibinfo {author} {\bibfnamefont {J.}~\bibnamefont
  {Capecelatro}},\ }\bibfield  {title} {\bibinfo {title} {Multiphase turbulence
  modeling using sparse regression and gene expression programming},\
  }\href@noop {} {\bibfield  {journal} {\bibinfo  {journal} {arXiv preprint
  arXiv:2106.10397}\ } (\bibinfo {year} {2021})}\BibitemShut {NoStop}%
\bibitem [{\citenamefont {Beetham}\ \emph {et~al.}(2021)\citenamefont
  {Beetham}, \citenamefont {Fox},\ and\ \citenamefont
  {Capecelatro}}]{beetham2021sparse}%
  \BibitemOpen
  \bibfield  {author} {\bibinfo {author} {\bibfnamefont {S.}~\bibnamefont
  {Beetham}}, \bibinfo {author} {\bibfnamefont {R.~O.}\ \bibnamefont {Fox}},\
  and\ \bibinfo {author} {\bibfnamefont {J.}~\bibnamefont {Capecelatro}},\
  }\bibfield  {title} {\bibinfo {title} {Sparse identification of multiphase
  turbulence closures for coupled fluid--particle flows},\ }\href@noop {}
  {\bibfield  {journal} {\bibinfo  {journal} {Journal of Fluid Mechanics}\
  }\textbf {\bibinfo {volume} {914}} (\bibinfo {year} {2021})}\BibitemShut
  {NoStop}%
\bibitem [{\citenamefont {Callaham}\ \emph
  {et~al.}(2022{\natexlab{b}})\citenamefont {Callaham}, \citenamefont {Rigas},
  \citenamefont {Loiseau},\ and\ \citenamefont
  {Brunton}}]{callaham2022empirical}%
  \BibitemOpen
  \bibfield  {author} {\bibinfo {author} {\bibfnamefont {J.~L.}\ \bibnamefont
  {Callaham}}, \bibinfo {author} {\bibfnamefont {G.}~\bibnamefont {Rigas}},
  \bibinfo {author} {\bibfnamefont {J.-C.}\ \bibnamefont {Loiseau}},\ and\
  \bibinfo {author} {\bibfnamefont {S.~L.}\ \bibnamefont {Brunton}},\
  }\bibfield  {title} {\bibinfo {title} {An empirical mean-field model of
  symmetry-breaking in a turbulent wake},\ }\href@noop {} {\bibfield  {journal}
  {\bibinfo  {journal} {Science Advances}\ }\textbf {\bibinfo {volume} {8}},\
  \bibinfo {pages} {eabm4786} (\bibinfo {year}
  {2022}{\natexlab{b}})}\BibitemShut {NoStop}%
\bibitem [{\citenamefont {Sansica}\ \emph {et~al.}(2022)\citenamefont
  {Sansica}, \citenamefont {Loiseau}, \citenamefont {Kanamori}, \citenamefont
  {Hashimoto},\ and\ \citenamefont {Robinet}}]{sansica2022system}%
  \BibitemOpen
  \bibfield  {author} {\bibinfo {author} {\bibfnamefont {A.}~\bibnamefont
  {Sansica}}, \bibinfo {author} {\bibfnamefont {J.-C.}\ \bibnamefont
  {Loiseau}}, \bibinfo {author} {\bibfnamefont {M.}~\bibnamefont {Kanamori}},
  \bibinfo {author} {\bibfnamefont {A.}~\bibnamefont {Hashimoto}},\ and\
  \bibinfo {author} {\bibfnamefont {J.-C.}\ \bibnamefont {Robinet}},\
  }\bibfield  {title} {\bibinfo {title} {System identification of
  two-dimensional transonic buffet},\ }\href@noop {} {\bibfield  {journal}
  {\bibinfo  {journal} {AIAA Journal}\ }\textbf {\bibinfo {volume} {60}},\
  \bibinfo {pages} {3090} (\bibinfo {year} {2022})}\BibitemShut {NoStop}%
\bibitem [{\citenamefont {Dam}\ \emph {et~al.}(2017)\citenamefont {Dam},
  \citenamefont {Br{\o}ns}, \citenamefont {Juul~Rasmussen}, \citenamefont
  {Naulin},\ and\ \citenamefont {Hesthaven}}]{Dam2017pf}%
  \BibitemOpen
  \bibfield  {author} {\bibinfo {author} {\bibfnamefont {M.}~\bibnamefont
  {Dam}}, \bibinfo {author} {\bibfnamefont {M.}~\bibnamefont {Br{\o}ns}},
  \bibinfo {author} {\bibfnamefont {J.}~\bibnamefont {Juul~Rasmussen}},
  \bibinfo {author} {\bibfnamefont {V.}~\bibnamefont {Naulin}},\ and\ \bibinfo
  {author} {\bibfnamefont {J.~S.}\ \bibnamefont {Hesthaven}},\ }\bibfield
  {title} {\bibinfo {title} {Sparse identification of a predator-prey system
  from simulation data of a convection model},\ }\href@noop {} {\bibfield
  {journal} {\bibinfo  {journal} {Physics of Plasmas}\ }\textbf {\bibinfo
  {volume} {24}},\ \bibinfo {pages} {022310} (\bibinfo {year}
  {2017})}\BibitemShut {NoStop}%
\bibitem [{\citenamefont {Kaptanoglu}\ \emph
  {et~al.}(2021{\natexlab{a}})\citenamefont {Kaptanoglu}, \citenamefont
  {Morgan}, \citenamefont {Hansen},\ and\ \citenamefont
  {Brunton}}]{kaptanoglu2021physics}%
  \BibitemOpen
  \bibfield  {author} {\bibinfo {author} {\bibfnamefont {A.~A.}\ \bibnamefont
  {Kaptanoglu}}, \bibinfo {author} {\bibfnamefont {K.~D.}\ \bibnamefont
  {Morgan}}, \bibinfo {author} {\bibfnamefont {C.~J.}\ \bibnamefont {Hansen}},\
  and\ \bibinfo {author} {\bibfnamefont {S.~L.}\ \bibnamefont {Brunton}},\
  }\bibfield  {title} {\bibinfo {title} {Physics-constrained, low-dimensional
  models for magnetohydrodynamics: First-principles and data-driven
  approaches},\ }\href@noop {} {\bibfield  {journal} {\bibinfo  {journal}
  {Physical Review E}\ }\textbf {\bibinfo {volume} {104}},\ \bibinfo {pages}
  {015206} (\bibinfo {year} {2021}{\natexlab{a}})}\BibitemShut {NoStop}%
\bibitem [{\citenamefont {Alves}\ and\ \citenamefont
  {Fiuza}(2022)}]{alves2022data}%
  \BibitemOpen
  \bibfield  {author} {\bibinfo {author} {\bibfnamefont {E.~P.}\ \bibnamefont
  {Alves}}\ and\ \bibinfo {author} {\bibfnamefont {F.}~\bibnamefont {Fiuza}},\
  }\bibfield  {title} {\bibinfo {title} {Data-driven discovery of reduced
  plasma physics models from fully kinetic simulations},\ }\href@noop {}
  {\bibfield  {journal} {\bibinfo  {journal} {Physical Review Research}\
  }\textbf {\bibinfo {volume} {4}},\ \bibinfo {pages} {033192} (\bibinfo {year}
  {2022})}\BibitemShut {NoStop}%
\bibitem [{\citenamefont {Lai}\ and\ \citenamefont
  {Nagarajaiah}(2019)}]{lai2019sparse}%
  \BibitemOpen
  \bibfield  {author} {\bibinfo {author} {\bibfnamefont {Z.}~\bibnamefont
  {Lai}}\ and\ \bibinfo {author} {\bibfnamefont {S.}~\bibnamefont
  {Nagarajaiah}},\ }\bibfield  {title} {\bibinfo {title} {Sparse structural
  system identification method for nonlinear dynamic systems with
  hysteresis/inelastic behavior},\ }\href@noop {} {\bibfield  {journal}
  {\bibinfo  {journal} {Mechanical Systems and Signal Processing}\ }\textbf
  {\bibinfo {volume} {117}},\ \bibinfo {pages} {813} (\bibinfo {year}
  {2019})}\BibitemShut {NoStop}%
\bibitem [{\citenamefont {de~Silva}\ \emph
  {et~al.}(2020{\natexlab{a}})\citenamefont {de~Silva}, \citenamefont {Higdon},
  \citenamefont {Brunton},\ and\ \citenamefont {Kutz}}]{de2020discovery}%
  \BibitemOpen
  \bibfield  {author} {\bibinfo {author} {\bibfnamefont {B.~M.}\ \bibnamefont
  {de~Silva}}, \bibinfo {author} {\bibfnamefont {D.~M.}\ \bibnamefont
  {Higdon}}, \bibinfo {author} {\bibfnamefont {S.~L.}\ \bibnamefont
  {Brunton}},\ and\ \bibinfo {author} {\bibfnamefont {J.~N.}\ \bibnamefont
  {Kutz}},\ }\bibfield  {title} {\bibinfo {title} {Discovery of physics from
  data: universal laws and discrepancies},\ }\href@noop {} {\bibfield
  {journal} {\bibinfo  {journal} {Frontiers in artificial intelligence}\
  }\textbf {\bibinfo {volume} {3}},\ \bibinfo {pages} {25} (\bibinfo {year}
  {2020}{\natexlab{a}})}\BibitemShut {NoStop}%
\bibitem [{\citenamefont {Pan}\ \emph {et~al.}(2021)\citenamefont {Pan},
  \citenamefont {Arnold-Medabalimi},\ and\ \citenamefont
  {Duraisamy}}]{pan2021sparsity}%
  \BibitemOpen
  \bibfield  {author} {\bibinfo {author} {\bibfnamefont {S.}~\bibnamefont
  {Pan}}, \bibinfo {author} {\bibfnamefont {N.}~\bibnamefont
  {Arnold-Medabalimi}},\ and\ \bibinfo {author} {\bibfnamefont
  {K.}~\bibnamefont {Duraisamy}},\ }\bibfield  {title} {\bibinfo {title}
  {Sparsity-promoting algorithms for the discovery of informative
  {K}oopman-invariant subspaces},\ }\href@noop {} {\bibfield  {journal}
  {\bibinfo  {journal} {Journal of Fluid Mechanics}\ }\textbf {\bibinfo
  {volume} {917}} (\bibinfo {year} {2021})}\BibitemShut {NoStop}%
\bibitem [{\citenamefont {Subramanian}\ \emph {et~al.}(2021)\citenamefont
  {Subramanian}, \citenamefont {Moar},\ and\ \citenamefont
  {Singh}}]{subramanian2021white}%
  \BibitemOpen
  \bibfield  {author} {\bibinfo {author} {\bibfnamefont {R.}~\bibnamefont
  {Subramanian}}, \bibinfo {author} {\bibfnamefont {R.~R.}\ \bibnamefont
  {Moar}},\ and\ \bibinfo {author} {\bibfnamefont {S.}~\bibnamefont {Singh}},\
  }\bibfield  {title} {\bibinfo {title} {White-box machine learning approaches
  to identify governing equations for overall dynamics of manufacturing
  systems: A case study on distillation column},\ }\href@noop {} {\bibfield
  {journal} {\bibinfo  {journal} {Machine Learning with Applications}\ }\textbf
  {\bibinfo {volume} {3}},\ \bibinfo {pages} {100014} (\bibinfo {year}
  {2021})}\BibitemShut {NoStop}%
\bibitem [{\citenamefont {Brenner}\ \emph {et~al.}(2022)\citenamefont
  {Brenner}, \citenamefont {Hess}, \citenamefont {Mikhaeil}, \citenamefont
  {Bereska}, \citenamefont {Monfared}, \citenamefont {Kuo},\ and\ \citenamefont
  {Durstewitz}}]{brenner2022tractable}%
  \BibitemOpen
  \bibfield  {author} {\bibinfo {author} {\bibfnamefont {M.}~\bibnamefont
  {Brenner}}, \bibinfo {author} {\bibfnamefont {F.}~\bibnamefont {Hess}},
  \bibinfo {author} {\bibfnamefont {J.~M.}\ \bibnamefont {Mikhaeil}}, \bibinfo
  {author} {\bibfnamefont {L.~F.}\ \bibnamefont {Bereska}}, \bibinfo {author}
  {\bibfnamefont {Z.}~\bibnamefont {Monfared}}, \bibinfo {author}
  {\bibfnamefont {P.-C.}\ \bibnamefont {Kuo}},\ and\ \bibinfo {author}
  {\bibfnamefont {D.}~\bibnamefont {Durstewitz}},\ }\bibfield  {title}
  {\bibinfo {title} {Tractable dendritic {RNNs} for reconstructing nonlinear
  dynamical systems},\ }in\ \href@noop {} {\emph {\bibinfo {booktitle}
  {International Conference on Machine Learning}}}\ (\bibinfo {organization}
  {PMLR},\ \bibinfo {year} {2022})\ pp.\ \bibinfo {pages}
  {2292--2320}\BibitemShut {NoStop}%
\bibitem [{\citenamefont {Zhang}\ \emph {et~al.}(2022)\citenamefont {Zhang},
  \citenamefont {Ahamed},\ and\ \citenamefont {Song}}]{zhang2022knowledge}%
  \BibitemOpen
  \bibfield  {author} {\bibinfo {author} {\bibfnamefont {S.}~\bibnamefont
  {Zhang}}, \bibinfo {author} {\bibfnamefont {F.}~\bibnamefont {Ahamed}},\ and\
  \bibinfo {author} {\bibfnamefont {H.-S.}\ \bibnamefont {Song}},\ }\bibfield
  {title} {\bibinfo {title} {Knowledge-informed data-driven modeling for sparse
  identification of governing equations for microbial inactivation processes in
  food},\ }\href@noop {} {\bibfield  {journal} {\bibinfo  {journal} {Frontiers
  in Food Science and Technology}\ ,\ \bibinfo {pages} {29}} (\bibinfo {year}
  {2022})}\BibitemShut {NoStop}%
\bibitem [{\citenamefont {Golden}\ \emph {et~al.}(2022)\citenamefont {Golden},
  \citenamefont {Grigoriev}, \citenamefont {Nambisan},\ and\ \citenamefont
  {Fernandez-Nieves}}]{golden2022physically}%
  \BibitemOpen
  \bibfield  {author} {\bibinfo {author} {\bibfnamefont {M.}~\bibnamefont
  {Golden}}, \bibinfo {author} {\bibfnamefont {R.}~\bibnamefont {Grigoriev}},
  \bibinfo {author} {\bibfnamefont {J.}~\bibnamefont {Nambisan}},\ and\
  \bibinfo {author} {\bibfnamefont {A.}~\bibnamefont {Fernandez-Nieves}},\
  }\bibfield  {title} {\bibinfo {title} {Physically-informed data-driven
  modeling of active nematics},\ }\href@noop {} {\bibfield  {journal} {\bibinfo
   {journal} {arXiv preprint arXiv:2202.12853}\ } (\bibinfo {year}
  {2022})}\BibitemShut {NoStop}%
\bibitem [{\citenamefont {Joshi}\ \emph {et~al.}(2022)\citenamefont {Joshi},
  \citenamefont {Ray}, \citenamefont {Lemma}, \citenamefont {Varghese},
  \citenamefont {Sharp}, \citenamefont {Dogic}, \citenamefont {Baskaran},\ and\
  \citenamefont {Hagan}}]{joshi2022data}%
  \BibitemOpen
  \bibfield  {author} {\bibinfo {author} {\bibfnamefont {C.}~\bibnamefont
  {Joshi}}, \bibinfo {author} {\bibfnamefont {S.}~\bibnamefont {Ray}}, \bibinfo
  {author} {\bibfnamefont {L.~M.}\ \bibnamefont {Lemma}}, \bibinfo {author}
  {\bibfnamefont {M.}~\bibnamefont {Varghese}}, \bibinfo {author}
  {\bibfnamefont {G.}~\bibnamefont {Sharp}}, \bibinfo {author} {\bibfnamefont
  {Z.}~\bibnamefont {Dogic}}, \bibinfo {author} {\bibfnamefont
  {A.}~\bibnamefont {Baskaran}},\ and\ \bibinfo {author} {\bibfnamefont
  {M.~F.}\ \bibnamefont {Hagan}},\ }\bibfield  {title} {\bibinfo {title}
  {Data-driven discovery of active nematic hydrodynamics},\ }\href@noop {}
  {\bibfield  {journal} {\bibinfo  {journal} {Physical review letters}\
  }\textbf {\bibinfo {volume} {129}},\ \bibinfo {pages} {258001} (\bibinfo
  {year} {2022})}\BibitemShut {NoStop}%
\bibitem [{\citenamefont {Schaeffer}(2017)}]{Schaeffer2017prsa}%
  \BibitemOpen
  \bibfield  {author} {\bibinfo {author} {\bibfnamefont {H.}~\bibnamefont
  {Schaeffer}},\ }\bibfield  {title} {\bibinfo {title} {Learning partial
  differential equations via data discovery and sparse optimization},\ }in\
  \href@noop {} {\emph {\bibinfo {booktitle} {Proc. R. Soc. A}}},\ Vol.\
  \bibinfo {volume} {473}\ (\bibinfo {organization} {The Royal Society},\
  \bibinfo {year} {2017})\ p.\ \bibinfo {pages} {20160446}\BibitemShut
  {NoStop}%
\bibitem [{\citenamefont {Rudy}\ \emph {et~al.}(2017)\citenamefont {Rudy},
  \citenamefont {Brunton}, \citenamefont {Proctor},\ and\ \citenamefont
  {Kutz}}]{Rudy2017sciadv}%
  \BibitemOpen
  \bibfield  {author} {\bibinfo {author} {\bibfnamefont {S.~H.}\ \bibnamefont
  {Rudy}}, \bibinfo {author} {\bibfnamefont {S.~L.}\ \bibnamefont {Brunton}},
  \bibinfo {author} {\bibfnamefont {J.~L.}\ \bibnamefont {Proctor}},\ and\
  \bibinfo {author} {\bibfnamefont {J.~N.}\ \bibnamefont {Kutz}},\ }\bibfield
  {title} {\bibinfo {title} {Data-driven discovery of partial differential
  equations},\ }\href@noop {} {\bibfield  {journal} {\bibinfo  {journal}
  {Science Advances}\ }\textbf {\bibinfo {volume} {3}},\ \bibinfo {pages}
  {e1602614} (\bibinfo {year} {2017})}\BibitemShut {NoStop}%
\bibitem [{\citenamefont {Sandoz}\ \emph {et~al.}(2023)\citenamefont {Sandoz},
  \citenamefont {Ducret}, \citenamefont {Gottwald}, \citenamefont {Vilmart},\
  and\ \citenamefont {Perron}}]{sandoz2022sindy}%
  \BibitemOpen
  \bibfield  {author} {\bibinfo {author} {\bibfnamefont {A.}~\bibnamefont
  {Sandoz}}, \bibinfo {author} {\bibfnamefont {V.}~\bibnamefont {Ducret}},
  \bibinfo {author} {\bibfnamefont {G.~A.}\ \bibnamefont {Gottwald}}, \bibinfo
  {author} {\bibfnamefont {G.}~\bibnamefont {Vilmart}},\ and\ \bibinfo {author}
  {\bibfnamefont {K.}~\bibnamefont {Perron}},\ }\bibfield  {title} {\bibinfo
  {title} {{SINDy} for delay-differential equations: application to model
  bacterial zinc response},\ }\href@noop {} {\bibfield  {journal} {\bibinfo
  {journal} {Proceedings of the Royal Society A}\ }\textbf {\bibinfo {volume}
  {479}},\ \bibinfo {pages} {20220556} (\bibinfo {year} {2023})}\BibitemShut
  {NoStop}%
\bibitem [{\citenamefont {Klimovskaia}\ \emph {et~al.}(2016)\citenamefont
  {Klimovskaia}, \citenamefont {Ganscha},\ and\ \citenamefont
  {Claassen}}]{klimovskaia2016sparse}%
  \BibitemOpen
  \bibfield  {author} {\bibinfo {author} {\bibfnamefont {A.}~\bibnamefont
  {Klimovskaia}}, \bibinfo {author} {\bibfnamefont {S.}~\bibnamefont
  {Ganscha}},\ and\ \bibinfo {author} {\bibfnamefont {M.}~\bibnamefont
  {Claassen}},\ }\bibfield  {title} {\bibinfo {title} {Sparse regression based
  structure learning of stochastic reaction networks from single cell snapshot
  time series},\ }\href@noop {} {\bibfield  {journal} {\bibinfo  {journal}
  {PLoS computational biology}\ }\textbf {\bibinfo {volume} {12}},\ \bibinfo
  {pages} {e1005234} (\bibinfo {year} {2016})}\BibitemShut {NoStop}%
\bibitem [{\citenamefont {Br{\"u}ckner}\ \emph {et~al.}(2020)\citenamefont
  {Br{\"u}ckner}, \citenamefont {Ronceray},\ and\ \citenamefont
  {Broedersz}}]{bruckner2020inferring}%
  \BibitemOpen
  \bibfield  {author} {\bibinfo {author} {\bibfnamefont {D.~B.}\ \bibnamefont
  {Br{\"u}ckner}}, \bibinfo {author} {\bibfnamefont {P.}~\bibnamefont
  {Ronceray}},\ and\ \bibinfo {author} {\bibfnamefont {C.~P.}\ \bibnamefont
  {Broedersz}},\ }\bibfield  {title} {\bibinfo {title} {Inferring the dynamics
  of underdamped stochastic systems},\ }\href@noop {} {\bibfield  {journal}
  {\bibinfo  {journal} {Physical review letters}\ }\textbf {\bibinfo {volume}
  {125}},\ \bibinfo {pages} {058103} (\bibinfo {year} {2020})}\BibitemShut
  {NoStop}%
\bibitem [{\citenamefont {Dai}\ \emph {et~al.}(2020)\citenamefont {Dai},
  \citenamefont {Gao}, \citenamefont {Lu}, \citenamefont {Zheng},\ and\
  \citenamefont {Duan}}]{dai2020detecting}%
  \BibitemOpen
  \bibfield  {author} {\bibinfo {author} {\bibfnamefont {M.}~\bibnamefont
  {Dai}}, \bibinfo {author} {\bibfnamefont {T.}~\bibnamefont {Gao}}, \bibinfo
  {author} {\bibfnamefont {Y.}~\bibnamefont {Lu}}, \bibinfo {author}
  {\bibfnamefont {Y.}~\bibnamefont {Zheng}},\ and\ \bibinfo {author}
  {\bibfnamefont {J.}~\bibnamefont {Duan}},\ }\bibfield  {title} {\bibinfo
  {title} {Detecting the maximum likelihood transition path from data of
  stochastic dynamical systems},\ }\href@noop {} {\bibfield  {journal}
  {\bibinfo  {journal} {Chaos: An Interdisciplinary Journal of Nonlinear
  Science}\ }\textbf {\bibinfo {volume} {30}},\ \bibinfo {pages} {113124}
  (\bibinfo {year} {2020})}\BibitemShut {NoStop}%
\bibitem [{\citenamefont {Callaham}\ \emph {et~al.}(2021)\citenamefont
  {Callaham}, \citenamefont {Loiseau}, \citenamefont {Rigas},\ and\
  \citenamefont {Brunton}}]{callaham2021nonlinear}%
  \BibitemOpen
  \bibfield  {author} {\bibinfo {author} {\bibfnamefont {J.~L.}\ \bibnamefont
  {Callaham}}, \bibinfo {author} {\bibfnamefont {J.-C.}\ \bibnamefont
  {Loiseau}}, \bibinfo {author} {\bibfnamefont {G.}~\bibnamefont {Rigas}},\
  and\ \bibinfo {author} {\bibfnamefont {S.~L.}\ \bibnamefont {Brunton}},\
  }\bibfield  {title} {\bibinfo {title} {Nonlinear stochastic modelling with
  {L}angevin regression},\ }\href@noop {} {\bibfield  {journal} {\bibinfo
  {journal} {Proceedings of the Royal Society A}\ }\textbf {\bibinfo {volume}
  {477}},\ \bibinfo {pages} {20210092} (\bibinfo {year} {2021})}\BibitemShut
  {NoStop}%
\bibitem [{\citenamefont {Huang}\ \emph {et~al.}(2022)\citenamefont {Huang},
  \citenamefont {Mabrouk}, \citenamefont {Gompper},\ and\ \citenamefont
  {Sabass}}]{huang2022sparse}%
  \BibitemOpen
  \bibfield  {author} {\bibinfo {author} {\bibfnamefont {Y.}~\bibnamefont
  {Huang}}, \bibinfo {author} {\bibfnamefont {Y.}~\bibnamefont {Mabrouk}},
  \bibinfo {author} {\bibfnamefont {G.}~\bibnamefont {Gompper}},\ and\ \bibinfo
  {author} {\bibfnamefont {B.}~\bibnamefont {Sabass}},\ }\bibfield  {title}
  {\bibinfo {title} {Sparse inference and active learning of stochastic
  differential equations from data},\ }\href@noop {} {\bibfield  {journal}
  {\bibinfo  {journal} {Scientific Reports}\ }\textbf {\bibinfo {volume}
  {12}},\ \bibinfo {pages} {21691} (\bibinfo {year} {2022})}\BibitemShut
  {NoStop}%
\bibitem [{\citenamefont {Hirsh}\ \emph {et~al.}(2022)\citenamefont {Hirsh},
  \citenamefont {Barajas-Solano},\ and\ \citenamefont
  {Kutz}}]{hirsh2022sparsifying}%
  \BibitemOpen
  \bibfield  {author} {\bibinfo {author} {\bibfnamefont {S.~M.}\ \bibnamefont
  {Hirsh}}, \bibinfo {author} {\bibfnamefont {D.~A.}\ \bibnamefont
  {Barajas-Solano}},\ and\ \bibinfo {author} {\bibfnamefont {J.~N.}\
  \bibnamefont {Kutz}},\ }\bibfield  {title} {\bibinfo {title} {Sparsifying
  priors for {B}ayesian uncertainty quantification in model discovery},\
  }\href@noop {} {\bibfield  {journal} {\bibinfo  {journal} {Royal Society Open
  Science}\ }\textbf {\bibinfo {volume} {9}},\ \bibinfo {pages} {211823}
  (\bibinfo {year} {2022})}\BibitemShut {NoStop}%
\bibitem [{\citenamefont {Bakarji}\ \emph
  {et~al.}(2022{\natexlab{a}})\citenamefont {Bakarji}, \citenamefont
  {Callaham}, \citenamefont {Brunton},\ and\ \citenamefont
  {Kutz}}]{bakarji2022dimensionally}%
  \BibitemOpen
  \bibfield  {author} {\bibinfo {author} {\bibfnamefont {J.}~\bibnamefont
  {Bakarji}}, \bibinfo {author} {\bibfnamefont {J.}~\bibnamefont {Callaham}},
  \bibinfo {author} {\bibfnamefont {S.~L.}\ \bibnamefont {Brunton}},\ and\
  \bibinfo {author} {\bibfnamefont {J.~N.}\ \bibnamefont {Kutz}},\ }\bibfield
  {title} {\bibinfo {title} {Dimensionally consistent learning with buckingham
  pi},\ }\href@noop {} {\bibfield  {journal} {\bibinfo  {journal} {Nature
  Computational Science}\ ,\ \bibinfo {pages} {1}} (\bibinfo {year}
  {2022}{\natexlab{a}})}\BibitemShut {NoStop}%
\bibitem [{\citenamefont {Brunton}\ \emph
  {et~al.}(2016{\natexlab{b}})\citenamefont {Brunton}, \citenamefont
  {Proctor},\ and\ \citenamefont {Kutz}}]{brunton2016sparse}%
  \BibitemOpen
  \bibfield  {author} {\bibinfo {author} {\bibfnamefont {S.~L.}\ \bibnamefont
  {Brunton}}, \bibinfo {author} {\bibfnamefont {J.~L.}\ \bibnamefont
  {Proctor}},\ and\ \bibinfo {author} {\bibfnamefont {J.~N.}\ \bibnamefont
  {Kutz}},\ }\bibfield  {title} {\bibinfo {title} {Sparse identification of
  nonlinear dynamics with control {(SINDYc)}},\ }\href@noop {} {\bibfield
  {journal} {\bibinfo  {journal} {IFAC-PapersOnLine}\ }\textbf {\bibinfo
  {volume} {49}},\ \bibinfo {pages} {710} (\bibinfo {year}
  {2016}{\natexlab{b}})}\BibitemShut {NoStop}%
\bibitem [{\citenamefont {Kaiser}\ \emph
  {et~al.}(2018{\natexlab{a}})\citenamefont {Kaiser}, \citenamefont {Kutz},\
  and\ \citenamefont {Brunton}}]{Kaiser2018prsa}%
  \BibitemOpen
  \bibfield  {author} {\bibinfo {author} {\bibfnamefont {E.}~\bibnamefont
  {Kaiser}}, \bibinfo {author} {\bibfnamefont {J.~N.}\ \bibnamefont {Kutz}},\
  and\ \bibinfo {author} {\bibfnamefont {S.~L.}\ \bibnamefont {Brunton}},\
  }\bibfield  {title} {\bibinfo {title} {Sparse identification of nonlinear
  dynamics for model predictive control in the low-data limit},\ }\href@noop {}
  {\bibfield  {journal} {\bibinfo  {journal} {Proceedings of the Royal Society
  of London A}\ }\textbf {\bibinfo {volume} {474}} (\bibinfo {year}
  {2018}{\natexlab{a}})}\BibitemShut {NoStop}%
\bibitem [{\citenamefont {Kaiser}\ \emph
  {et~al.}(2018{\natexlab{b}})\citenamefont {Kaiser}, \citenamefont {Kutz},\
  and\ \citenamefont {Brunton}}]{kaiser2018discovering}%
  \BibitemOpen
  \bibfield  {author} {\bibinfo {author} {\bibfnamefont {E.}~\bibnamefont
  {Kaiser}}, \bibinfo {author} {\bibfnamefont {J.~N.}\ \bibnamefont {Kutz}},\
  and\ \bibinfo {author} {\bibfnamefont {S.~L.}\ \bibnamefont {Brunton}},\
  }\bibfield  {title} {\bibinfo {title} {Discovering conservation laws from
  data for control},\ }in\ \href@noop {} {\emph {\bibinfo {booktitle} {2018
  IEEE Conference on Decision and Control (CDC)}}}\ (\bibinfo {organization}
  {IEEE},\ \bibinfo {year} {2018})\ pp.\ \bibinfo {pages}
  {6415--6421}\BibitemShut {NoStop}%
\bibitem [{\citenamefont {Mangan}\ \emph {et~al.}(2016)\citenamefont {Mangan},
  \citenamefont {Brunton}, \citenamefont {Proctor},\ and\ \citenamefont
  {Kutz}}]{Mangan2016ieee}%
  \BibitemOpen
  \bibfield  {author} {\bibinfo {author} {\bibfnamefont {N.~M.}\ \bibnamefont
  {Mangan}}, \bibinfo {author} {\bibfnamefont {S.~L.}\ \bibnamefont {Brunton}},
  \bibinfo {author} {\bibfnamefont {J.~L.}\ \bibnamefont {Proctor}},\ and\
  \bibinfo {author} {\bibfnamefont {J.~N.}\ \bibnamefont {Kutz}},\ }\bibfield
  {title} {\bibinfo {title} {Inferring biological networks by sparse
  identification of nonlinear dynamics},\ }\href@noop {} {\bibfield  {journal}
  {\bibinfo  {journal} {IEEE Transactions on Molecular, Biological, and
  Multi-Scale Communications}\ }\textbf {\bibinfo {volume} {2}},\ \bibinfo
  {pages} {52} (\bibinfo {year} {2016})}\BibitemShut {NoStop}%
\bibitem [{\citenamefont {Kaheman}\ \emph {et~al.}(2020)\citenamefont
  {Kaheman}, \citenamefont {Kutz},\ and\ \citenamefont
  {Brunton}}]{kaheman2020sindy}%
  \BibitemOpen
  \bibfield  {author} {\bibinfo {author} {\bibfnamefont {K.}~\bibnamefont
  {Kaheman}}, \bibinfo {author} {\bibfnamefont {J.~N.}\ \bibnamefont {Kutz}},\
  and\ \bibinfo {author} {\bibfnamefont {S.~L.}\ \bibnamefont {Brunton}},\
  }\bibfield  {title} {\bibinfo {title} {{SIND}y-{PI}: a robust algorithm for
  parallel implicit sparse identification of nonlinear dynamics},\ }\href@noop
  {} {\bibfield  {journal} {\bibinfo  {journal} {Proceedings of the Royal
  Society A}\ }\textbf {\bibinfo {volume} {476}},\ \bibinfo {pages} {20200279}
  (\bibinfo {year} {2020})}\BibitemShut {NoStop}%
\bibitem [{\citenamefont {Mangan}\ \emph {et~al.}(2019)\citenamefont {Mangan},
  \citenamefont {Askham}, \citenamefont {Brunton}, \citenamefont {Kutz},\ and\
  \citenamefont {Proctor}}]{mangan2019model}%
  \BibitemOpen
  \bibfield  {author} {\bibinfo {author} {\bibfnamefont {N.~M.}\ \bibnamefont
  {Mangan}}, \bibinfo {author} {\bibfnamefont {T.}~\bibnamefont {Askham}},
  \bibinfo {author} {\bibfnamefont {S.~L.}\ \bibnamefont {Brunton}}, \bibinfo
  {author} {\bibfnamefont {J.~N.}\ \bibnamefont {Kutz}},\ and\ \bibinfo
  {author} {\bibfnamefont {J.~L.}\ \bibnamefont {Proctor}},\ }\bibfield
  {title} {\bibinfo {title} {Model selection for hybrid dynamical systems via
  sparse regression},\ }\href@noop {} {\bibfield  {journal} {\bibinfo
  {journal} {Proceedings of the Royal Society A}\ }\textbf {\bibinfo {volume}
  {475}},\ \bibinfo {pages} {20180534} (\bibinfo {year} {2019})}\BibitemShut
  {NoStop}%
\bibitem [{\citenamefont {Thiele}\ \emph {et~al.}(2020)\citenamefont {Thiele},
  \citenamefont {Fey}, \citenamefont {Sommer},\ and\ \citenamefont
  {Kr{\"u}ger}}]{thiele2020system}%
  \BibitemOpen
  \bibfield  {author} {\bibinfo {author} {\bibfnamefont {G.}~\bibnamefont
  {Thiele}}, \bibinfo {author} {\bibfnamefont {A.}~\bibnamefont {Fey}},
  \bibinfo {author} {\bibfnamefont {D.}~\bibnamefont {Sommer}},\ and\ \bibinfo
  {author} {\bibfnamefont {J.}~\bibnamefont {Kr{\"u}ger}},\ }\bibfield  {title}
  {\bibinfo {title} {System identification of a hysteresis-controlled pump
  system using {SIND}y},\ }in\ \href@noop {} {\emph {\bibinfo {booktitle} {2020
  24th International Conference on System Theory, Control and Computing
  (ICSTCC)}}}\ (\bibinfo {organization} {IEEE},\ \bibinfo {year} {2020})\ pp.\
  \bibinfo {pages} {457--464}\BibitemShut {NoStop}%
\bibitem [{\citenamefont {Champion}\ \emph {et~al.}(2020)\citenamefont
  {Champion}, \citenamefont {Zheng}, \citenamefont {Aravkin}, \citenamefont
  {Brunton},\ and\ \citenamefont {Kutz}}]{champion2020unified}%
  \BibitemOpen
  \bibfield  {author} {\bibinfo {author} {\bibfnamefont {K.}~\bibnamefont
  {Champion}}, \bibinfo {author} {\bibfnamefont {P.}~\bibnamefont {Zheng}},
  \bibinfo {author} {\bibfnamefont {A.~Y.}\ \bibnamefont {Aravkin}}, \bibinfo
  {author} {\bibfnamefont {S.~L.}\ \bibnamefont {Brunton}},\ and\ \bibinfo
  {author} {\bibfnamefont {J.~N.}\ \bibnamefont {Kutz}},\ }\bibfield  {title}
  {\bibinfo {title} {A unified sparse optimization framework to learn
  parsimonious physics-informed models from data},\ }\href@noop {} {\bibfield
  {journal} {\bibinfo  {journal} {IEEE Access}\ }\textbf {\bibinfo {volume}
  {8}},\ \bibinfo {pages} {169259} (\bibinfo {year} {2020})}\BibitemShut
  {NoStop}%
\bibitem [{\citenamefont {Mangan}\ \emph {et~al.}(2017)\citenamefont {Mangan},
  \citenamefont {Kutz}, \citenamefont {Brunton},\ and\ \citenamefont
  {Proctor}}]{mangan2017model}%
  \BibitemOpen
  \bibfield  {author} {\bibinfo {author} {\bibfnamefont {N.~M.}\ \bibnamefont
  {Mangan}}, \bibinfo {author} {\bibfnamefont {J.~N.}\ \bibnamefont {Kutz}},
  \bibinfo {author} {\bibfnamefont {S.~L.}\ \bibnamefont {Brunton}},\ and\
  \bibinfo {author} {\bibfnamefont {J.~L.}\ \bibnamefont {Proctor}},\
  }\bibfield  {title} {\bibinfo {title} {Model selection for dynamical systems
  via sparse regression and information criteria},\ }\href@noop {} {\bibfield
  {journal} {\bibinfo  {journal} {Proceedings of the Royal Society A:
  Mathematical, Physical and Engineering Sciences}\ }\textbf {\bibinfo {volume}
  {473}},\ \bibinfo {pages} {20170009} (\bibinfo {year} {2017})}\BibitemShut
  {NoStop}%
\bibitem [{\citenamefont {Dong}\ \emph {et~al.}(2022)\citenamefont {Dong},
  \citenamefont {Bai}, \citenamefont {Lu},\ and\ \citenamefont
  {Fan}}]{dong2022improved}%
  \BibitemOpen
  \bibfield  {author} {\bibinfo {author} {\bibfnamefont {X.}~\bibnamefont
  {Dong}}, \bibinfo {author} {\bibfnamefont {Y.-L.}\ \bibnamefont {Bai}},
  \bibinfo {author} {\bibfnamefont {Y.}~\bibnamefont {Lu}},\ and\ \bibinfo
  {author} {\bibfnamefont {M.}~\bibnamefont {Fan}},\ }\bibfield  {title}
  {\bibinfo {title} {An improved sparse identification of nonlinear dynamics
  with {A}kaike information criterion and group sparsity},\ }\href@noop {}
  {\bibfield  {journal} {\bibinfo  {journal} {Nonlinear Dynamics}\ ,\ \bibinfo
  {pages} {1}} (\bibinfo {year} {2022})}\BibitemShut {NoStop}%
\bibitem [{\citenamefont {Kaptanoglu}\ \emph
  {et~al.}(2021{\natexlab{b}})\citenamefont {Kaptanoglu}, \citenamefont
  {Callaham}, \citenamefont {Aravkin}, \citenamefont {Hansen},\ and\
  \citenamefont {Brunton}}]{kaptanoglu2021promoting}%
  \BibitemOpen
  \bibfield  {author} {\bibinfo {author} {\bibfnamefont {A.~A.}\ \bibnamefont
  {Kaptanoglu}}, \bibinfo {author} {\bibfnamefont {J.~L.}\ \bibnamefont
  {Callaham}}, \bibinfo {author} {\bibfnamefont {A.}~\bibnamefont {Aravkin}},
  \bibinfo {author} {\bibfnamefont {C.~J.}\ \bibnamefont {Hansen}},\ and\
  \bibinfo {author} {\bibfnamefont {S.~L.}\ \bibnamefont {Brunton}},\
  }\bibfield  {title} {\bibinfo {title} {Promoting global stability in
  data-driven models of quadratic nonlinear dynamics},\ }\href@noop {}
  {\bibfield  {journal} {\bibinfo  {journal} {Physical Review Fluids}\ }\textbf
  {\bibinfo {volume} {6}},\ \bibinfo {pages} {094401} (\bibinfo {year}
  {2021}{\natexlab{b}})}\BibitemShut {NoStop}%
\bibitem [{\citenamefont {Tran}\ and\ \citenamefont
  {Ward}(2017)}]{tran2017exact}%
  \BibitemOpen
  \bibfield  {author} {\bibinfo {author} {\bibfnamefont {G.}~\bibnamefont
  {Tran}}\ and\ \bibinfo {author} {\bibfnamefont {R.}~\bibnamefont {Ward}},\
  }\bibfield  {title} {\bibinfo {title} {Exact recovery of chaotic systems from
  highly corrupted data},\ }\href {https://doi.org/10.1137/16m1086637}
  {\bibfield  {journal} {\bibinfo  {journal} {Multiscale Modeling \&
  Simulation}\ }\textbf {\bibinfo {volume} {15}},\ \bibinfo {pages} {1108}
  (\bibinfo {year} {2017})}\BibitemShut {NoStop}%
\bibitem [{\citenamefont {Schaeffer}\ \emph {et~al.}(2018)\citenamefont
  {Schaeffer}, \citenamefont {Tran},\ and\ \citenamefont
  {Ward}}]{schaeffer2018extracting}%
  \BibitemOpen
  \bibfield  {author} {\bibinfo {author} {\bibfnamefont {H.}~\bibnamefont
  {Schaeffer}}, \bibinfo {author} {\bibfnamefont {G.}~\bibnamefont {Tran}},\
  and\ \bibinfo {author} {\bibfnamefont {R.}~\bibnamefont {Ward}},\ }\bibfield
  {title} {\bibinfo {title} {Extracting sparse high-dimensional dynamics from
  limited data},\ }\href@noop {} {\bibfield  {journal} {\bibinfo  {journal}
  {SIAM Journal on Applied Mathematics}\ }\textbf {\bibinfo {volume} {78}},\
  \bibinfo {pages} {3279} (\bibinfo {year} {2018})}\BibitemShut {NoStop}%
\bibitem [{\citenamefont {Delahunt}\ and\ \citenamefont
  {Kutz}(2022)}]{delahunt2022toolkit}%
  \BibitemOpen
  \bibfield  {author} {\bibinfo {author} {\bibfnamefont {C.~B.}\ \bibnamefont
  {Delahunt}}\ and\ \bibinfo {author} {\bibfnamefont {J.~N.}\ \bibnamefont
  {Kutz}},\ }\bibfield  {title} {\bibinfo {title} {A toolkit for data-driven
  discovery of governing equations in high-noise regimes},\ }\href@noop {}
  {\bibfield  {journal} {\bibinfo  {journal} {IEEE Access}\ }\textbf {\bibinfo
  {volume} {10}},\ \bibinfo {pages} {31210} (\bibinfo {year}
  {2022})}\BibitemShut {NoStop}%
\bibitem [{\citenamefont {Wentz}\ and\ \citenamefont
  {Doostan}(2022)}]{wentz2022derivative}%
  \BibitemOpen
  \bibfield  {author} {\bibinfo {author} {\bibfnamefont {J.}~\bibnamefont
  {Wentz}}\ and\ \bibinfo {author} {\bibfnamefont {A.}~\bibnamefont
  {Doostan}},\ }\bibfield  {title} {\bibinfo {title} {Derivative-based {SINDy
  (DSINDy): A}ddressing the challenge of discovering governing equations from
  noisy data},\ }\href@noop {} {\bibfield  {journal} {\bibinfo  {journal}
  {arXiv preprint arXiv:2211.05918}\ } (\bibinfo {year} {2022})}\BibitemShut
  {NoStop}%
\bibitem [{\citenamefont {Kaheman}\ \emph {et~al.}(2022)\citenamefont
  {Kaheman}, \citenamefont {Brunton},\ and\ \citenamefont
  {Kutz}}]{kaheman2022automatic}%
  \BibitemOpen
  \bibfield  {author} {\bibinfo {author} {\bibfnamefont {K.}~\bibnamefont
  {Kaheman}}, \bibinfo {author} {\bibfnamefont {S.~L.}\ \bibnamefont
  {Brunton}},\ and\ \bibinfo {author} {\bibfnamefont {J.~N.}\ \bibnamefont
  {Kutz}},\ }\bibfield  {title} {\bibinfo {title} {Automatic differentiation to
  simultaneously identify nonlinear dynamics and extract noise probability
  distributions from data},\ }\href@noop {} {\bibfield  {journal} {\bibinfo
  {journal} {Machine Learning: Science and Technology}\ }\textbf {\bibinfo
  {volume} {3}},\ \bibinfo {pages} {015031} (\bibinfo {year}
  {2022})}\BibitemShut {NoStop}%
\bibitem [{\citenamefont {Somacal}\ \emph {et~al.}(2022)\citenamefont
  {Somacal}, \citenamefont {Barrera}, \citenamefont {Boechi}, \citenamefont
  {Jonckheere}, \citenamefont {Lefieux}, \citenamefont {Picard},\ and\
  \citenamefont {Smucler}}]{somacal2022uncovering}%
  \BibitemOpen
  \bibfield  {author} {\bibinfo {author} {\bibfnamefont {A.}~\bibnamefont
  {Somacal}}, \bibinfo {author} {\bibfnamefont {Y.}~\bibnamefont {Barrera}},
  \bibinfo {author} {\bibfnamefont {L.}~\bibnamefont {Boechi}}, \bibinfo
  {author} {\bibfnamefont {M.}~\bibnamefont {Jonckheere}}, \bibinfo {author}
  {\bibfnamefont {V.}~\bibnamefont {Lefieux}}, \bibinfo {author} {\bibfnamefont
  {D.}~\bibnamefont {Picard}},\ and\ \bibinfo {author} {\bibfnamefont
  {E.}~\bibnamefont {Smucler}},\ }\bibfield  {title} {\bibinfo {title}
  {Uncovering differential equations from data with hidden variables},\
  }\href@noop {} {\bibfield  {journal} {\bibinfo  {journal} {Physical Review
  E}\ }\textbf {\bibinfo {volume} {105}},\ \bibinfo {pages} {054209} (\bibinfo
  {year} {2022})}\BibitemShut {NoStop}%
\bibitem [{\citenamefont {Bakarji}\ \emph
  {et~al.}(2022{\natexlab{b}})\citenamefont {Bakarji}, \citenamefont
  {Champion}, \citenamefont {Kutz},\ and\ \citenamefont
  {Brunton}}]{bakarji2022discovering}%
  \BibitemOpen
  \bibfield  {author} {\bibinfo {author} {\bibfnamefont {J.}~\bibnamefont
  {Bakarji}}, \bibinfo {author} {\bibfnamefont {K.}~\bibnamefont {Champion}},
  \bibinfo {author} {\bibfnamefont {J.~N.}\ \bibnamefont {Kutz}},\ and\
  \bibinfo {author} {\bibfnamefont {S.~L.}\ \bibnamefont {Brunton}},\
  }\bibfield  {title} {\bibinfo {title} {Discovering governing equations from
  partial measurements with deep delay autoencoders},\ }\href@noop {}
  {\bibfield  {journal} {\bibinfo  {journal} {arXiv preprint arXiv:2201.05136}\
  } (\bibinfo {year} {2022}{\natexlab{b}})}\BibitemShut {NoStop}%
\bibitem [{\citenamefont {Conti}\ \emph {et~al.}(2022)\citenamefont {Conti},
  \citenamefont {Gobat}, \citenamefont {Fresca}, \citenamefont {Manzoni},\ and\
  \citenamefont {Frangi}}]{conti2022reduced}%
  \BibitemOpen
  \bibfield  {author} {\bibinfo {author} {\bibfnamefont {P.}~\bibnamefont
  {Conti}}, \bibinfo {author} {\bibfnamefont {G.}~\bibnamefont {Gobat}},
  \bibinfo {author} {\bibfnamefont {S.}~\bibnamefont {Fresca}}, \bibinfo
  {author} {\bibfnamefont {A.}~\bibnamefont {Manzoni}},\ and\ \bibinfo {author}
  {\bibfnamefont {A.}~\bibnamefont {Frangi}},\ }\bibfield  {title} {\bibinfo
  {title} {Reduced order modeling of parametrized systems through autoencoders
  and {SINDy} approach: continuation of periodic solutions},\ }\href@noop {}
  {\bibfield  {journal} {\bibinfo  {journal} {arXiv preprint arXiv:2211.06786}\
  } (\bibinfo {year} {2022})}\BibitemShut {NoStop}%
\bibitem [{\citenamefont {Gao}\ and\ \citenamefont
  {Kutz}(2022)}]{gao2022bayesian}%
  \BibitemOpen
  \bibfield  {author} {\bibinfo {author} {\bibfnamefont {L.}~\bibnamefont
  {Gao}}\ and\ \bibinfo {author} {\bibfnamefont {J.~N.}\ \bibnamefont {Kutz}},\
  }\bibfield  {title} {\bibinfo {title} {{B}ayesian autoencoders for
  data-driven discovery of coordinates, governing equations and fundamental
  constants},\ }\href@noop {} {\bibfield  {journal} {\bibinfo  {journal} {arXiv
  preprint arXiv:2211.10575}\ } (\bibinfo {year} {2022})}\BibitemShut {NoStop}%
\bibitem [{\citenamefont {Zhao}\ and\ \citenamefont
  {Li}(2022)}]{zhao2022adaptive}%
  \BibitemOpen
  \bibfield  {author} {\bibinfo {author} {\bibfnamefont {Z.}~\bibnamefont
  {Zhao}}\ and\ \bibinfo {author} {\bibfnamefont {Q.}~\bibnamefont {Li}},\
  }\bibfield  {title} {\bibinfo {title} {Adaptive sampling methods for learning
  dynamical systems},\ }in\ \href@noop {} {\emph {\bibinfo {booktitle}
  {Mathematical and Scientific Machine Learning}}}\ (\bibinfo {organization}
  {PMLR},\ \bibinfo {year} {2022})\ pp.\ \bibinfo {pages}
  {335--350}\BibitemShut {NoStop}%
\bibitem [{\citenamefont {Wu}\ and\ \citenamefont
  {Xiu}(2019)}]{wu2019numerical}%
  \BibitemOpen
  \bibfield  {author} {\bibinfo {author} {\bibfnamefont {K.}~\bibnamefont
  {Wu}}\ and\ \bibinfo {author} {\bibfnamefont {D.}~\bibnamefont {Xiu}},\
  }\bibfield  {title} {\bibinfo {title} {Numerical aspects for approximating
  governing equations using data},\ }\href@noop {} {\bibfield  {journal}
  {\bibinfo  {journal} {Journal of Computational Physics}\ }\textbf {\bibinfo
  {volume} {384}},\ \bibinfo {pages} {200} (\bibinfo {year}
  {2019})}\BibitemShut {NoStop}%
\bibitem [{\citenamefont {Fasel}\ \emph {et~al.}(2022)\citenamefont {Fasel},
  \citenamefont {Kutz}, \citenamefont {Brunton},\ and\ \citenamefont
  {Brunton}}]{fasel2022ensemble}%
  \BibitemOpen
  \bibfield  {author} {\bibinfo {author} {\bibfnamefont {U.}~\bibnamefont
  {Fasel}}, \bibinfo {author} {\bibfnamefont {J.~N.}\ \bibnamefont {Kutz}},
  \bibinfo {author} {\bibfnamefont {B.~W.}\ \bibnamefont {Brunton}},\ and\
  \bibinfo {author} {\bibfnamefont {S.~L.}\ \bibnamefont {Brunton}},\
  }\bibfield  {title} {\bibinfo {title} {Ensemble-{SINDy: R}obust sparse model
  discovery in the low-data, high-noise limit, with active learning and
  control},\ }\href@noop {} {\bibfield  {journal} {\bibinfo  {journal}
  {Proceedings of the Royal Society A}\ }\textbf {\bibinfo {volume} {478}},\
  \bibinfo {pages} {20210904} (\bibinfo {year} {2022})}\BibitemShut {NoStop}%
\bibitem [{\citenamefont {Gao}\ \emph {et~al.}(2023)\citenamefont {Gao},
  \citenamefont {Fasel}, \citenamefont {Brunton},\ and\ \citenamefont
  {Kutz}}]{gao2023convergence}%
  \BibitemOpen
  \bibfield  {author} {\bibinfo {author} {\bibfnamefont {L.}~\bibnamefont
  {Gao}}, \bibinfo {author} {\bibfnamefont {U.}~\bibnamefont {Fasel}}, \bibinfo
  {author} {\bibfnamefont {S.~L.}\ \bibnamefont {Brunton}},\ and\ \bibinfo
  {author} {\bibfnamefont {J.~N.}\ \bibnamefont {Kutz}},\ }\bibfield  {title}
  {\bibinfo {title} {Convergence of uncertainty estimates in ensemble and
  {B}ayesian sparse model discovery},\ }\href@noop {} {\bibfield  {journal}
  {\bibinfo  {journal} {arXiv preprint arXiv:2301.12649}\ } (\bibinfo {year}
  {2023})}\BibitemShut {NoStop}%
\bibitem [{\citenamefont {Jiang}\ \emph {et~al.}(2023)\citenamefont {Jiang},
  \citenamefont {Du}, \citenamefont {Yang},\ and\ \citenamefont
  {Deng}}]{jiang2023regularized}%
  \BibitemOpen
  \bibfield  {author} {\bibinfo {author} {\bibfnamefont {F.}~\bibnamefont
  {Jiang}}, \bibinfo {author} {\bibfnamefont {L.}~\bibnamefont {Du}}, \bibinfo
  {author} {\bibfnamefont {F.}~\bibnamefont {Yang}},\ and\ \bibinfo {author}
  {\bibfnamefont {Z.-C.}\ \bibnamefont {Deng}},\ }\bibfield  {title} {\bibinfo
  {title} {Regularized least absolute deviation-based sparse identification of
  dynamical systems},\ }\href@noop {} {\bibfield  {journal} {\bibinfo
  {journal} {Chaos: An Interdisciplinary Journal of Nonlinear Science}\
  }\textbf {\bibinfo {volume} {33}},\ \bibinfo {pages} {013103} (\bibinfo
  {year} {2023})}\BibitemShut {NoStop}%
\bibitem [{\citenamefont {Schaeffer}\ and\ \citenamefont
  {McCalla}(2017)}]{Schaeffer2017pre}%
  \BibitemOpen
  \bibfield  {author} {\bibinfo {author} {\bibfnamefont {H.}~\bibnamefont
  {Schaeffer}}\ and\ \bibinfo {author} {\bibfnamefont {S.~G.}\ \bibnamefont
  {McCalla}},\ }\bibfield  {title} {\bibinfo {title} {Sparse model selection
  via integral terms},\ }\href@noop {} {\bibfield  {journal} {\bibinfo
  {journal} {Physical Review E}\ }\textbf {\bibinfo {volume} {96}},\ \bibinfo
  {pages} {023302} (\bibinfo {year} {2017})}\BibitemShut {NoStop}%
\bibitem [{\citenamefont {Gurevich}\ \emph {et~al.}(2019)\citenamefont
  {Gurevich}, \citenamefont {Reinbold},\ and\ \citenamefont
  {Grigoriev}}]{gurevich2019robust}%
  \BibitemOpen
  \bibfield  {author} {\bibinfo {author} {\bibfnamefont {D.~R.}\ \bibnamefont
  {Gurevich}}, \bibinfo {author} {\bibfnamefont {P.~A.}\ \bibnamefont
  {Reinbold}},\ and\ \bibinfo {author} {\bibfnamefont {R.~O.}\ \bibnamefont
  {Grigoriev}},\ }\bibfield  {title} {\bibinfo {title} {Robust and optimal
  sparse regression for nonlinear {PDE} models},\ }\href@noop {} {\bibfield
  {journal} {\bibinfo  {journal} {Chaos: An Interdisciplinary Journal of
  Nonlinear Science}\ }\textbf {\bibinfo {volume} {29}},\ \bibinfo {pages}
  {103113} (\bibinfo {year} {2019})}\BibitemShut {NoStop}%
\bibitem [{\citenamefont {Reinbold}\ \emph {et~al.}(2020)\citenamefont
  {Reinbold}, \citenamefont {Gurevich},\ and\ \citenamefont
  {Grigoriev}}]{reinbold2020using}%
  \BibitemOpen
  \bibfield  {author} {\bibinfo {author} {\bibfnamefont {P.~A.}\ \bibnamefont
  {Reinbold}}, \bibinfo {author} {\bibfnamefont {D.~R.}\ \bibnamefont
  {Gurevich}},\ and\ \bibinfo {author} {\bibfnamefont {R.~O.}\ \bibnamefont
  {Grigoriev}},\ }\bibfield  {title} {\bibinfo {title} {Using noisy or
  incomplete data to discover models of spatiotemporal dynamics},\ }\href@noop
  {} {\bibfield  {journal} {\bibinfo  {journal} {Physical Review E}\ }\textbf
  {\bibinfo {volume} {101}},\ \bibinfo {pages} {010203} (\bibinfo {year}
  {2020})}\BibitemShut {NoStop}%
\bibitem [{\citenamefont {Messenger}\ and\ \citenamefont
  {Bortz}(2021)}]{messenger2021weakpde}%
  \BibitemOpen
  \bibfield  {author} {\bibinfo {author} {\bibfnamefont {D.~A.}\ \bibnamefont
  {Messenger}}\ and\ \bibinfo {author} {\bibfnamefont {D.~M.}\ \bibnamefont
  {Bortz}},\ }\bibfield  {title} {\bibinfo {title} {Weak {SIND}y for partial
  differential equations},\ }\href@noop {} {\bibfield  {journal} {\bibinfo
  {journal} {Journal of Computational Physics}\ ,\ \bibinfo {pages} {110525}}
  (\bibinfo {year} {2021})}\BibitemShut {NoStop}%
\bibitem [{\citenamefont {Kageorge}\ \emph {et~al.}(2021)\citenamefont
  {Kageorge}, \citenamefont {Grigoriev},\ and\ \citenamefont
  {Schatz}}]{kageorge2021data}%
  \BibitemOpen
  \bibfield  {author} {\bibinfo {author} {\bibfnamefont {L.~M.}\ \bibnamefont
  {Kageorge}}, \bibinfo {author} {\bibfnamefont {R.~O.}\ \bibnamefont
  {Grigoriev}},\ and\ \bibinfo {author} {\bibfnamefont {M.~F.}\ \bibnamefont
  {Schatz}},\ }\bibfield  {title} {\bibinfo {title} {Data-driven detection of
  drifting system parameters},\ }\href@noop {} {\bibfield  {journal} {\bibinfo
  {journal} {arXiv preprint arXiv:2111.12114}\ } (\bibinfo {year}
  {2021})}\BibitemShut {NoStop}%
\bibitem [{\citenamefont {Reinbold}\ \emph {et~al.}(2021)\citenamefont
  {Reinbold}, \citenamefont {Kageorge}, \citenamefont {Schatz},\ and\
  \citenamefont {Grigoriev}}]{reinbold2021robust}%
  \BibitemOpen
  \bibfield  {author} {\bibinfo {author} {\bibfnamefont {P.~A.}\ \bibnamefont
  {Reinbold}}, \bibinfo {author} {\bibfnamefont {L.~M.}\ \bibnamefont
  {Kageorge}}, \bibinfo {author} {\bibfnamefont {M.~F.}\ \bibnamefont
  {Schatz}},\ and\ \bibinfo {author} {\bibfnamefont {R.~O.}\ \bibnamefont
  {Grigoriev}},\ }\bibfield  {title} {\bibinfo {title} {Robust learning from
  noisy, incomplete, high-dimensional experimental data via physically
  constrained symbolic regression},\ }\href@noop {} {\bibfield  {journal}
  {\bibinfo  {journal} {Nature communications}\ }\textbf {\bibinfo {volume}
  {12}},\ \bibinfo {pages} {1} (\bibinfo {year} {2021})}\BibitemShut {NoStop}%
\bibitem [{\citenamefont {Gurevich}\ \emph {et~al.}(2021)\citenamefont
  {Gurevich}, \citenamefont {Reinbold},\ and\ \citenamefont
  {Grigoriev}}]{gurevich2021learning}%
  \BibitemOpen
  \bibfield  {author} {\bibinfo {author} {\bibfnamefont {D.~R.}\ \bibnamefont
  {Gurevich}}, \bibinfo {author} {\bibfnamefont {P.~A.}\ \bibnamefont
  {Reinbold}},\ and\ \bibinfo {author} {\bibfnamefont {R.~O.}\ \bibnamefont
  {Grigoriev}},\ }\bibfield  {title} {\bibinfo {title} {Learning fluid physics
  from highly turbulent data using sparse physics-informed discovery of
  empirical relations {(SPIDER)}},\ }\href@noop {} {\bibfield  {journal}
  {\bibinfo  {journal} {arXiv preprint arXiv:2105.00048}\ } (\bibinfo {year}
  {2021})}\BibitemShut {NoStop}%
\bibitem [{\citenamefont {Russo}\ and\ \citenamefont
  {Laiu}(2022)}]{russo2022convergence}%
  \BibitemOpen
  \bibfield  {author} {\bibinfo {author} {\bibfnamefont {B.}~\bibnamefont
  {Russo}}\ and\ \bibinfo {author} {\bibfnamefont {M.~P.}\ \bibnamefont
  {Laiu}},\ }\bibfield  {title} {\bibinfo {title} {Convergence of weak-{SINDy}
  surrogate models},\ }\href@noop {} {\bibfield  {journal} {\bibinfo  {journal}
  {arXiv preprint arXiv:2209.15573}\ } (\bibinfo {year} {2022})}\BibitemShut
  {NoStop}%
\bibitem [{\citenamefont {Messenger}\ and\ \citenamefont
  {Bortz}(2022{\natexlab{a}})}]{messenger2022asymptotic}%
  \BibitemOpen
  \bibfield  {author} {\bibinfo {author} {\bibfnamefont {D.~A.}\ \bibnamefont
  {Messenger}}\ and\ \bibinfo {author} {\bibfnamefont {D.~M.}\ \bibnamefont
  {Bortz}},\ }\bibfield  {title} {\bibinfo {title} {Asymptotic consistency of
  the {WSINDy} algorithm in the limit of continuum data},\ }\href@noop {}
  {\bibfield  {journal} {\bibinfo  {journal} {arXiv preprint arXiv:2211.16000}\
  } (\bibinfo {year} {2022}{\natexlab{a}})}\BibitemShut {NoStop}%
\bibitem [{\citenamefont {Messenger}\ and\ \citenamefont
  {Bortz}(2022{\natexlab{b}})}]{messenger2022learning}%
  \BibitemOpen
  \bibfield  {author} {\bibinfo {author} {\bibfnamefont {D.~A.}\ \bibnamefont
  {Messenger}}\ and\ \bibinfo {author} {\bibfnamefont {D.~M.}\ \bibnamefont
  {Bortz}},\ }\bibfield  {title} {\bibinfo {title} {Learning mean-field
  equations from particle data using {WSINDy}},\ }\href@noop {} {\bibfield
  {journal} {\bibinfo  {journal} {Physica D: Nonlinear Phenomena}\ }\textbf
  {\bibinfo {volume} {439}},\ \bibinfo {pages} {133406} (\bibinfo {year}
  {2022}{\natexlab{b}})}\BibitemShut {NoStop}%
\bibitem [{\citenamefont {Messenger}\ \emph {et~al.}(2022)\citenamefont
  {Messenger}, \citenamefont {Dall’Anese},\ and\ \citenamefont
  {Bortz}}]{messenger2022online}%
  \BibitemOpen
  \bibfield  {author} {\bibinfo {author} {\bibfnamefont {D.~A.}\ \bibnamefont
  {Messenger}}, \bibinfo {author} {\bibfnamefont {E.}~\bibnamefont
  {Dall’Anese}},\ and\ \bibinfo {author} {\bibfnamefont {D.}~\bibnamefont
  {Bortz}},\ }\bibfield  {title} {\bibinfo {title} {Online weak-form sparse
  identification of partial differential equations},\ }in\ \href@noop {} {\emph
  {\bibinfo {booktitle} {Mathematical and Scientific Machine Learning}}}\
  (\bibinfo {organization} {PMLR},\ \bibinfo {year} {2022})\ pp.\ \bibinfo
  {pages} {241--256}\BibitemShut {NoStop}%
\bibitem [{\citenamefont {Gel{\ss}}\ \emph {et~al.}(2019)\citenamefont
  {Gel{\ss}}, \citenamefont {Klus}, \citenamefont {Eisert},\ and\ \citenamefont
  {Sch{\"u}tte}}]{gelss2019multidimensional}%
  \BibitemOpen
  \bibfield  {author} {\bibinfo {author} {\bibfnamefont {P.}~\bibnamefont
  {Gel{\ss}}}, \bibinfo {author} {\bibfnamefont {S.}~\bibnamefont {Klus}},
  \bibinfo {author} {\bibfnamefont {J.}~\bibnamefont {Eisert}},\ and\ \bibinfo
  {author} {\bibfnamefont {C.}~\bibnamefont {Sch{\"u}tte}},\ }\bibfield
  {title} {\bibinfo {title} {Multidimensional approximation of nonlinear
  dynamical systems},\ }\href@noop {} {\bibfield  {journal} {\bibinfo
  {journal} {Journal of Computational and Nonlinear Dynamics}\ }\textbf
  {\bibinfo {volume} {14}} (\bibinfo {year} {2019})}\BibitemShut {NoStop}%
\bibitem [{\citenamefont {Goe{\ss}mann}\ \emph {et~al.}(2020)\citenamefont
  {Goe{\ss}mann}, \citenamefont {G{\"o}tte}, \citenamefont {Roth},
  \citenamefont {Sweke}, \citenamefont {Kutyniok},\ and\ \citenamefont
  {Eisert}}]{goessmann2020tensor}%
  \BibitemOpen
  \bibfield  {author} {\bibinfo {author} {\bibfnamefont {A.}~\bibnamefont
  {Goe{\ss}mann}}, \bibinfo {author} {\bibfnamefont {M.}~\bibnamefont
  {G{\"o}tte}}, \bibinfo {author} {\bibfnamefont {I.}~\bibnamefont {Roth}},
  \bibinfo {author} {\bibfnamefont {R.}~\bibnamefont {Sweke}}, \bibinfo
  {author} {\bibfnamefont {G.}~\bibnamefont {Kutyniok}},\ and\ \bibinfo
  {author} {\bibfnamefont {J.}~\bibnamefont {Eisert}},\ }\bibfield  {title}
  {\bibinfo {title} {Tensor network approaches for learning non-linear
  dynamical laws},\ }\href@noop {} {\bibfield  {journal} {\bibinfo  {journal}
  {arXiv preprint arXiv:2002.12388}\ } (\bibinfo {year} {2020})}\BibitemShut
  {NoStop}%
\bibitem [{\citenamefont {Boninsegna}\ \emph {et~al.}(2018)\citenamefont
  {Boninsegna}, \citenamefont {N{\"u}ske},\ and\ \citenamefont
  {Clementi}}]{boninsegna2018sparse}%
  \BibitemOpen
  \bibfield  {author} {\bibinfo {author} {\bibfnamefont {L.}~\bibnamefont
  {Boninsegna}}, \bibinfo {author} {\bibfnamefont {F.}~\bibnamefont
  {N{\"u}ske}},\ and\ \bibinfo {author} {\bibfnamefont {C.}~\bibnamefont
  {Clementi}},\ }\bibfield  {title} {\bibinfo {title} {Sparse learning of
  stochastic dynamical equations},\ }\href@noop {} {\bibfield  {journal}
  {\bibinfo  {journal} {The Journal of Chemical Physics}\ }\textbf {\bibinfo
  {volume} {148}},\ \bibinfo {pages} {241723} (\bibinfo {year}
  {2018})}\BibitemShut {NoStop}%
\bibitem [{\citenamefont {Lorenz}(1963)}]{lorenz1963deterministic}%
  \BibitemOpen
  \bibfield  {author} {\bibinfo {author} {\bibfnamefont {E.~N.}\ \bibnamefont
  {Lorenz}},\ }\bibfield  {title} {\bibinfo {title} {Deterministic nonperiodic
  flow},\ }\href@noop {} {\bibfield  {journal} {\bibinfo  {journal} {Journal of
  atmospheric sciences}\ }\textbf {\bibinfo {volume} {20}},\ \bibinfo {pages}
  {130} (\bibinfo {year} {1963})}\BibitemShut {NoStop}%
\bibitem [{\citenamefont {de~Silva}\ \emph
  {et~al.}(2020{\natexlab{b}})\citenamefont {de~Silva}, \citenamefont
  {Champion}, \citenamefont {Quade}, \citenamefont {Loiseau}, \citenamefont
  {Kutz},\ and\ \citenamefont {Brunton}}]{silva2020pysindy}%
  \BibitemOpen
  \bibfield  {author} {\bibinfo {author} {\bibfnamefont {B.}~\bibnamefont
  {de~Silva}}, \bibinfo {author} {\bibfnamefont {K.}~\bibnamefont {Champion}},
  \bibinfo {author} {\bibfnamefont {M.}~\bibnamefont {Quade}}, \bibinfo
  {author} {\bibfnamefont {J.-C.}\ \bibnamefont {Loiseau}}, \bibinfo {author}
  {\bibfnamefont {J.~N.}\ \bibnamefont {Kutz}},\ and\ \bibinfo {author}
  {\bibfnamefont {S.}~\bibnamefont {Brunton}},\ }\bibfield  {title} {\bibinfo
  {title} {{PySINDy: A P}ython package for the sparse identification of
  nonlinear dynamical systems from data},\ }\href@noop {} {\bibfield  {journal}
  {\bibinfo  {journal} {Journal of Open Source Software}\ }\textbf {\bibinfo
  {volume} {5}},\ \bibinfo {pages} {1} (\bibinfo {year}
  {2020}{\natexlab{b}})}\BibitemShut {NoStop}%
\bibitem [{\citenamefont {Kaptanoglu}\ \emph
  {et~al.}(2022{\natexlab{a}})\citenamefont {Kaptanoglu}, \citenamefont
  {de~Silva}, \citenamefont {Fasel}, \citenamefont {Kaheman}, \citenamefont
  {Goldschmidt}, \citenamefont {Callaham}, \citenamefont {Delahunt},
  \citenamefont {Nicolaou}, \citenamefont {Champion}, \citenamefont {Loiseau},
  \citenamefont {Kutz},\ and\ \citenamefont {Brunton}}]{Kaptanoglu2022}%
  \BibitemOpen
  \bibfield  {author} {\bibinfo {author} {\bibfnamefont {A.~A.}\ \bibnamefont
  {Kaptanoglu}}, \bibinfo {author} {\bibfnamefont {B.~M.}\ \bibnamefont
  {de~Silva}}, \bibinfo {author} {\bibfnamefont {U.}~\bibnamefont {Fasel}},
  \bibinfo {author} {\bibfnamefont {K.}~\bibnamefont {Kaheman}}, \bibinfo
  {author} {\bibfnamefont {A.~J.}\ \bibnamefont {Goldschmidt}}, \bibinfo
  {author} {\bibfnamefont {J.}~\bibnamefont {Callaham}}, \bibinfo {author}
  {\bibfnamefont {C.~B.}\ \bibnamefont {Delahunt}}, \bibinfo {author}
  {\bibfnamefont {Z.~G.}\ \bibnamefont {Nicolaou}}, \bibinfo {author}
  {\bibfnamefont {K.}~\bibnamefont {Champion}}, \bibinfo {author}
  {\bibfnamefont {J.-C.}\ \bibnamefont {Loiseau}}, \bibinfo {author}
  {\bibfnamefont {J.~N.}\ \bibnamefont {Kutz}},\ and\ \bibinfo {author}
  {\bibfnamefont {S.~L.}\ \bibnamefont {Brunton}},\ }\bibfield  {title}
  {\bibinfo {title} {{PySINDy}: {A} comprehensive {P}ython package for robust
  sparse system identification},\ }\href {https://doi.org/10.21105/joss.03994}
  {\bibfield  {journal} {\bibinfo  {journal} {Journal of Open Source Software}\
  }\textbf {\bibinfo {volume} {7}},\ \bibinfo {pages} {3994} (\bibinfo {year}
  {2022}{\natexlab{a}})}\BibitemShut {NoStop}%
\bibitem [{\citenamefont {Zheng}\ \emph {et~al.}(2019)\citenamefont {Zheng},
  \citenamefont {Askham}, \citenamefont {Brunton}, \citenamefont {Kutz},\ and\
  \citenamefont {Aravkin}}]{zheng2019unified}%
  \BibitemOpen
  \bibfield  {author} {\bibinfo {author} {\bibfnamefont {P.}~\bibnamefont
  {Zheng}}, \bibinfo {author} {\bibfnamefont {T.}~\bibnamefont {Askham}},
  \bibinfo {author} {\bibfnamefont {S.~L.}\ \bibnamefont {Brunton}}, \bibinfo
  {author} {\bibfnamefont {J.~N.}\ \bibnamefont {Kutz}},\ and\ \bibinfo
  {author} {\bibfnamefont {A.~Y.}\ \bibnamefont {Aravkin}},\ }\bibfield
  {title} {\bibinfo {title} {A unified framework for sparse relaxed regularized
  regression: {SR}3},\ }\href@noop {} {\bibfield  {journal} {\bibinfo
  {journal} {IEEE Access}\ }\textbf {\bibinfo {volume} {7}},\ \bibinfo {pages}
  {1404} (\bibinfo {year} {2019})}\BibitemShut {NoStop}%
\bibitem [{\citenamefont {Tibshirani}\ \emph {et~al.}(2015)\citenamefont
  {Tibshirani}, \citenamefont {Wainwright},\ and\ \citenamefont
  {Hastie}}]{tibshirani2015statistical}%
  \BibitemOpen
  \bibfield  {author} {\bibinfo {author} {\bibfnamefont {R.}~\bibnamefont
  {Tibshirani}}, \bibinfo {author} {\bibfnamefont {M.}~\bibnamefont
  {Wainwright}},\ and\ \bibinfo {author} {\bibfnamefont {T.}~\bibnamefont
  {Hastie}},\ }\href@noop {} {\emph {\bibinfo {title} {Statistical learning
  with sparsity: the lasso and generalizations}}}\ (\bibinfo  {publisher}
  {Chapman and Hall/CRC},\ \bibinfo {year} {2015})\BibitemShut {NoStop}%
\bibitem [{\citenamefont {Bertsimas}\ and\ \citenamefont
  {Gurnee}(2023)}]{bertsimas2022learning}%
  \BibitemOpen
  \bibfield  {author} {\bibinfo {author} {\bibfnamefont {D.}~\bibnamefont
  {Bertsimas}}\ and\ \bibinfo {author} {\bibfnamefont {W.}~\bibnamefont
  {Gurnee}},\ }\bibfield  {title} {\bibinfo {title} {Learning sparse nonlinear
  dynamics via mixed-integer optimization},\ }\href@noop {} {\bibfield
  {journal} {\bibinfo  {journal} {Nonlinear Dynamics}\ ,\ \bibinfo {pages} {1}}
  (\bibinfo {year} {2023})}\BibitemShut {NoStop}%
\bibitem [{\citenamefont {Schaeffer}\ \emph {et~al.}(2017)\citenamefont
  {Schaeffer}, \citenamefont {Tran},\ and\ \citenamefont
  {Ward}}]{schaeffer2017learning}%
  \BibitemOpen
  \bibfield  {author} {\bibinfo {author} {\bibfnamefont {H.}~\bibnamefont
  {Schaeffer}}, \bibinfo {author} {\bibfnamefont {G.}~\bibnamefont {Tran}},\
  and\ \bibinfo {author} {\bibfnamefont {R.}~\bibnamefont {Ward}},\ }\bibfield
  {title} {\bibinfo {title} {Learning dynamical systems and bifurcation via
  group sparsity},\ }\href@noop {} {\bibfield  {journal} {\bibinfo  {journal}
  {arXiv preprint arXiv:1709.01558}\ } (\bibinfo {year} {2017})}\BibitemShut
  {NoStop}%
\bibitem [{\citenamefont {Uy}\ \emph {et~al.}(2011)\citenamefont {Uy},
  \citenamefont {Hoai}, \citenamefont {O’Neill}, \citenamefont {McKay},\ and\
  \citenamefont {Galv{\'a}n-L{\'o}pez}}]{uy2011semantically}%
  \BibitemOpen
  \bibfield  {author} {\bibinfo {author} {\bibfnamefont {N.~Q.}\ \bibnamefont
  {Uy}}, \bibinfo {author} {\bibfnamefont {N.~X.}\ \bibnamefont {Hoai}},
  \bibinfo {author} {\bibfnamefont {M.}~\bibnamefont {O’Neill}}, \bibinfo
  {author} {\bibfnamefont {R.~I.}\ \bibnamefont {McKay}},\ and\ \bibinfo
  {author} {\bibfnamefont {E.}~\bibnamefont {Galv{\'a}n-L{\'o}pez}},\
  }\bibfield  {title} {\bibinfo {title} {Semantically-based crossover in
  genetic programming: application to real-valued symbolic regression},\
  }\href@noop {} {\bibfield  {journal} {\bibinfo  {journal} {Genetic
  Programming and Evolvable Machines}\ }\textbf {\bibinfo {volume} {12}},\
  \bibinfo {pages} {91} (\bibinfo {year} {2011})}\BibitemShut {NoStop}%
\bibitem [{\citenamefont {Petersen}\ \emph {et~al.}(2021)\citenamefont
  {Petersen}, \citenamefont {Larma}, \citenamefont {Mundhenk}, \citenamefont
  {Santiago}, \citenamefont {Kim},\ and\ \citenamefont
  {Kim}}]{petersen2019deep}%
  \BibitemOpen
  \bibfield  {author} {\bibinfo {author} {\bibfnamefont {B.~K.}\ \bibnamefont
  {Petersen}}, \bibinfo {author} {\bibfnamefont {M.~L.}\ \bibnamefont {Larma}},
  \bibinfo {author} {\bibfnamefont {T.~N.}\ \bibnamefont {Mundhenk}}, \bibinfo
  {author} {\bibfnamefont {C.~P.}\ \bibnamefont {Santiago}}, \bibinfo {author}
  {\bibfnamefont {S.~K.}\ \bibnamefont {Kim}},\ and\ \bibinfo {author}
  {\bibfnamefont {J.~T.}\ \bibnamefont {Kim}},\ }\bibfield  {title} {\bibinfo
  {title} {Deep symbolic regression: {R}ecovering mathematical expressions from
  data via risk-seeking policy gradients},\ }in\ \href
  {https://openreview.net/forum?id=m5Qsh0kBQG} {\emph {\bibinfo {booktitle}
  {International Conference on Learning Representations}}}\ (\bibinfo {year}
  {2021})\BibitemShut {NoStop}%
\bibitem [{\citenamefont {Kantz}\ and\ \citenamefont
  {Schreiber}(2004)}]{kantz2004nonlinear}%
  \BibitemOpen
  \bibfield  {author} {\bibinfo {author} {\bibfnamefont {H.}~\bibnamefont
  {Kantz}}\ and\ \bibinfo {author} {\bibfnamefont {T.}~\bibnamefont
  {Schreiber}},\ }\href@noop {} {\emph {\bibinfo {title} {Nonlinear time series
  analysis}}},\ Vol.~\bibinfo {volume} {7}\ (\bibinfo  {publisher} {Cambridge
  university press},\ \bibinfo {year} {2004})\BibitemShut {NoStop}%
\bibitem [{\citenamefont {Kaptanoglu}(2021)}]{kaptanoglu2021exploration}%
  \BibitemOpen
  \bibfield  {author} {\bibinfo {author} {\bibfnamefont {A.}~\bibnamefont
  {Kaptanoglu}},\ }\href@noop {} {\emph {\bibinfo {title} {An Exploration of
  Data-Driven System Identification and Machine Learning for Plasma Physics}}}\
  (\bibinfo  {publisher} {University of Washington},\ \bibinfo {year}
  {2021})\BibitemShut {NoStop}%
\bibitem [{\citenamefont {Van~Breugel}\ \emph {et~al.}(2020)\citenamefont
  {Van~Breugel}, \citenamefont {Kutz},\ and\ \citenamefont
  {Brunton}}]{van2020numerical}%
  \BibitemOpen
  \bibfield  {author} {\bibinfo {author} {\bibfnamefont {F.}~\bibnamefont
  {Van~Breugel}}, \bibinfo {author} {\bibfnamefont {J.~N.}\ \bibnamefont
  {Kutz}},\ and\ \bibinfo {author} {\bibfnamefont {B.~W.}\ \bibnamefont
  {Brunton}},\ }\bibfield  {title} {\bibinfo {title} {Numerical differentiation
  of noisy data: A unifying multi-objective optimization framework},\
  }\href@noop {} {\bibfield  {journal} {\bibinfo  {journal} {IEEE Access}\
  }\textbf {\bibinfo {volume} {8}},\ \bibinfo {pages} {196865} (\bibinfo {year}
  {2020})}\BibitemShut {NoStop}%
\bibitem [{\citenamefont {Van~Breugel}\ \emph {et~al.}(2022)\citenamefont
  {Van~Breugel}, \citenamefont {Liu}, \citenamefont {Brunton},\ and\
  \citenamefont {Kutz}}]{van2022pynumdiff}%
  \BibitemOpen
  \bibfield  {author} {\bibinfo {author} {\bibfnamefont {F.}~\bibnamefont
  {Van~Breugel}}, \bibinfo {author} {\bibfnamefont {Y.}~\bibnamefont {Liu}},
  \bibinfo {author} {\bibfnamefont {B.~W.}\ \bibnamefont {Brunton}},\ and\
  \bibinfo {author} {\bibfnamefont {J.~N.}\ \bibnamefont {Kutz}},\ }\bibfield
  {title} {\bibinfo {title} {Pynumdiff: A python package for numerical
  differentiation of noisy time-series data},\ }\href@noop {} {\bibfield
  {journal} {\bibinfo  {journal} {Journal of Open Source Software}\ }\textbf
  {\bibinfo {volume} {7}},\ \bibinfo {pages} {4078} (\bibinfo {year}
  {2022})}\BibitemShut {NoStop}%
\bibitem [{\citenamefont {Blasco}\ \emph {et~al.}(2008)\citenamefont {Blasco},
  \citenamefont {Herrero}, \citenamefont {Sanchis},\ and\ \citenamefont
  {Mart{\'\i}nez}}]{blasco2008new}%
  \BibitemOpen
  \bibfield  {author} {\bibinfo {author} {\bibfnamefont {X.}~\bibnamefont
  {Blasco}}, \bibinfo {author} {\bibfnamefont {J.~M.}\ \bibnamefont {Herrero}},
  \bibinfo {author} {\bibfnamefont {J.}~\bibnamefont {Sanchis}},\ and\ \bibinfo
  {author} {\bibfnamefont {M.}~\bibnamefont {Mart{\'\i}nez}},\ }\bibfield
  {title} {\bibinfo {title} {A new graphical visualization of n-dimensional
  {P}areto front for decision-making in multiobjective optimization},\
  }\href@noop {} {\bibfield  {journal} {\bibinfo  {journal} {Information
  Sciences}\ }\textbf {\bibinfo {volume} {178}},\ \bibinfo {pages} {3908}
  (\bibinfo {year} {2008})}\BibitemShut {NoStop}%
\bibitem [{\citenamefont {La~Cava}\ \emph {et~al.}()\citenamefont {La~Cava},
  \citenamefont {Orzechowski}, \citenamefont {Burlacu}, \citenamefont
  {de~Franca}, \citenamefont {Virgolin}, \citenamefont {Jin}, \citenamefont
  {Kommenda},\ and\ \citenamefont {Moore}}]{la2021contemporary}%
  \BibitemOpen
  \bibfield  {author} {\bibinfo {author} {\bibfnamefont {W.}~\bibnamefont
  {La~Cava}}, \bibinfo {author} {\bibfnamefont {P.}~\bibnamefont
  {Orzechowski}}, \bibinfo {author} {\bibfnamefont {B.}~\bibnamefont
  {Burlacu}}, \bibinfo {author} {\bibfnamefont {F.~O.}\ \bibnamefont
  {de~Franca}}, \bibinfo {author} {\bibfnamefont {M.}~\bibnamefont {Virgolin}},
  \bibinfo {author} {\bibfnamefont {Y.}~\bibnamefont {Jin}}, \bibinfo {author}
  {\bibfnamefont {M.}~\bibnamefont {Kommenda}},\ and\ \bibinfo {author}
  {\bibfnamefont {J.~H.}\ \bibnamefont {Moore}},\ }\bibfield  {title} {\bibinfo
  {title} {Contemporary symbolic regression methods and their relative
  performance},\ }in\ \href@noop {} {\emph {\bibinfo {booktitle} {Thirty-fifth
  Conference on Neural Information Processing Systems Datasets and Benchmarks
  Track (Round 1)}}}\BibitemShut {NoStop}%
\bibitem [{\citenamefont {Orzechowski}\ \emph {et~al.}(2018)\citenamefont
  {Orzechowski}, \citenamefont {La~Cava},\ and\ \citenamefont
  {Moore}}]{orzechowski2018we}%
  \BibitemOpen
  \bibfield  {author} {\bibinfo {author} {\bibfnamefont {P.}~\bibnamefont
  {Orzechowski}}, \bibinfo {author} {\bibfnamefont {W.}~\bibnamefont
  {La~Cava}},\ and\ \bibinfo {author} {\bibfnamefont {J.~H.}\ \bibnamefont
  {Moore}},\ }\bibfield  {title} {\bibinfo {title} {Where are we now? {A} large
  benchmark study of recent symbolic regression methods},\ }in\ \href@noop {}
  {\emph {\bibinfo {booktitle} {Proceedings of the Genetic and Evolutionary
  Computation Conference}}}\ (\bibinfo {year} {2018})\ pp.\ \bibinfo {pages}
  {1183--1190}\BibitemShut {NoStop}%
\bibitem [{\citenamefont {Bhat}(2019)}]{bhat2019learning}%
  \BibitemOpen
  \bibfield  {author} {\bibinfo {author} {\bibfnamefont {H.~S.}\ \bibnamefont
  {Bhat}},\ }\bibfield  {title} {\bibinfo {title} {Learning and interpreting
  potentials for classical {H}amiltonian systems},\ }in\ \href@noop {} {\emph
  {\bibinfo {booktitle} {Joint European Conference on Machine Learning and
  Knowledge Discovery in Databases}}}\ (\bibinfo {organization} {Springer},\
  \bibinfo {year} {2019})\ pp.\ \bibinfo {pages} {217--228}\BibitemShut
  {NoStop}%
\bibitem [{\citenamefont {Chu}\ and\ \citenamefont
  {Hayashibe}(2020)}]{chu2020discovering}%
  \BibitemOpen
  \bibfield  {author} {\bibinfo {author} {\bibfnamefont {H.~K.}\ \bibnamefont
  {Chu}}\ and\ \bibinfo {author} {\bibfnamefont {M.}~\bibnamefont
  {Hayashibe}},\ }\bibfield  {title} {\bibinfo {title} {Discovering
  interpretable dynamics by sparsity promotion on energy and the
  {L}agrangian},\ }\href@noop {} {\bibfield  {journal} {\bibinfo  {journal}
  {IEEE Robotics and Automation Letters}\ }\textbf {\bibinfo {volume} {5}},\
  \bibinfo {pages} {2154} (\bibinfo {year} {2020})}\BibitemShut {NoStop}%
\bibitem [{\citenamefont {Bertalan}\ \emph {et~al.}(2019)\citenamefont
  {Bertalan}, \citenamefont {Dietrich}, \citenamefont {Mezi{\'c}},\ and\
  \citenamefont {Kevrekidis}}]{bertalan2019learning}%
  \BibitemOpen
  \bibfield  {author} {\bibinfo {author} {\bibfnamefont {T.}~\bibnamefont
  {Bertalan}}, \bibinfo {author} {\bibfnamefont {F.}~\bibnamefont {Dietrich}},
  \bibinfo {author} {\bibfnamefont {I.}~\bibnamefont {Mezi{\'c}}},\ and\
  \bibinfo {author} {\bibfnamefont {I.~G.}\ \bibnamefont {Kevrekidis}},\
  }\bibfield  {title} {\bibinfo {title} {On learning {H}amiltonian systems from
  data},\ }\href@noop {} {\bibfield  {journal} {\bibinfo  {journal} {Chaos: An
  Interdisciplinary Journal of Nonlinear Science}\ }\textbf {\bibinfo {volume}
  {29}},\ \bibinfo {pages} {121107} (\bibinfo {year} {2019})}\BibitemShut
  {NoStop}%
\bibitem [{\citenamefont {Mikhaeil}\ \emph {et~al.}(2021)\citenamefont
  {Mikhaeil}, \citenamefont {Monfared},\ and\ \citenamefont
  {Durstewitz}}]{mikhaeil2021difficulty}%
  \BibitemOpen
  \bibfield  {author} {\bibinfo {author} {\bibfnamefont {J.~M.}\ \bibnamefont
  {Mikhaeil}}, \bibinfo {author} {\bibfnamefont {Z.}~\bibnamefont {Monfared}},\
  and\ \bibinfo {author} {\bibfnamefont {D.}~\bibnamefont {Durstewitz}},\
  }\bibfield  {title} {\bibinfo {title} {On the difficulty of learning chaotic
  dynamics with {RNNs}},\ }\href@noop {} {\bibfield  {journal} {\bibinfo
  {journal} {arXiv preprint arXiv:2110.07238}\ } (\bibinfo {year}
  {2021})}\BibitemShut {NoStop}%
\bibitem [{\citenamefont {Ouala}\ \emph {et~al.}(2023)\citenamefont {Ouala},
  \citenamefont {Brunton}, \citenamefont {Chapron}, \citenamefont {Pascual},
  \citenamefont {Collard}, \citenamefont {Gaultier},\ and\ \citenamefont
  {Fablet}}]{ouala2023bounded}%
  \BibitemOpen
  \bibfield  {author} {\bibinfo {author} {\bibfnamefont {S.}~\bibnamefont
  {Ouala}}, \bibinfo {author} {\bibfnamefont {S.~L.}\ \bibnamefont {Brunton}},
  \bibinfo {author} {\bibfnamefont {B.}~\bibnamefont {Chapron}}, \bibinfo
  {author} {\bibfnamefont {A.}~\bibnamefont {Pascual}}, \bibinfo {author}
  {\bibfnamefont {F.}~\bibnamefont {Collard}}, \bibinfo {author} {\bibfnamefont
  {L.}~\bibnamefont {Gaultier}},\ and\ \bibinfo {author} {\bibfnamefont
  {R.}~\bibnamefont {Fablet}},\ }\bibfield  {title} {\bibinfo {title} {Bounded
  nonlinear forecasts of partially observed geophysical systems with
  physics-constrained deep learning},\ }\href@noop {} {\bibfield  {journal}
  {\bibinfo  {journal} {Physica D: Nonlinear Phenomena}\ ,\ \bibinfo {pages}
  {133630}} (\bibinfo {year} {2023})}\BibitemShut {NoStop}%
\bibitem [{\citenamefont {Diamond}\ and\ \citenamefont
  {Boyd}(2016)}]{diamond2016cvxpy}%
  \BibitemOpen
  \bibfield  {author} {\bibinfo {author} {\bibfnamefont {S.}~\bibnamefont
  {Diamond}}\ and\ \bibinfo {author} {\bibfnamefont {S.}~\bibnamefont {Boyd}},\
  }\bibfield  {title} {\bibinfo {title} {{CVXPY}: A {P}ython-embedded modeling
  language for convex optimization},\ }\href@noop {} {\bibfield  {journal}
  {\bibinfo  {journal} {The Journal of Machine Learning Research}\ }\textbf
  {\bibinfo {volume} {17}},\ \bibinfo {pages} {2909} (\bibinfo {year}
  {2016})}\BibitemShut {NoStop}%
\bibitem [{\citenamefont {Kaptanoglu}\ \emph
  {et~al.}(2022{\natexlab{b}})\citenamefont {Kaptanoglu}, \citenamefont {Qian},
  \citenamefont {Wechsung},\ and\ \citenamefont
  {Landreman}}]{kaptanoglu2022permanent}%
  \BibitemOpen
  \bibfield  {author} {\bibinfo {author} {\bibfnamefont {A.~A.}\ \bibnamefont
  {Kaptanoglu}}, \bibinfo {author} {\bibfnamefont {T.}~\bibnamefont {Qian}},
  \bibinfo {author} {\bibfnamefont {F.}~\bibnamefont {Wechsung}},\ and\
  \bibinfo {author} {\bibfnamefont {M.}~\bibnamefont {Landreman}},\ }\bibfield
  {title} {\bibinfo {title} {Permanent-magnet optimization for stellarators as
  sparse regression},\ }\href@noop {} {\bibfield  {journal} {\bibinfo
  {journal} {Physical Review Applied}\ }\textbf {\bibinfo {volume} {18}},\
  \bibinfo {pages} {044006} (\bibinfo {year} {2022}{\natexlab{b}})}\BibitemShut
  {NoStop}%
\bibitem [{\citenamefont {Wainwright}(2009)}]{wainwright2009sharp}%
  \BibitemOpen
  \bibfield  {author} {\bibinfo {author} {\bibfnamefont {M.~J.}\ \bibnamefont
  {Wainwright}},\ }\bibfield  {title} {\bibinfo {title} {Sharp thresholds for
  high-dimensional and noisy sparsity recovery using $l-1$-constrained
  quadratic programming {(L}asso)},\ }\href@noop {} {\bibfield  {journal}
  {\bibinfo  {journal} {IEEE transactions on information theory}\ }\textbf
  {\bibinfo {volume} {55}},\ \bibinfo {pages} {2183} (\bibinfo {year}
  {2009})}\BibitemShut {NoStop}%
\bibitem [{\citenamefont {Bertsimas}\ \emph {et~al.}(2020)\citenamefont
  {Bertsimas}, \citenamefont {Pauphilet},\ and\ \citenamefont
  {Parys}}]{bertsimas_review}%
  \BibitemOpen
  \bibfield  {author} {\bibinfo {author} {\bibfnamefont {D.}~\bibnamefont
  {Bertsimas}}, \bibinfo {author} {\bibfnamefont {J.}~\bibnamefont
  {Pauphilet}},\ and\ \bibinfo {author} {\bibfnamefont {B.~V.}\ \bibnamefont
  {Parys}},\ }\bibfield  {title} {\bibinfo {title} {{Sparse Regression:
  {S}calable Algorithms and Empirical Performance}},\ }\href
  {https://doi.org/10.1214/19-STS701} {\bibfield  {journal} {\bibinfo
  {journal} {Statistical Science}\ }\textbf {\bibinfo {volume} {35}},\ \bibinfo
  {pages} {555 } (\bibinfo {year} {2020})}\BibitemShut {NoStop}%
\bibitem [{\citenamefont {Sommerer}\ and\ \citenamefont
  {Ott}(1993)}]{sommerer1993particles}%
  \BibitemOpen
  \bibfield  {author} {\bibinfo {author} {\bibfnamefont {J.~C.}\ \bibnamefont
  {Sommerer}}\ and\ \bibinfo {author} {\bibfnamefont {E.}~\bibnamefont {Ott}},\
  }\bibfield  {title} {\bibinfo {title} {Particles floating on a moving fluid:
  {A} dynamically comprehensible physical fractal},\ }\href@noop {} {\bibfield
  {journal} {\bibinfo  {journal} {Science}\ }\textbf {\bibinfo {volume}
  {259}},\ \bibinfo {pages} {335} (\bibinfo {year} {1993})}\BibitemShut
  {NoStop}%
\bibitem [{\citenamefont {Pavliotis}\ and\ \citenamefont
  {Stuart}(2008)}]{pavliotis2008multiscale}%
  \BibitemOpen
  \bibfield  {author} {\bibinfo {author} {\bibfnamefont {G.}~\bibnamefont
  {Pavliotis}}\ and\ \bibinfo {author} {\bibfnamefont {A.}~\bibnamefont
  {Stuart}},\ }\href@noop {} {\emph {\bibinfo {title} {Multiscale methods:
  averaging and homogenization}}}\ (\bibinfo  {publisher} {Springer Science \&
  Business Media},\ \bibinfo {year} {2008})\BibitemShut {NoStop}%
\bibitem [{\citenamefont {Bucci}\ \emph {et~al.}(2021)\citenamefont {Bucci},
  \citenamefont {Semeraro}, \citenamefont {Allauzen}, \citenamefont
  {Chibbaro},\ and\ \citenamefont {Mathelin}}]{bucci2022}%
  \BibitemOpen
  \bibfield  {author} {\bibinfo {author} {\bibfnamefont {A.}~\bibnamefont
  {Bucci}}, \bibinfo {author} {\bibfnamefont {O.}~\bibnamefont {Semeraro}},
  \bibinfo {author} {\bibfnamefont {A.}~\bibnamefont {Allauzen}}, \bibinfo
  {author} {\bibfnamefont {S.}~\bibnamefont {Chibbaro}},\ and\ \bibinfo
  {author} {\bibfnamefont {L.}~\bibnamefont {Mathelin}},\ }\href
  {https://doi.org/10.48550/ARXIV.2112.08458} {\bibinfo {title} {Curriculum
  learning for data-driven modeling of dynamical systems}} (\bibinfo {year}
  {2021})\BibitemShut {NoStop}%
\bibitem [{\citenamefont {Bramburger}\ \emph {et~al.}(2020)\citenamefont
  {Bramburger}, \citenamefont {Dylewsky},\ and\ \citenamefont
  {Kutz}}]{bramburger2020sparse}%
  \BibitemOpen
  \bibfield  {author} {\bibinfo {author} {\bibfnamefont {J.~J.}\ \bibnamefont
  {Bramburger}}, \bibinfo {author} {\bibfnamefont {D.}~\bibnamefont
  {Dylewsky}},\ and\ \bibinfo {author} {\bibfnamefont {J.~N.}\ \bibnamefont
  {Kutz}},\ }\bibfield  {title} {\bibinfo {title} {Sparse identification of
  slow timescale dynamics},\ }\href@noop {} {\bibfield  {journal} {\bibinfo
  {journal} {Physical Review E}\ }\textbf {\bibinfo {volume} {102}},\ \bibinfo
  {pages} {022204} (\bibinfo {year} {2020})}\BibitemShut {NoStop}%
\bibitem [{\citenamefont {Cenedese}\ \emph {et~al.}(2022)\citenamefont
  {Cenedese}, \citenamefont {Ax{\aa}s}, \citenamefont {B{\"a}uerlein},
  \citenamefont {Avila},\ and\ \citenamefont {Haller}}]{cenedese2022data}%
  \BibitemOpen
  \bibfield  {author} {\bibinfo {author} {\bibfnamefont {M.}~\bibnamefont
  {Cenedese}}, \bibinfo {author} {\bibfnamefont {J.}~\bibnamefont {Ax{\aa}s}},
  \bibinfo {author} {\bibfnamefont {B.}~\bibnamefont {B{\"a}uerlein}}, \bibinfo
  {author} {\bibfnamefont {K.}~\bibnamefont {Avila}},\ and\ \bibinfo {author}
  {\bibfnamefont {G.}~\bibnamefont {Haller}},\ }\bibfield  {title} {\bibinfo
  {title} {Data-driven modeling and prediction of non-linearizable dynamics via
  spectral submanifolds},\ }\href@noop {} {\bibfield  {journal} {\bibinfo
  {journal} {Nature communications}\ }\textbf {\bibinfo {volume} {13}},\
  \bibinfo {pages} {1} (\bibinfo {year} {2022})}\BibitemShut {NoStop}%
\bibitem [{\citenamefont {Szalai}(2022)}]{szalai2022data}%
  \BibitemOpen
  \bibfield  {author} {\bibinfo {author} {\bibfnamefont {R.}~\bibnamefont
  {Szalai}},\ }\bibfield  {title} {\bibinfo {title} {Data-driven reduced order
  models using invariant foliations, manifolds and autoencoders},\ }\href@noop
  {} {\bibfield  {journal} {\bibinfo  {journal} {arXiv preprint
  arXiv:2206.12269}\ } (\bibinfo {year} {2022})}\BibitemShut {NoStop}%
\bibitem [{\citenamefont {Ax{\aa}s}\ \emph {et~al.}(2022)\citenamefont
  {Ax{\aa}s}, \citenamefont {Cenedese},\ and\ \citenamefont
  {Haller}}]{axaas2022fast}%
  \BibitemOpen
  \bibfield  {author} {\bibinfo {author} {\bibfnamefont {J.}~\bibnamefont
  {Ax{\aa}s}}, \bibinfo {author} {\bibfnamefont {M.}~\bibnamefont {Cenedese}},\
  and\ \bibinfo {author} {\bibfnamefont {G.}~\bibnamefont {Haller}},\
  }\bibfield  {title} {\bibinfo {title} {Fast data-driven model reduction for
  nonlinear dynamical systems},\ }\href@noop {} {\bibfield  {journal} {\bibinfo
   {journal} {Nonlinear Dynamics}\ ,\ \bibinfo {pages} {1}} (\bibinfo {year}
  {2022})}\BibitemShut {NoStop}%
\bibitem [{\citenamefont {Udrescu}\ \emph {et~al.}(2020)\citenamefont
  {Udrescu}, \citenamefont {Tan}, \citenamefont {Feng}, \citenamefont {Neto},
  \citenamefont {Wu},\ and\ \citenamefont {Tegmark}}]{udrescu2020ai2}%
  \BibitemOpen
  \bibfield  {author} {\bibinfo {author} {\bibfnamefont {S.-M.}\ \bibnamefont
  {Udrescu}}, \bibinfo {author} {\bibfnamefont {A.}~\bibnamefont {Tan}},
  \bibinfo {author} {\bibfnamefont {J.}~\bibnamefont {Feng}}, \bibinfo {author}
  {\bibfnamefont {O.}~\bibnamefont {Neto}}, \bibinfo {author} {\bibfnamefont
  {T.}~\bibnamefont {Wu}},\ and\ \bibinfo {author} {\bibfnamefont
  {M.}~\bibnamefont {Tegmark}},\ }\bibfield  {title} {\bibinfo {title} {{AI
  Feynman 2.0:} {P}areto-optimal symbolic regression exploiting graph
  modularity},\ }\href@noop {} {\bibfield  {journal} {\bibinfo  {journal}
  {Advances in Neural Information Processing Systems}\ }\textbf {\bibinfo
  {volume} {33}},\ \bibinfo {pages} {4860} (\bibinfo {year}
  {2020})}\BibitemShut {NoStop}%
\bibitem [{\citenamefont {Gr{\"u}nwald}\ \emph {et~al.}(2005)\citenamefont
  {Gr{\"u}nwald}, \citenamefont {Myung},\ and\ \citenamefont
  {Pitt}}]{grunwald2005advances}%
  \BibitemOpen
  \bibfield  {author} {\bibinfo {author} {\bibfnamefont {P.~D.}\ \bibnamefont
  {Gr{\"u}nwald}}, \bibinfo {author} {\bibfnamefont {I.~J.}\ \bibnamefont
  {Myung}},\ and\ \bibinfo {author} {\bibfnamefont {M.~A.}\ \bibnamefont
  {Pitt}},\ }\href@noop {} {\emph {\bibinfo {title} {Advances in minimum
  description length: {T}heory and applications}}}\ (\bibinfo  {publisher} {MIT
  press},\ \bibinfo {year} {2005})\BibitemShut {NoStop}%
\bibitem [{\citenamefont {Vladislavleva}\ \emph {et~al.}(2008)\citenamefont
  {Vladislavleva}, \citenamefont {Smits},\ and\ \citenamefont
  {Den~Hertog}}]{vladislavleva2008order}%
  \BibitemOpen
  \bibfield  {author} {\bibinfo {author} {\bibfnamefont {E.~J.}\ \bibnamefont
  {Vladislavleva}}, \bibinfo {author} {\bibfnamefont {G.~F.}\ \bibnamefont
  {Smits}},\ and\ \bibinfo {author} {\bibfnamefont {D.}~\bibnamefont
  {Den~Hertog}},\ }\bibfield  {title} {\bibinfo {title} {Order of nonlinearity
  as a complexity measure for models generated by symbolic regression via
  pareto genetic programming},\ }\href@noop {} {\bibfield  {journal} {\bibinfo
  {journal} {IEEE Transactions on Evolutionary Computation}\ }\textbf {\bibinfo
  {volume} {13}},\ \bibinfo {pages} {333} (\bibinfo {year} {2008})}\BibitemShut
  {NoStop}%
\bibitem [{\citenamefont {Murdoch}\ \emph {et~al.}(2019)\citenamefont
  {Murdoch}, \citenamefont {Singh}, \citenamefont {Kumbier}, \citenamefont
  {Abbasi-Asl},\ and\ \citenamefont {Yu}}]{murdoch2019definitions}%
  \BibitemOpen
  \bibfield  {author} {\bibinfo {author} {\bibfnamefont {W.~J.}\ \bibnamefont
  {Murdoch}}, \bibinfo {author} {\bibfnamefont {C.}~\bibnamefont {Singh}},
  \bibinfo {author} {\bibfnamefont {K.}~\bibnamefont {Kumbier}}, \bibinfo
  {author} {\bibfnamefont {R.}~\bibnamefont {Abbasi-Asl}},\ and\ \bibinfo
  {author} {\bibfnamefont {B.}~\bibnamefont {Yu}},\ }\bibfield  {title}
  {\bibinfo {title} {Definitions, methods, and applications in interpretable
  machine learning},\ }\href@noop {} {\bibfield  {journal} {\bibinfo  {journal}
  {Proceedings of the National Academy of Sciences}\ }\textbf {\bibinfo
  {volume} {116}},\ \bibinfo {pages} {22071} (\bibinfo {year}
  {2019})}\BibitemShut {NoStop}%
\bibitem [{\citenamefont {Tesi}\ \emph {et~al.}(1996)\citenamefont {Tesi},
  \citenamefont {Villoresi},\ and\ \citenamefont
  {Genesio}}]{tesi1996stability}%
  \BibitemOpen
  \bibfield  {author} {\bibinfo {author} {\bibfnamefont {A.}~\bibnamefont
  {Tesi}}, \bibinfo {author} {\bibfnamefont {F.}~\bibnamefont {Villoresi}},\
  and\ \bibinfo {author} {\bibfnamefont {R.}~\bibnamefont {Genesio}},\
  }\bibfield  {title} {\bibinfo {title} {On the stability domain estimation via
  a quadratic {L}yapunov function: convexity and optimality properties for
  polynomial systems},\ }\href@noop {} {\bibfield  {journal} {\bibinfo
  {journal} {IEEE Transactions on Automatic Control}\ }\textbf {\bibinfo
  {volume} {41}},\ \bibinfo {pages} {1650} (\bibinfo {year}
  {1996})}\BibitemShut {NoStop}%
\bibitem [{\citenamefont {Ahmadi}\ \emph {et~al.}(2013)\citenamefont {Ahmadi},
  \citenamefont {Majumdar},\ and\ \citenamefont
  {Tedrake}}]{ahmadi2013complexity}%
  \BibitemOpen
  \bibfield  {author} {\bibinfo {author} {\bibfnamefont {A.~A.}\ \bibnamefont
  {Ahmadi}}, \bibinfo {author} {\bibfnamefont {A.}~\bibnamefont {Majumdar}},\
  and\ \bibinfo {author} {\bibfnamefont {R.}~\bibnamefont {Tedrake}},\
  }\bibfield  {title} {\bibinfo {title} {Complexity of ten decision problems in
  continuous time dynamical systems},\ }in\ \href@noop {} {\emph {\bibinfo
  {booktitle} {2013 American Control Conference}}}\ (\bibinfo {organization}
  {IEEE},\ \bibinfo {year} {2013})\ pp.\ \bibinfo {pages}
  {6376--6381}\BibitemShut {NoStop}%
\bibitem [{\citenamefont {Dikeman}\ \emph {et~al.}(2022)\citenamefont
  {Dikeman}, \citenamefont {Zhang},\ and\ \citenamefont
  {Yang}}]{dikeman2022stiffness}%
  \BibitemOpen
  \bibfield  {author} {\bibinfo {author} {\bibfnamefont {H.~E.}\ \bibnamefont
  {Dikeman}}, \bibinfo {author} {\bibfnamefont {H.}~\bibnamefont {Zhang}},\
  and\ \bibinfo {author} {\bibfnamefont {S.}~\bibnamefont {Yang}},\ }\bibfield
  {title} {\bibinfo {title} {Stiffness-reduced neural {ODE} models for
  data-driven reduced-order modeling of combustion chemical kinetics},\ }in\
  \href@noop {} {\emph {\bibinfo {booktitle} {AIAA SCITECH 2022 Forum}}}\
  (\bibinfo {year} {2022})\ p.\ \bibinfo {pages} {0226}\BibitemShut {NoStop}%
\bibitem [{\citenamefont {Ott}\ \emph {et~al.}(1990)\citenamefont {Ott},
  \citenamefont {Grebogi},\ and\ \citenamefont {Yorke}}]{ott1990controlling}%
  \BibitemOpen
  \bibfield  {author} {\bibinfo {author} {\bibfnamefont {E.}~\bibnamefont
  {Ott}}, \bibinfo {author} {\bibfnamefont {C.}~\bibnamefont {Grebogi}},\ and\
  \bibinfo {author} {\bibfnamefont {J.~A.}\ \bibnamefont {Yorke}},\ }\bibfield
  {title} {\bibinfo {title} {Controlling chaos},\ }\href@noop {} {\bibfield
  {journal} {\bibinfo  {journal} {Physical review letters}\ }\textbf {\bibinfo
  {volume} {64}},\ \bibinfo {pages} {1196} (\bibinfo {year}
  {1990})}\BibitemShut {NoStop}%
\end{thebibliography}%

\end{document}